\documentclass[runningheads]{llncs}

\usepackage{eccv}

\usepackage{eccvabbrv}

\usepackage{graphicx}
\usepackage{booktabs}

\usepackage[accsupp]{axessibility}  

\makeatletter
\def\thanks#1{\protected@xdef\@thanks{\@thanks\protect\footnotetext{#1}}}
\makeatother

\definecolor{C-Model}{HTML}{ffefe0}
\definecolor{I-Model}{HTML}{E6ECE3}
\definecolor{O-Model}{HTML}{E9F9D1}
\definecolor{darkgreen}{HTML}{04bf29}
\definecolor{darkred}{HTML}{D1191F}
\usepackage[pagebackref,breaklinks,colorlinks,citecolor=eccvblue]{hyperref}

\usepackage{orcidlink}

\usepackage{multirow}
\usepackage{xcolor}  
\usepackage{color, colortbl}
\usepackage{array}
\usepackage{wrapfig}
\usepackage[utf8]{inputenc} 
\usepackage[T1]{fontenc}    
\usepackage{url}            
\usepackage{booktabs}       
\usepackage{amsfonts}       
\usepackage{nicefrac}       
\usepackage{microtype}      
\usepackage{xcolor}         
\usepackage{listings}
\usepackage{epsfig}
\usepackage{graphicx}
\usepackage{amsmath}
\usepackage{amssymb}
\usepackage{multirow}
\usepackage{color, colortbl}
\usepackage{subcaption}
\usepackage{pifont}
\usepackage{todonotes}
\usepackage{bbding}
\usepackage{wasysym}
\usepackage{comment}
\usepackage{bm}
\usepackage{makecell}
\usepackage{wrapfig}
\usepackage{multicol}
\usepackage[misc]{ifsym}

\begin{document}

\title{
  \includegraphics[width=0.044\textwidth]{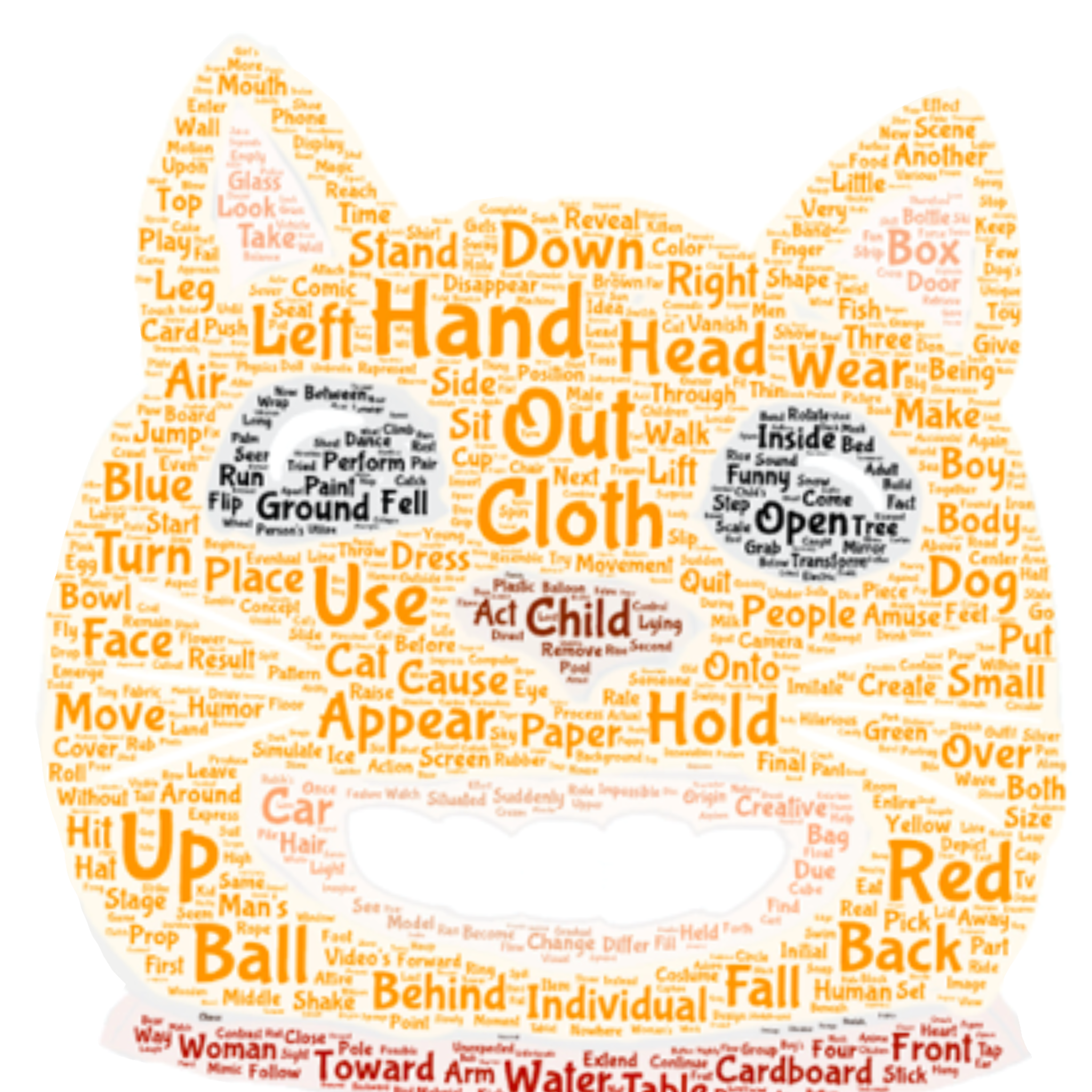}
  FunQA: Towards Surprising Video Comprehension 
}




\author{Binzhu Xie$^{\ast}$\thanks{\scriptsize $^{\ast}$ indicates equal contribution.\quad\textsuperscript{\Letter} Corresponding author. Contact: \texttt{ziwei.liu@ntu.edu.sg}}\inst{1}\and
Sicheng Zhang$^{\ast}$\inst{1} \and
Zitang Zhou$^{\ast}$\inst{1} \and \\
Bo Li\inst{2} \and
Yuanhan Zhang\inst{2} \and
Jack Hessel\inst{3} \and
Jingkang Yang\inst{2} \and
Ziwei Liu\inst{2}\textsuperscript{\Letter}
}

\authorrunning{Xie et al.}

\institute{Beijing University of Posts and Telecommunications, Beijing, China \and
S-Lab, Nanyang Technological University, Singapore \and
The Allen Institute for AI, WA, USA\\
\texttt{\url{https://github.com/Jingkang50/FunQA}
}
}

\maketitle
\vspace{-25pt}
\begin{figure}[htbp]
\centering
\includegraphics[width=0.9\linewidth]{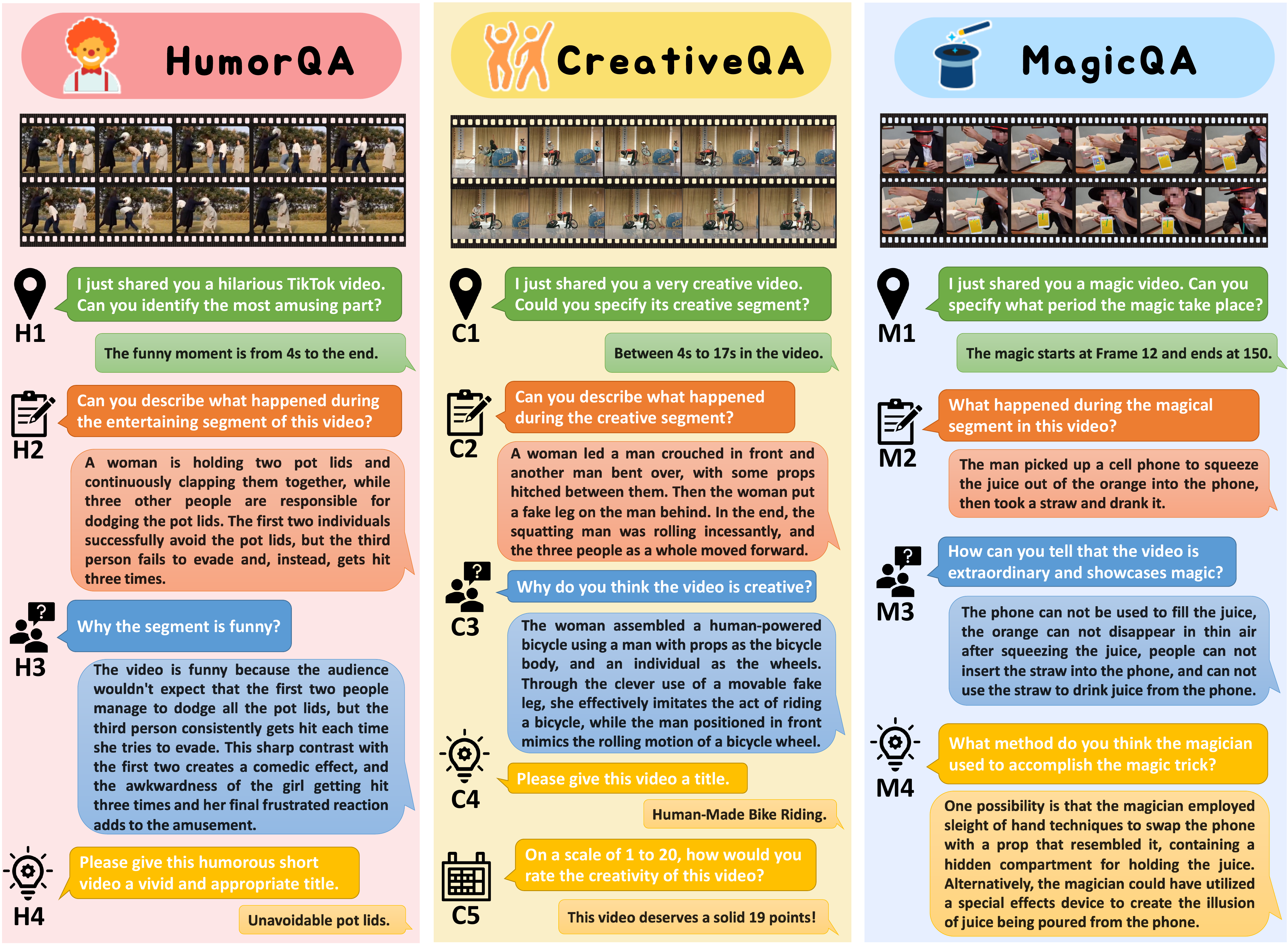}
\captionof{figure}{
\textbf{Overview of FunQA.} FunQA comprises three subsets of surprising videos: \textit{1) HumorQA, 2) CreativeQA,} and \textit{3) MagicQA.} Each subset is associated with three common tasks: \textit{1) counter-intuitive timestamp localization}, \textit{2) detailed video description}, and \textit{3) reasoning around counter-intuitiveness} (see \textbf{H1-3}, \textbf{C1-3}, and \textbf{M1-3}). Furthermore, we offer higher-level tasks tailored for each video type, such as \textit{attributing a fitting and vivid title} for HumorQA and CreativeQA (see \textbf{H4}, \textbf{C4}), etc.}
\vspace{-35pt}
\label{fig:motivation}
\end{figure}

\begin{abstract}
Surprising videos, \emph{e.g.}, funny clips, creative performances, or visual illusions, attract significant attention. Enjoyment of these videos is not simply a response to visual stimuli; rather, it hinges on the human capacity to understand (and appreciate) commonsense violations depicted in these videos.
We introduce \textbf{FunQA}, a challenging video question answering (QA) dataset specifically designed to evaluate and enhance the depth of video reasoning based on counter-intuitive and fun videos. Unlike most video QA benchmarks which focus on less surprising contexts, e.g., cooking or instructional videos, FunQA covers three previously unexplored types of surprising videos: \textbf{1) HumorQA}, \textbf{2) CreativeQA}, and \textbf{3) MagicQA}.
For each subset, we establish rigorous QA tasks designed to assess the model's capability in counter-intuitive timestamp localization, detailed video description, and reasoning around counter-intuitiveness.
We also pose higher-level tasks, such as attributing a fitting and vivid title to the video, and scoring the video creativity.
In total, the FunQA benchmark consists of 312K free-text QA pairs derived from 4.3K video clips, spanning a total of 24 video hours.
Moreover, we propose \textbf{FunMentor}, an agent designed for Vision-Language Models (VLMs) that uses multi-turn dialogues to enhance models' understanding of counter-intuitiveness. Extensive experiments with existing VLMs demonstrate the effectiveness of FunMentor and reveal significant performance gaps for the FunQA videos across spatial-temporal reasoning, visual-centered reasoning, and free-text generation.
\end{abstract}

\section{Introduction}
The charm of surprising videos, being funny, creative, and filled with visual illusions, 
offers enjoyment and attracts engagement from viewers.
This type of media elicits \emph{positive surprise}\footnote{c.f., \emph{negative} surprise, e.g., a surprising medical bill.} \cite{noordewier2013valence},
a captivating emotion that stems not merely from perceiving surface-level visual stimuli, but rather, the innate ability of humans to understand and find delight in unexpected and counter-intuitive moments \cite{martin1983humour}. However, despite significant advancements in today's computer vision models, the question remains: can video models ``understand'' the humor/creativity in surprising videos? 
Consider the \href{https://www.facebook.com/kiosgim/videos/3766868883332108}{humorous video} depicted in Fig.~\ref{fig:motivation} (left) as an example. We observe a woman in black holding two pot lids and clapping them together. The remaining three individuals are responsible for avoiding the pot lids. The first two people successfully dodge, but the third girl, in a panic, fails to avoid any hits and gets struck three times. The embarrassed demeanor of the third girl along with her final frustrated reaction, elicits laughter\footnote{The hostility/superiority theory of humor posits that
humor can arise from claiming superiority over someone or something \cite{gruner1978understanding,billig2005laughter}; but alternate (more optimistic) theories of humor exist, \cite{attardo2008primer} offers a survey.}. While humans effortlessly recognize this as an unusual (and potentially entertaining) event, 
the reasoning required to holistically understand the scene is complex:
a model needs to 
recognize that this is not a video depicting harm but rather girls \textit{engaging in playful pranks together}, and discern that the comedic element arises from \textit{the stark contrast between the third girl being hit by the pot lids every time and the first two girls skillfully avoiding them}.

\smallskip
\begin{table}[!t]
\centering
\caption{\textbf{Comparison between FunQA and other existing benchmarks.}
Compared to other datasets, FunQA revolves around the captivating realm of interesting and counter-intuitive videos. The tasks within FunQA are specifically designed to challenge the vision capabilities of models, requiring strong skills in producing an in-depth description, interpretation, and spatial-temporal reasoning.
Here we clarify the abbreviation in the table. For annotation type:  \includegraphics[width=0.02\textwidth]{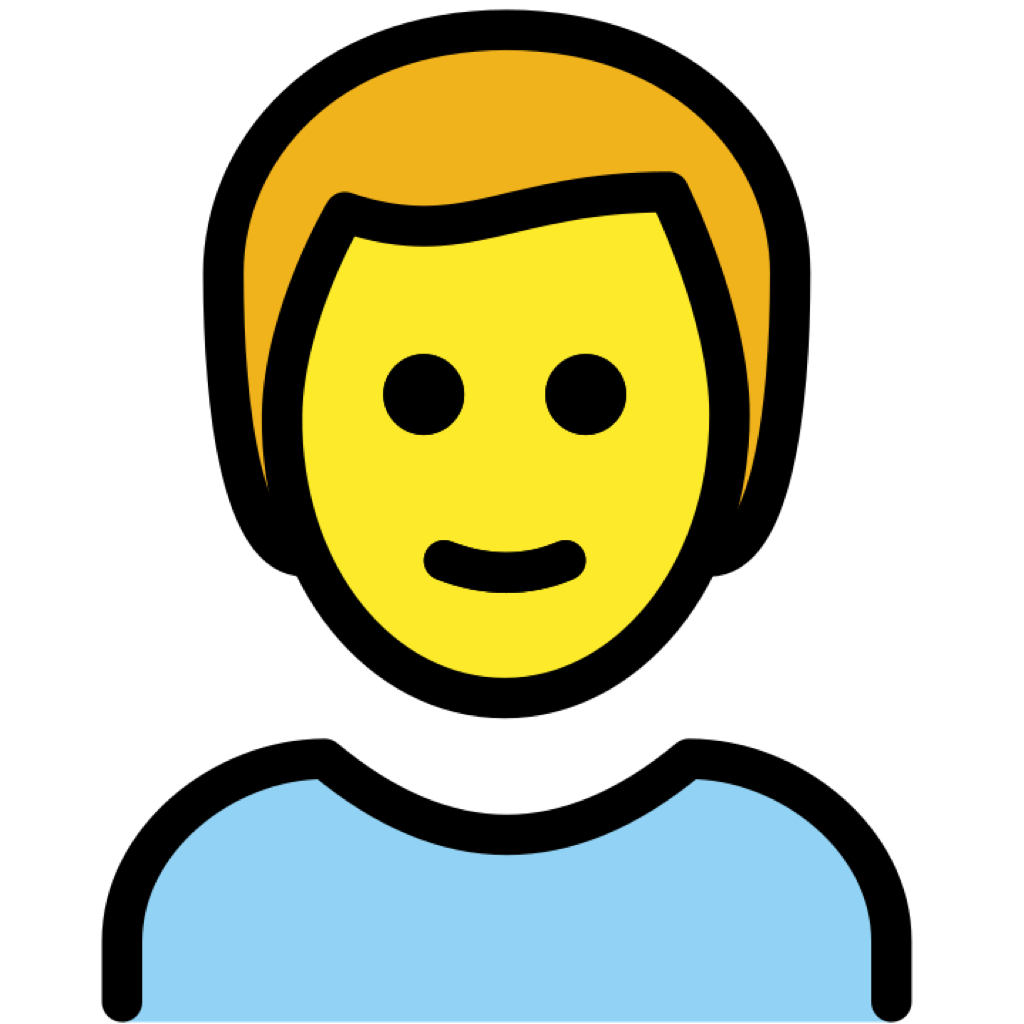} denotes Manual Annotation and  \includegraphics[width=0.02\textwidth]{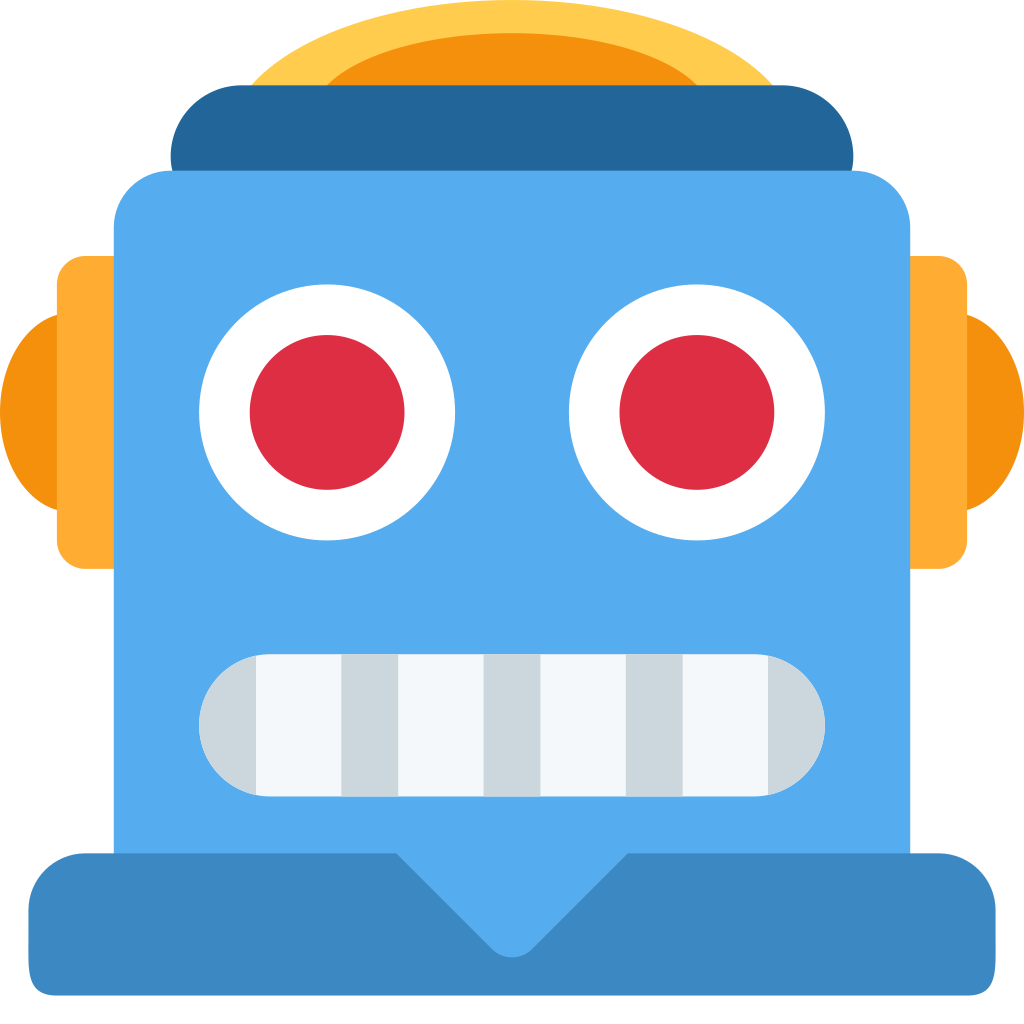} for Automatic Annotation; \textbf{Avg Len} denotes video average length; \textbf{\# Clips} means number of video clips; \textbf{VC} for visual-centric, \textbf{Des.} for Description, \textbf{Exp.} for Explanation, \textbf{STR} for Spatial-temporal Reasoning, \textbf{MC} means Multiple Choice QA, and \textbf{OE} shows Open Ended QA with \textbf{Average Word Count} per response. 
}
\vspace{-5pt}
\label{T:main}
\setlength\tabcolsep{10pt}
\resizebox{1.0\linewidth}{!}{
\begin{tabular}{llccccccccccl}
\toprule
\multirow{2}{*}{Dataset} & \multirow{2}{*}{Domain} & \multirow{2}{*}{
\includegraphics[width=0.02\textwidth]{figures/human.png} or \includegraphics[width=0.02\textwidth]{figures/robot.png}
} & \multicolumn{2}{c}{Video} & \multicolumn{7}{c}{Question Answer}  \\
\cmidrule(l){4-5} \cmidrule(l){6-12} 
 &  &  & Avg Len & \# Clips                      & \# QA  & VC & Des. & Exp. & STR & MC & OE   \\
 \toprule
TGIF-QA \cite{jang-IJCV-2019} 
& Social Media & \includegraphics[width=0.02\textwidth]{figures/robot.png} & 3s & 72K & 165K & \textcolor{darkgreen}{\ding{51}} & \textcolor{darkgreen}{\ding{51}} & \textcolor{darkred}{\ding{55}} & \textcolor{darkgreen}{\ding{51}} & \textcolor{darkred}{\ding{55}} & 2.1 \\

MSRVTT-QA \cite{xu2017video}
& Social Media & \includegraphics[width=0.02\textwidth]{figures/robot.png} & 15s & 10K & 244K & \textcolor{darkgreen}{\ding{51}} & \textcolor{darkred}{\ding{55}} & \textcolor{darkred}{\ding{55}} & \textcolor{darkgreen}{\ding{51}} & \textcolor{darkred}{\ding{55}} & 1.0 \\

ActivityNet-QA \cite{yu2019activitynetqa}
& Social Media & \includegraphics[width=0.02\textwidth]{figures/human.png} & 180s & 6K & 58K & \textcolor{darkgreen}{\ding{51}} & \textcolor{darkred}{\ding{55}} & \textcolor{darkred}{\ding{55}} & \textcolor{darkgreen}{\ding{51}} & \textcolor{darkred}{\ding{55}} & 1.9 \\
\toprule

NExT-QA \cite{xiao2021nextqanext}            
& Daily life & \includegraphics[width=0.02\textwidth]{figures/human.png} & 44s & 5K & 52K & \textcolor{darkgreen}{\ding{51}} & \textcolor{darkgreen}{\ding{51}} & \textcolor{darkgreen}{\ding{51}} & \textcolor{darkgreen}{\ding{51}} & \textcolor{darkgreen}{\ding{51}} & 2.6 \\

Social-IQ \cite{zadeh2019social}                        
& Daily life & \includegraphics[width=0.02\textwidth]{figures/human.png} & 99s & 1K & 8K & \textcolor{darkgreen}{\ding{51}} & \textcolor{darkred}{\ding{55}} & \textcolor{darkgreen}{\ding{51}} & \textcolor{darkred}{\ding{55}} & \textcolor{darkgreen}{\ding{51}} & N/A \\
\toprule

MovieQA \cite{tapaswi2016movieqa}                       & TV shows & \includegraphics[width=0.02\textwidth]{figures/robot.png} & 203s & 7K & 6K & \textcolor{darkred}{\ding{55}} & \textcolor{darkred}{\ding{55}} & \textcolor{darkgreen}{\ding{51}} & \textcolor{darkgreen}{\ding{51}} & \textcolor{darkgreen}{\ding{51}} & N/A \\

TVQA+ \cite{lei2019tvqa}  
& TV shows & \includegraphics[width=0.02\textwidth]{figures/human.png} & 8s & 4K & 30K & \textcolor{darkred}{\ding{55}} & \textcolor{darkred}{\ding{55}} & \textcolor{darkgreen}{\ding{51}} & \textcolor{darkgreen}{\ding{51}} & \textcolor{darkgreen}{\ding{51}} & N/A \\
\toprule

SUTD-TrafficQA \cite{xu2021sutd}
& Traffic & \includegraphics[width=0.02\textwidth]{figures/human.png} & 5s & 10K & 623K & \textcolor{darkgreen}{\ding{51}} & \textcolor{darkred}{\ding{55}} & \textcolor{darkred}{\ding{55}} & \textcolor{darkgreen}{\ding{51}} & \textcolor{darkgreen}{\ding{51}} & N/A \\
MarioQA \cite{mun2017marioqa} 
& Games  & \includegraphics[width=0.02\textwidth]{figures/human.png} & 5s & 188K & 188K & \textcolor{darkgreen}{\ding{51}} & \textcolor{darkred}{\ding{55}} & \textcolor{darkgreen}{\ding{51}} & \textcolor{darkgreen}{\ding{51}} & \textcolor{darkred}{\ding{55}} & 2.0 \\

CLEVRER \cite{yi2020clevrer}     
& Synthetic Videos & \includegraphics[width=0.02\textwidth]{figures/human.png} & 5s & 20K & 305K & \textcolor{darkgreen}{\ding{51}} & \textcolor{darkred}{\ding{55}} & \textcolor{darkgreen}{\ding{51}} & \textcolor{darkgreen}{\ding{51}} & \textcolor{darkgreen}{\ding{51}} & N/A \\




\midrule
\textbf{FunQA (Ours)} 
& \textbf{Surprising Videos} 
& \includegraphics[width=0.02\textwidth]{figures/human.png} & 19s & 4K & \textbf{312K} & \textcolor{darkgreen}{\ding{51}} & \textcolor{darkgreen}{\ding{51}} & \textcolor{darkgreen}{\ding{51}} & \textcolor{darkgreen}{\ding{51}} & \textcolor{darkgreen}{\ding{51}} & \textbf{34.2} \\ 
\bottomrule 

\end{tabular}}
\vspace{-0.5cm}
\end{table}

\noindent While there have been some efforts to enhance computer vision models' performance in Video Question Answering (VideoQA), these works have primarily focused on the common, less surprising videos found in existing VideoQA datasets. Examples of commonly employed VideoQA datasets include YouCook2 \cite{zhou2018towards} which contains video clips from 2K cooking videos, Howto100M \cite{miech2019howto100m} which consists of only instructional videos.
While there exist video datasets that explore the humor in TV shows \cite{mustard, hasan-etal-2019-ur} and include tasks such as predicting laughter tracks \cite{Patro_2021_WACV}, these tasks often heavily rely on audio and narrative cues, with visual clues might playing a lesser role. Beyond datasets centered on factual queries, it is worth noting that NExT-QA targets the explanation of video content, which is widely employed for evaluating reasoning abilities. However, it was found in the experiment (see Section~\ref{sec:compare_nextqa}) that VLMs such as GPT-4V(ision) already achieved an accuracy of 80\% on NExTQA. This demonstrates that with the development of VLMs, the demand for datasets with deeper reasoning capabilities and presenting greater challenges is increasing.

\noindent To revitalize the visual reasoning field and further improve model capabilities to identify and understand visual commonsense violations in videos, we introduce \textbf{FunQA}, an extensive and carefully curated VideoQA dataset comprising 4.3K surprising videos and 312K manually annotated \textbf{free-text} QA pairs. Unlike some VideoQA datasets that feature open-ended questions but short answers (e.g., an average of 2.6 words per answer in NExT-QA~\cite{xiao2021nextqanext}), FunQA's responses average 34.2 words in length. This significantly increases the demand for advanced video comprehension capabilities in the model. Therefore, here we use \textbf{free-text} QA to distinguish from open-ended QA. Our dataset consists of three subsets: \textbf{1) HumorQA}, \textbf{2) CreativeQA}, and \textbf{3) MagicQA}. Each subset covers different sources and contents, but the commonality lies in their surprising nature, e.g., the unexpected contrasts in humorous videos, the intriguing disguises in creative videos, and the seemingly impossible performances in magic videos. Our experiments suggest that these surprising videos require different types of reasoning than common videos, as existing VideoQA methods perform poorly on the corpus. With FunQA, we hope to provide a benchmark that covers the popular, important, and sophisticated genre of counter-intuitive/surprising videos. 

\noindent In FunQA, we formulate three rigorous tasks to measure models' understanding of surprise:
\textbf{1) Counter-intuitive timestamp localization}: a model must identify the specific time period within a video when an unexpected event takes place.
\textbf{2) Detailed video description}: a model must generate coherent and objective descriptions of the video content, evaluating models' fundamental video understanding capabilities.
\textbf{3) Counter-intuitiveness reasoning}: a model must generate concrete explanations of why the video is surprising. 
These tasks progressively assess the model's ability to perceive, articulate, and reason about the counter-intuitive elements present in surprising videos. We also propose additional tasks that pose higher-level challenges, such as assigning an appropriate and vivid title to the video.

\noindent In continuation of our efforts to enhance models' comprehension of surprising content, we introduce \textbf{FunMentor}, a specialized agent designed to boost counter-intuitive reasoning in VLMs. Operating like a seasoned coach in a variety show, FunMentor engages in detailed, multi-turn dialogues, honing the models' responses to accurately grasp the essence of both amusing and astonishing content. Beyond just evaluation, FunMentor actively steers VLMs with precise prompts, fostering fluent, logical, and persuasive responses. Experiments have demonstrated its effectiveness in augmenting VLMs' ability to comprehend.


\noindent To summarize our contributions:\\
\textbf{1) New VideoQA Dataset:} We build a large-scale dataset \textbf{FunQA}, which complements the existing VideoQA dataset with intriguing videos.\\
\textbf{2) Novel and Challenging Tasks:} We design a number of novel tasks that allow the model to explore previously untouched problems, such as timestamp localization, and reasoning around counter-intuitiveness. These tasks push video reasoning beyond superficial descriptions, demanding deeper understanding.\\
\textbf{3) Novel Method named FunMentor:} We propose a novel agent that refines the model's understanding of counter-intuitiveness through multi-turn dialogues with VLMs.\\
\textbf{4) Comprehensive Evaluation:} We have done an extensive and comprehensive evaluation of cutting-edge baselines, giving the field an insight and future research direction.

\section{Related Work}


\noindent\textbf{Video Question Answering Benchmarks}\quad
While the visual question answering (VQA) task focuses on enhancing models' ability in image comprehension \cite{goyal2017making, zhu2016visual7w, krishna2016visual}, video question answering (VideoQA) shifts the attention towards video comprehension. VideoQA is generally more challenging than VQA as it requires a comprehensive understanding of visual content, utilization of temporal and spatial information, and exploration of relationships between recognized objects and activities~\cite{yi2020clevrer}. To address the VideoQA task, the research community has introduced various benchmarks. As depicted in Table \ref{T:main} (the complete table are shown in Appendix \ref{appendix_a: FunQA more details}), most commonly used VideoQA datasets are sourced from human-centric videos like movies \cite{tapaswi2016movieqa}, TV shows \cite{lei2018tvqa, lei2019tvqa, garcia2020knowit}, and social media \cite{jang-IJCV-2019, zadeh2019social, xiao2021nextqanext, grundemclaughlin2021agqa, yang2022avqa, wu2021star, castro2022fiber}, and there are also object-centric datasets of game videos \cite{mun2017marioqa}, synthetic videos \cite{yi2020clevrer} and egocentric videos~\cite{Gao_2021_ICCV}. MovieQA \cite{tapaswi2016movieqa} and TVQA \cite{lei2018tvqa} are commonly employed by VideoQA methods, which put forward tasks related to temporal and causal reasoning. However, they rely heavily on dialogue comprehension and textual plot summaries, which severely limits the challenge of visual reasoning. TGIF-QA \cite{jang-IJCV-2019} uses animated GIFs to challenge spatial-temporal reasoning, but as most GIFs are short videos of 3 seconds, and its tasks mainly focus on action description, TGIF-QA lacks complex reasoning evaluation ability. When most datasets use multiple choice questions as QA tasks, some methods, such as NExT-QA \cite{xiao2021nextqanext}, try to join open-ended questions. NExT-QA mainly focuses on daily life videos, but the open-ended answers are mostly simple sentences containing only a few words. To sum up, most existing methods focus on ordinary videos, lack of understanding of intriguing or unexpected videos, and advanced reasoning tasks such as generating complete explanatory texts of videos remain to be explored.\\
\noindent\textbf{Video Question Answering Methods}\quad
Earlier studies have explored various models, including LSTMs and graph-based neural networks, to capture cross-modal information \cite{li2019beyond, zhao2018open}. With the advent of Transformers, video understanding models, like ClipBERT \cite{leiless} and CoMVT \cite{seo2021look} emerged, focusing on understanding specific frames within a video. Subsequent models like Violet \cite{fu2021violet}, extended their ability to encompass temporal and spatial information. However, these methods have primarily been applied to short videos. For long videos, MIST \cite{gao2022mist} stands out by achieving state-of-the-art (SOTA) performance and excelling in terms of computation efficiency and interpretability. Furthermore, recent VLMs \cite{li2023otter,2023videochat,maaz2023videochatgpt} have showcased remarkable video understanding capabilities.\\
\noindent\textbf{Counter-Intuitive Benchmarks}\quad
While many current computer vision benchmarks primarily focus on understanding commonsense content, there is a growing interest in addressing the realm of counter-intuitiveness. Several emerging benchmarks and models cater to this domain, such as Whoops~\cite{Whoop}, which emphasizes weird, unusual, and uncanny images, OOPS \cite{Epstein_2020_CVPR}, which centers on recognizing and predicting unintentional events, and MemeGraphs \cite{kougia2023memegraphs}, which revolves around memes featuring humor and sarcasm. Furthermore, some works even challenges models to comprehend complex multimodal humor in comics~\cite{hessel2022androids}. In the realm of large language models, exemplified by GPT-4 \cite{openai2023gpt4}, there is a particular focus on showcasing their ability to provide explanations for funny pictures. However, regarding videos, existing datasets exploring humor in TV shows or comedy tend to heavily rely on audio and narrative cues ~\cite{mustard, hasan-etal-2019-ur,Patro_2021_WACV}, with visual clues playing a comparatively lesser role.
\section{The FunQA Dataset}
\begin{figure}[!t]
    \centering



    \begin{subfigure}[b]{0.45\textwidth}
        \centering
        \includegraphics[width=1\textwidth]
        {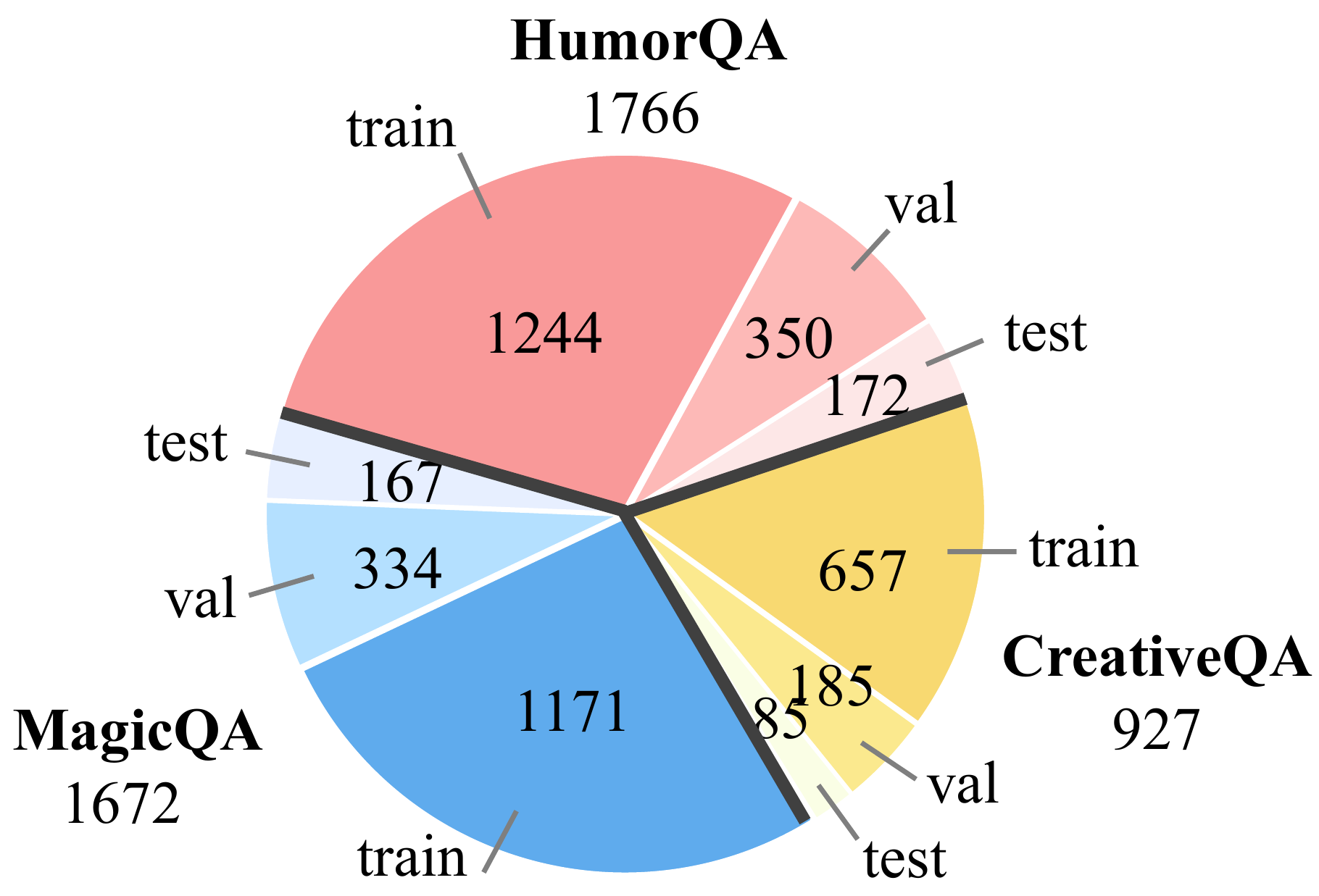}
        \caption{Number of videos and splits}
        \label{fig:fun_split}
    \end{subfigure}
    \vspace{-25pt}
    \begin{subfigure}[b]{0.42\textwidth}
        \centering
        \includegraphics[width=1\textwidth]{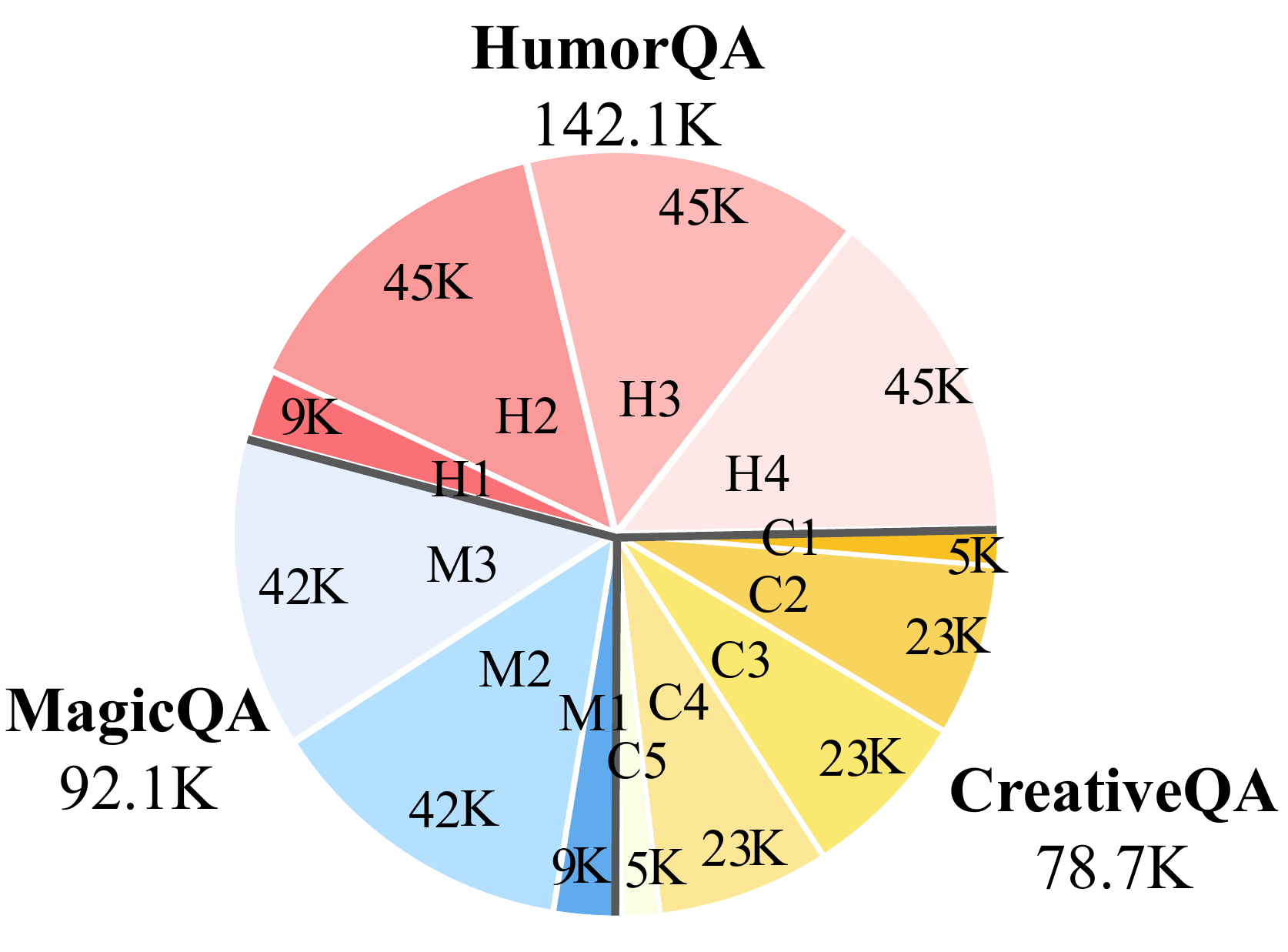}
        \caption{Number of QA pairs}
        \label{fig:fun_qa}
    \end{subfigure}
    \begin{subfigure}[b]{0.5\textwidth}
        \centering
        \includegraphics[width=1\textwidth]{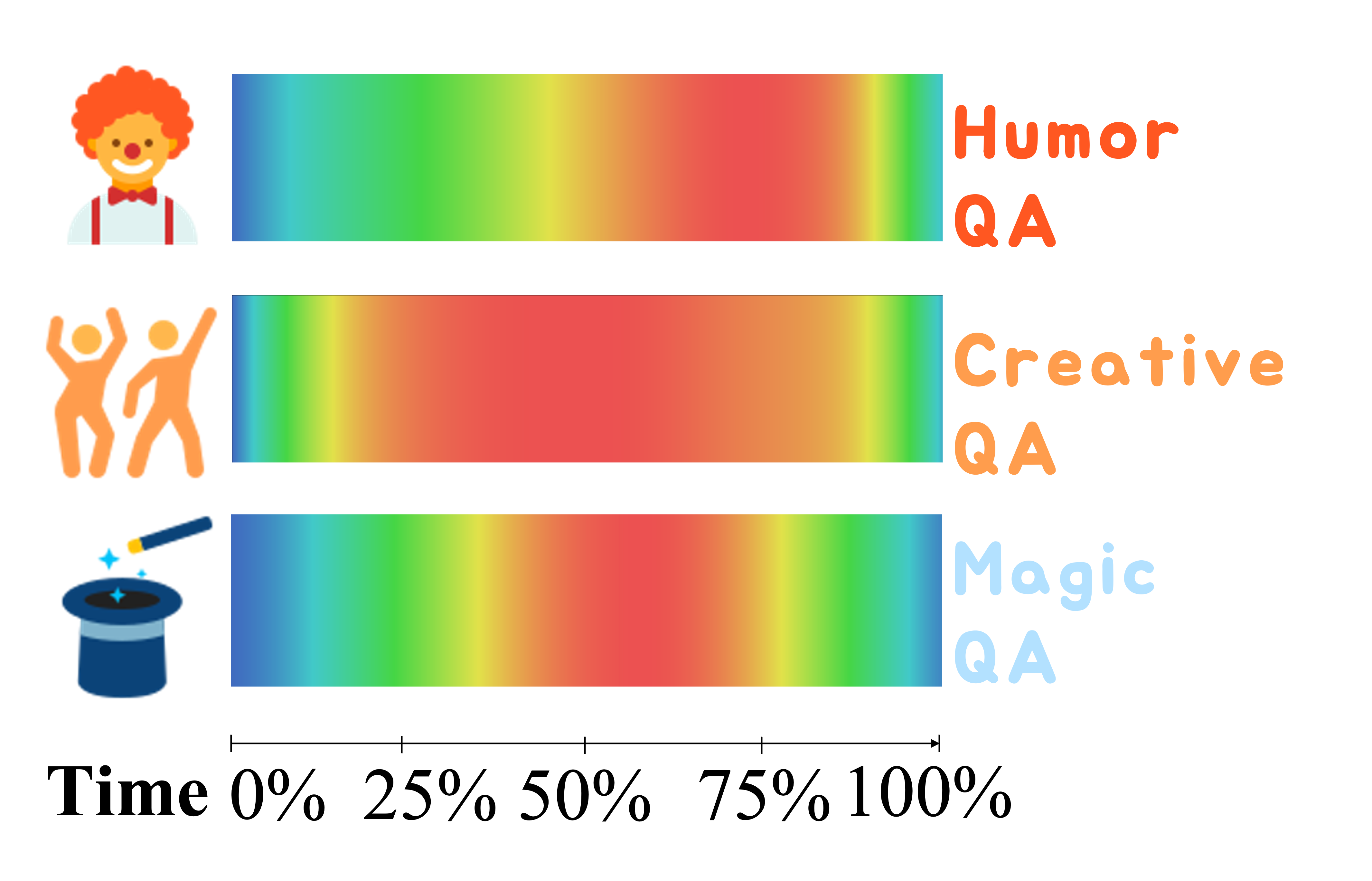}
        \caption{Time Dist. of Localization}
        \label{fig:fun_moment}
    \end{subfigure}
    \begin{subfigure}[b]{0.4\textwidth}
        \centering
        \includegraphics[width=1\textwidth]{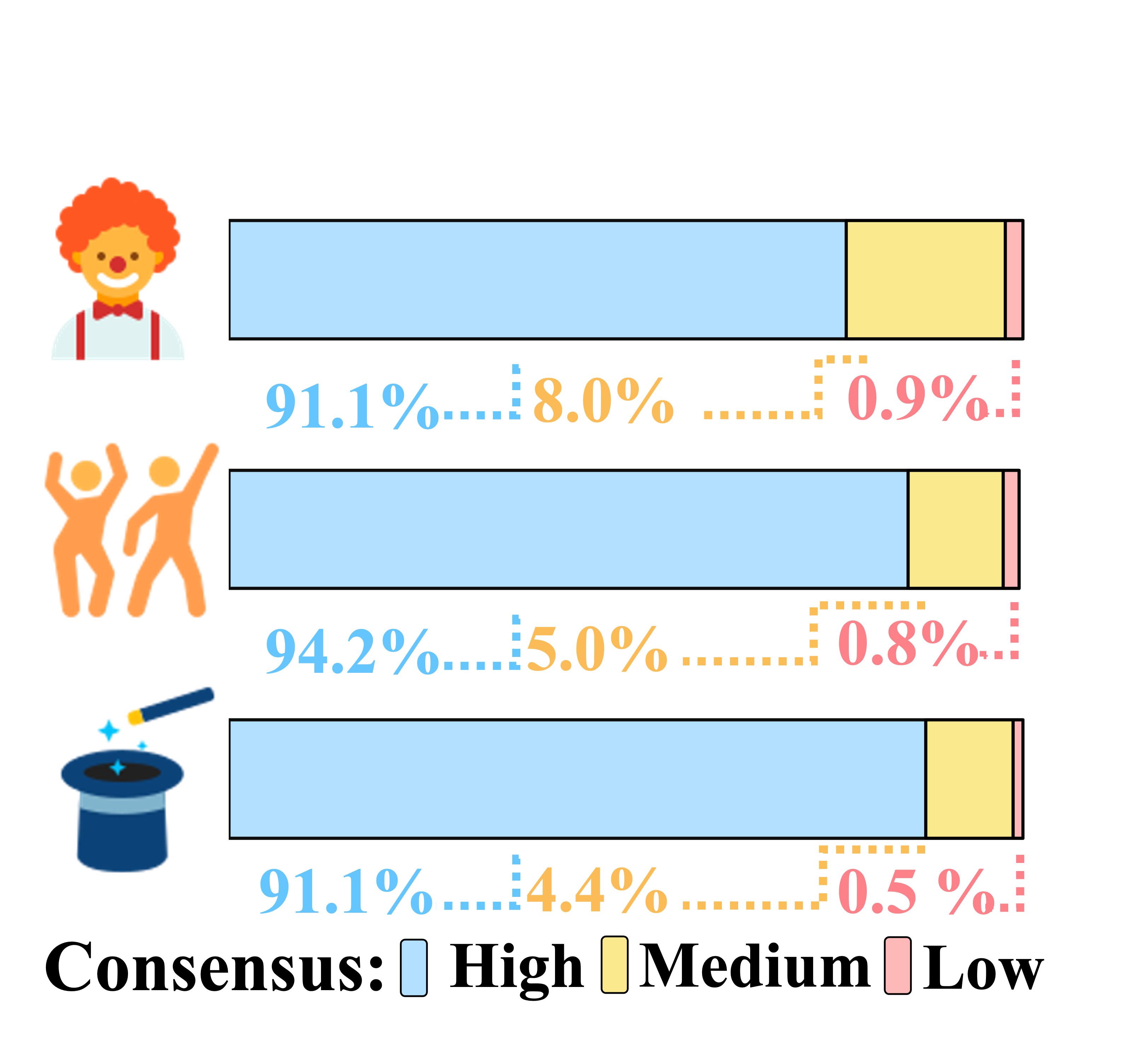}
        \caption{Consensus Evaluation}
        \label{fig:fun_val}
    \end{subfigure}

  \vspace{-10pt}
  \caption{\textbf{Statistics of FunQA Dataset.} 
 Figure (a) and (b) showcase vital statistics, including the number of videos for different videos, splits, and QA pairs count for three subsets. Figure (c) highlights the high-frequency time span of the answer for localization questions in red. Figure (d) presents the percentage of consensus between annotators for the same QA pair. The consensus is categorized into three levels \textbf{High} for consistent understanding, \textbf{Medium} for partial agreement but with mutual acknowledgment, and \textbf{Low} for complete disagreement.
  }
  \vspace{-0.4cm}
  \label{fig:dataset statistics}
\end{figure}

In this section, we provide a detailed explanation of the design principles that guided the creation of the FunQA dataset and its subsets. Additionally, we introduce our novel VideoQA tasks tailored for FunQA, and FunQA data statistics in Figure~\ref{fig:dataset statistics}. 
Toward the end we present our Construction Pipeline and Quality Control, highlighting our efforts in maintaining data quality and objectivity.

\subsection{Task Definition}

To comprehensively evaluate the model's ability to understand surprising videos, we designed the following 4 types of tasks for each subset: \\
\noindent\textbf{Counter-intuitive Timestamp Localization Task}\quad
The localization task is the base task to assess the model's comprehension abilities. It involves localizing counter-intuitive segments within the video, answers expressed in either seconds or frames. This task serves as the basis for the subsequent two main tasks in the three subsets, where the focus shifts to locating moments of humor, creativity, and magical effects, respectively. Successfully completing this task demands the model's understanding of the video's overall content, incorporating both temporal and spatial information.\\
\noindent\textbf{Detailed Description Task}\quad
The description task aims to evaluate the model's information extraction capabilities, serving as a fundamental aspect of video understanding. This task requires providing a free-text answer that describes the selected moment. Furthermore, this task allows for analysis of how the model extracts information and generates answers for subsequent tasks. By examining the model's performance in this task, we gain insights into its ability to extract relevant information and generate meaningful responses.\\
\noindent\textbf{Counter-intuitiveness Reasoning Task}\quad
The reasoning task is designed to test the model's ability to reason about the video, and in the three subsets, this question is Why Humorous, Why creative, and Why counter-intuitive and the answer is a free-text explanation. This task is very difficult and involves the model's deep reasoning ability; it requires the model to give a complete explanation using information from the entire video and its own common sense.\\
\noindent\textbf{Higher Level Tasks}\quad
In addition to the three main tasks, we design higher-level tasks to enhance the model's inference abilities on counter-intuitive videos. \textit{Title Task} in HumorQA and CreativeQA requires generating a concise title summarizing the video's content. \textit{Creative Scoring Task} in CreativeQA involves rating the creativity of videos between 1 and 20 (provided officially). \textit{Magic Method Task} in MagicQA requires the model to explain clearly the rationale behind the magic, and its purpose is to test the model's ability to reason more deeply. To ensure the accuracy of the answers, this task is only partially annotated and appears only in the test set, details of which can be found in Appendix \ref{appendix_a:dataset con}.


\subsection{Dataset Statistics}
FunQA contains \textbf{4,365} counter-intuitive video clips and \textbf{311,950} QA pairs, the total length of these videos is \textbf{23.9h} and the average length of each clips is \textbf{19} seconds. FunQA consists three fine-grained subsets, each one containing well-designed tasks. The specific numbers of videos and splits can be seen in Fig. \ref{fig:dataset statistics} (a).
The specific number of QA pairs for each task can be seen in Fig.~\ref{fig:dataset statistics} (b). For our localization task, the heat map for the three different types of videos can be seen in Fig.~\ref{fig:dataset statistics} (c), which shows the high-frequency time span of the answer. 
For the description and reasoning tasks, the average length of the words in their free-text answers reached \textbf{34.24}, which is much longer than existing VideoQA datasets (e.g., 2.6 in NExT-QA~\cite{xiao2021nextqanext}).
FunQA has a well-established annotation process and high annotation quality, the result of our annotation consensus evaluation are illustrated in Fig. \ref{fig:dataset statistics} (d).  Impressively, over 90\% of the annotations demonstrate a high level of consensus, while only 1\% exhibit low consensus. This clearly underscores the objectivity and reliability of the FunQA dataset.
\subsection{Dataset Construction Pipeline} 
\label{Section: dataset construction pipeline}
FunQA dataset construction pipeline was in three stages: Pre-processing, Manual Annotation, and Post-Processing. The whole process took about 900 hours with over 50 highly educated undergraduates as part-time annotators. See Appendix~\ref{appendix_a:dataset con} for more details on the dataset construction.\\
\noindent\textbf{Pre-Processing}\quad
Initially, we crawled videos from YouTube. Then we performed a two-stage manual cleaning and cutting process on the collection to ensure counter-intuitive features and video quality and to exclude non-ethical and sensitive content, resulting in video clips.\\
\noindent\textbf{Manual Annotation}\quad
We annotated the videos according to the characteristics of different task designs in Chinese. We screen and train the annotators to ensure the accuracy and high quality of the annotation, and finally produce the original annotated files. After the first round, we conducted a secondary round of 10\% of the tasks and performed Consensus Evaluation to ensure the objectivity.\\
\noindent\textbf{Post-Processing}\quad 
Based on our carefully designed tasks and high-quality annotations, we expanded our dataset using GPT-3.5. Firstly, we automatically translated the Chinese annotations into English. Subsequently, we generated more QA pairs that were faithful to the original ideas but presented differently. This not only made FunQA multilingual but also expanded its QA pair count to 312K. Additionally, we created diverse sub-dataset, FunQA-MC (multi-choice QA) and FunQA-DIA (dialogue QA). In addition, to focus on exploring the ability to handle counter-intuitive reasoning, we released FunQA-MC-R (a multi-choice version specifically containing counter-intuitive reasoning questions). More details are given in Appendix \ref{appendix_a: Augmentation} and Appendix \ref{appendix_b: FunQA Extended dataset}.
\begin{center}
\begin{minipage}{\textwidth}
\begin{minipage}[t]{0.48\textwidth}
\makeatletter\def\@captype{table}
\caption{\textbf{Consensus Evaluation Experiment.} This table shows the results from a random 10\% sample of QA pairs, cross-validated by annotators to assess agreement with existing annotations. `Low,' `Medium,' and `High' indicate the strength of the consensus.}
\resizebox{\linewidth}{!}{
\begin{tabular}{l|cccc}
\toprule
Consensus & HumorQA & CreativeQA & MagicQA & \# Total \\
\hline
Low & 1 & 1 & 2 & 4 \textbf{(1\%)} \\
Medium & 9 & 6 & 17 & 32 \textbf{(8\%}) \\
High & 199 & 111 & 194 & 504 \textbf{(91\%)}\\
\bottomrule
\end{tabular}}
\label{T:coneval}
\end{minipage}
\hfill
\begin{minipage}[t]{0.48\textwidth}
\makeatletter\def\@captype{table}
\caption{\textbf{Can FunQA be Solved Solely Based on Images?} The left two columns show the average number of questions answerable or not by humans using only 8 static, uniformly selected frames.}
\label{T:onlyframes}
\centering
\resizebox{\linewidth}{!}{
\begin{tabular}{l|ccc}
\toprule
Dataset  & \# Can & \# Cannot & Cannot Rate  \\
\hline
HumorQA & 7.8 & 32.2 & 80\% \\
CreativeQA & 6.1 & 33.9 & 85\% \\
MagicQA & 2.6 & 37.4 & 94\% \\
\midrule
\textbf{FunQA} & 16.5 & 103.5 & \textbf{86\%} \\
\bottomrule
\end{tabular}
}
\end{minipage}
\end{minipage}
\vspace{10pt}
\end{center}
\subsection{Quality Control}
\noindent\textbf{Minimal Errors and High Objectivity in FunQA}\quad
We assure that every annotation included in the final release of FunQA has been subjected to rigorous \textbf{multi-person}, \textbf{multi-round} review processes. Furthermore, We did manual consensus evaluation on released FunQA dataset, randomly sampling 10\% of the data from all three sub-datasets (HumorQA, CreativeQA, and MagicQA). As shown in Table \ref{T:coneval}, we get the 91\% high consensus.\\
\noindent\textbf{FunQA Emphasis on Temporal Dynamics}\quad
FunQA requires a strong emphases on temporal dynamics rather than solely on few frames of images. To prove that, we did the quantitative human experiments - We randomly selected 40 videos from each of the three sub-datasets, totaling 120 videos. For each video, we sampled 8 frames evenly. We enlisted 10 individuals who had not seen any FunQA videos before. We had them view the sequence of 8 consecutive frames and then watch the original video along with its annotations. They were asked to determine whether they can understand and answer the counter-intuitive understanding of the original video solely based on the images. Nearly \textbf{86\%} people thought that FunQA cannot be solved only by images, as shown in Table~\ref{T:onlyframes}.\\

\section{FunMentor}
\begin{wrapfigure}{r}{0.5\textwidth}
    \centering
    \vspace{-30pt}
    \includegraphics[width=\linewidth]{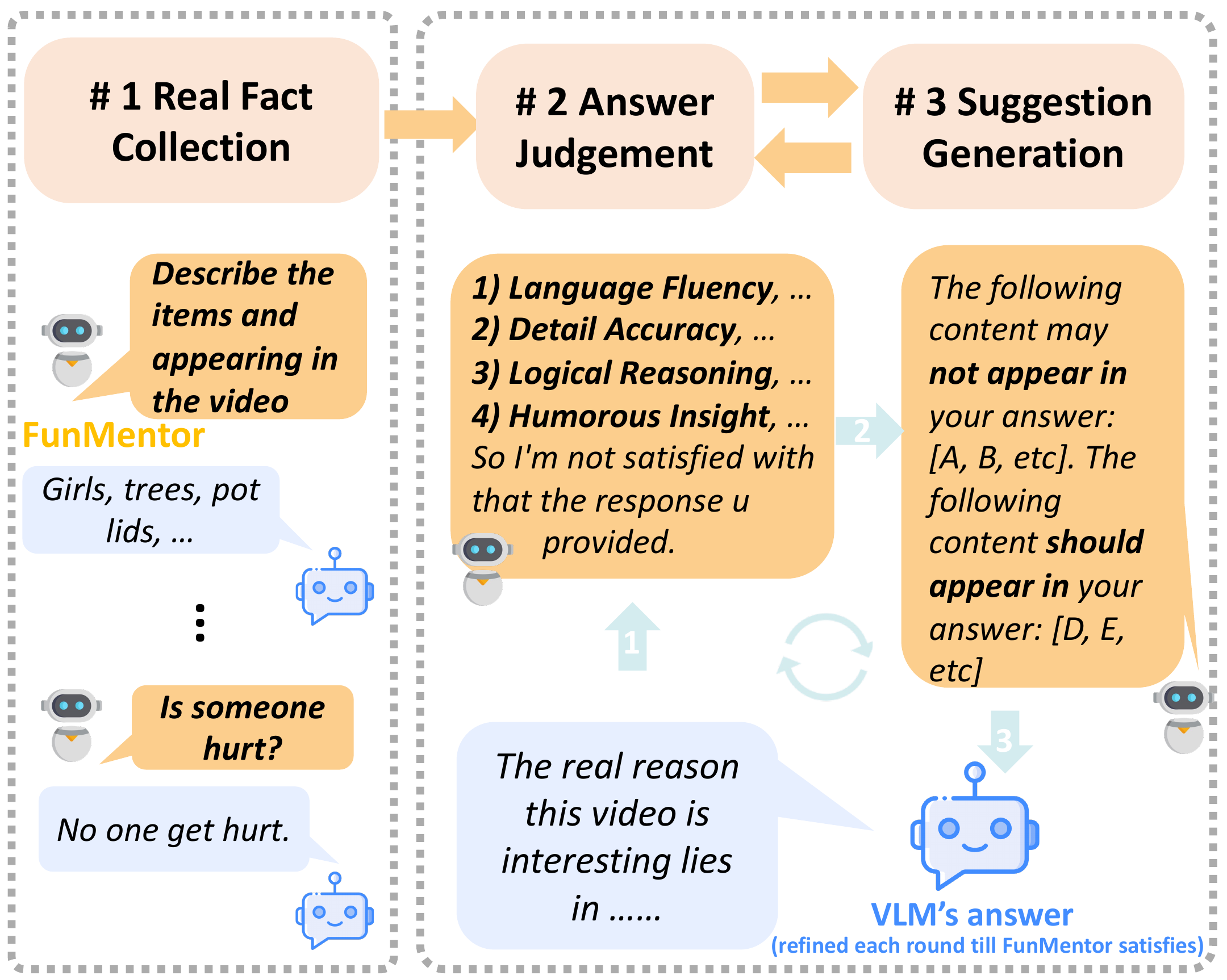}
    \vspace{-20pt}
    \caption{\textbf{FunMentor's Refining Process.} FunMentor asks multi-round questions to help VLMs to generate persuading answers.}
    \vspace{-30pt}
    \label{fig:funmentor}
    \vspace{10pt}
\end{wrapfigure}
This section presents the details of \textbf{FunMentor} for counter-intuitiveness understanding. FunMentor is an agent that refines a VLM's answer through multi-turn dialogues, ensuring it generates answers that best explain the given surprising content. The refining process comprises three components:\\
\noindent\textbf{Real Fact Collection}\quad 
Initially, FunMentor poses a series of inquiries to the VLM model it aims to assist, focusing on fact-aware questions to accurately comprehend the objective content of the video (e.g., FunMentor might ask: ``Please describe the items and characters appearing in each frame of the video''). This step is crucial because, without this preliminary context, FunMentor would have no knowledge of the video content and might be gullible to VLM's humorous explanations of the video. Thus, providing it with some factual clues is very important.\\
\noindent\textbf{Answer Judgement}\quad 
FunMentor assesses VLMs' responses based on several key aspects: \textbf{1) Language Fluency}, ensuring responses are grammatically fluent; \textbf{2) Detail Accuracy}, verifying factual correctness in relation to the video's content; \textbf{3) Logical Reasoning}, checking for coherent logic and smooth transitions; and \textbf{4) Humorous Insight}, expecting responses to provide humor beyond a mere description of the video. If not passed, FunMentor will try to generate some feedback, which is explained below.\\
\noindent\textbf{Suggestion Generation}\quad 
Upon receiving unsatisfactory responses, FunMentor formulates constructive suggestions and tailored prompts, which are then sent back to the VLMs for revision, anticipating their improved answers. This process involves the agent analyzing the model's initial response, blending pre-defined prompts with the context of the original question, to generate specific feedback instructions. For example, FunMentor examines the issues in the VLM's response and advises on what should and shouldn't be included in the revised answer, guiding the VLM towards a more accurate and relevant reply.\\
Fig. \ref{fig:funmentor} shows the pipeline of FunMentor. More details as shown in Appendix \ref{appendix_c: More Details of FunMentor}.

\section{Experiments}
In this section, we begin by introducing various caption-based and instruction-based models for evaluation. We then delve into diverse metrics tailored for FunQA tasks, with extensive details provided in Appendix \ref{appendix_d: Experiments}. Our comprehensive experiments and deep analysis of the results are then presented. Additionally, we conduct a comparative study contrasting FunQA with the previous VQA dataset, NExT-QA, which similarly features multiple-choice and open-ended questions. This comparison underscores the unique attributes and importance of FunQA.

\subsection{Model Zoo}
We categorize the models capable of addressing the majority of VQA tasks into two classes: Caption-based models and Instruction-based models. Caption-based models generate suitable captions for videos by taking different prompts as input. For this category, we evaluate mPLUG~\cite{li2022mplug} and GIT \cite{wang2022git}. Meanwhile, Instruction-based models are capable of answering a wide array of questions. For this category, we evaluate models VideoChat \cite{2023videochat}, Video-ChatGPT \cite{maaz2023videochatgpt}, mPLUG-Owl \cite{ye2023mplugowl}, Video-LLaMA \cite{damonlpsg2023videollama} and Otter \cite{li2023otter}.
For Otter, we evaluate two versions: one is fine-tuned on the Dense Caption \cite{krishna2017dense}, and another is fine-tuned on the FunQA training set. We evaluate the proposed FunMentor based on Video-ChatGPT and Otter.

\subsection{Evaluation Metrics}
\noindent\textbf{Timestamp Localization (H1, C1, M1)}\quad 
We employ the intersection of unions (IOU) based on time span. \\
\noindent\textbf{Description \& Reasoning (H2-4, C2-4, M2-3)}\quad
For all the \textbf{free-text} tasks, we employ two approaches for evaluation. Firstly, we utilize traditional NLG (Natural Language Generation) metrics. We use BLEU-4 \cite{papineni2002bleu}, ROUGE-L \cite{lin2004rouge}, CIDEr \cite{vedantam2015cider}, and BLEURT \cite{pu2021learning} as our metrics. The first two rely on N-gram overlap, which is only sensitive to lexical variations and cannot identify changes in sentence semantics or grammar. The latter two are reference-based evaluation metrics. Secondly, several works \cite{wang2023chatgpt, fu2023gptscore, kamalloo2023evaluating, chiang2023can} have shown promising results in utilizing GPT as a metric for NLG. Therefore, we introduce GPT-4 to assist in evaluating free-text similarity. We carefully design the prompts to make it possible to give objective ratings as much as possible like a human being. More details of GPT-4 prompts and evaluation criteria are provided in Appendix 
\ref{appendix_d: Evaluation Metrics}.\\
\noindent\textbf{Creative Scoring (C5)}\quad $Creative Score Metric = 100 \times \left(1 - \frac{{\lvert Predict - GT \rvert}}{{20}}\right)$.
\subsection{Results and Observations}
\begin{table}[t]
\caption{\textbf{Main Results on FunQA Benchmark.} The FunQA benchmark consists of four task categories. H1, C1, M1 represent the counter-intuitive timestamp localization task, where \textbf{IOU} is used as the metric. H2, C2, M2 represent the detailed video description task, and H3, C3, M3 represent reasoning around counter-intuitiveness. For the higher-level tasks, H4, C4 involve attributing a fitting and vivid title. The responses for all these tasks in free-text format. We utilize the \textbf{BLEURT} (B.) and \textbf{GPT-4} metrics for evaluation. The scores for the traditional metrics (BLEU-4 and CIDEr) are all close to zero. Here, we present only the BLEURT and GPT-4 scores, with the full results available in the Appendix \ref{appendix_d: Exp_result}). C5 represents scoring the video creativity, and the metric is the \textbf{Accuracy} between the predicted score and the official score. We tested the caption-based and instruction-based models. 
Here we clarify the abbreviation in the table.  \textbf{L.M.}: GIT\_LARGE\_MSRVTT; \textbf{L.V.}: GIT\_LARGE\_VATEX; \textbf{D.C.} means finetuned on Dense Caption; \textbf{FunQA} means finetuned on FunQA. }
\label{T:experiment}
\vspace{-10pt}
{\renewcommand\baselinestretch{1.5}\selectfont
\setlength\tabcolsep{10pt} 
\resizebox{1.0\linewidth}{!}{
\begin{tabular}{lp{1cm}<{\centering} cccp{1cm}<{\centering}ccccp{1cm}<{\centering}cc}
\toprule
\multirow{2}{*}{} & \multicolumn{4}{c}{HumorQA} & \multicolumn{5}{c}{CreativeQA} & \multicolumn{3}{c}{MagicQA} \\

\cmidrule(l){2-5} \cmidrule(l){6-10} \cmidrule(l){11-13}
Task & H1 & H2-Des. & H3-Rea. & H4-Title & C1 & C2-Des. & C3-Rea. & C4-Title & C5-Score & M1 & M2-Des. & M3-Rea. \\
\toprule
Metrics & IOU & B. / GPT-4 & B. / GPT-4 & B. / GPT-4 & IOU & B. / GPT-4 & B. / GPT-4 & B. / GPT-4 & Acc & IOU & B. / GPT-4 & B. / GPT-4 \\
\toprule
\rowcolor{C-Model}
\multicolumn{13}{l}{\textbf{- Caption-based Model}} \\
\rowcolor{C-Model}
mPLUG \cite{li2022mplug}  
& 0.0
& 19.9 / 3.9
& \textbf{25.7} / \textbf{6.0}
& 22.1 / \textbf{11.2}

& 0.0 
& 14.9 / 3.0
& \textbf{24.2} / \textbf{6.9}
& \textbf{20.8} / \textbf{18.8}
& 0.0 / 0.0 

& 0.0 
& 19.7 / 4.0
& \textbf{21.2} / \textbf{8.1} \\
\rowcolor{C-Model}
GIT (L.M.) \cite{wang2022git} 
& 0.0 
& 22.4 / 3.6
& 0.0 / 0.0
& 17.0 / 8.9

& 0.0 
& 14.4 / 3.8
& 0.0 / 0.0
& 7.1 / 11.3
& 0.0 / 0.0

& 0.0
& 19.4 / 8.2
& 0.0 / 0.0 \\

\rowcolor{C-Model}
GIT (L.V.) \cite{wang2022git} 
& 0.0 
& \textbf{33.3} / \textbf{4.0}
& 0.0 / 0.0
& \textbf{25.9} / 10.0

& 0.0 
& \textbf{20.5} / \textbf{4.2}
& 0.0 / 0.0
& 10.5 / 12.0
& 0.0 / 0.0 

& 0.0 
& \textbf{29.8} / \textbf{8.6}
& 0.0 / 0.0 \\
\midrule
\rowcolor{I-Model}
\multicolumn{13}{l}{\textbf{- Instruction-based Model}} \\
\rowcolor{I-Model}
VideoChat \cite{2023videochat} 
& 0.0 
& 44.0 / 17.9
& 45.4 / 31.9
& 20.2 / 31.7

& 0.0 
& 21.7 / 5.9
& 22.8 / \textbf{17.7}
& 7.3 / 31.1
& 67.5 

& 0.0 
& 47.4 / 8.2
& 43.1 / 44.6 \\

\rowcolor{I-Model}
Video-ChatGPT \cite{maaz2023videochatgpt} 
& 0.0 
& 39.9 / \textbf{24.3}
& 40.1 / 24.9
& 36.5 / 41.2

& 0.0 
& \textbf{45.8} / 6.6
& 45.2 / 9.1
& 30.9 / \textbf{48.8}
& \textbf{85.4} 

& 0.0 
& \textbf{50.8} / \textbf{11.2}
& 43.3 / 40.4 \\

\rowcolor{I-Model}
mPLUG-Owl \cite{ye2023mplugowl} 
& 0.0 
& 44.5 / 10,7
& \textbf{47.3} / \textbf{35.0}
& 29.8 / \textbf{48.8}

& 0.0 
& 43.0 / 5.0
& \textbf{44.7 }/ 10.6
& 23.9 / 36.3
& 66.7 

& 0.0 
& 46.4 / 8.6
& 43.9 / 30.9 \\

\rowcolor{I-Model}
Video-LLaMA \cite{damonlpsg2023videollama} 
& 0.0 
& \textbf{48.4} / 7.7
& 42.9 / 29.0
& 46.5 / 34.1

& 0.0 
& 45.5 / 7.2
& 41.1 / 17.2
& 42.3 / 31.2
& 64.2 

& 0.0 
& 50.1 / 10.2
& 39.0 / 28.0 \\

\rowcolor{I-Model}
Otter (D.C.) \cite{li2023otter} 
& 0.0 
& 30.2 / 7.7
& 32.3 / 28.3
& 21.7 / 20.0

& 0.0 
& 28.7 / 1.7
& 32.9 / 7.9
& 17.7 / 36.3
& 45.0

& 0.0 
& 32.5 / 2.1
& 27.3 / 36.8 \\
\rowcolor{I-Model}
Otter (FunQA) \cite{li2023otter} 
& 0.0 
& 38.4 / 8.9
& 42.6 / 31.7
& \textbf{47.5} / 32.1

& 0.0 
& 40.0 / \textbf{7.3}
& 41.1 / 8.8
& \textbf{44.5} / 38.8
& 69.4

& 0.0 
& 44.7 / 10.3
& \textbf{44.5} / \textbf{47.5} \\
\midrule
\rowcolor{O-Model}
Video-ChatGPT \cite{maaz2023videochatgpt}  + \textbf{FunMentor (Ours)}
& 0.0 
& \textbf{65.2} / \textbf{33.2}
& \textbf{57.5} / 36.5
& 50.2 / \textbf{65.1}

& 0.0 
& \textbf{66.3} / \textbf{14.2}
& \textbf{58.7} / \textbf{23.4}
& 45.3 / \textbf{52.2}
& \textbf{85.4}

& 0.0
& \textbf{55.1} / \textbf{13.3}
& \textbf{46.3} / 54.8 \\
\rowcolor{O-Model}
Otter (FunQA) \cite{li2023otter} + \textbf{FunMentor (Ours)}
& 0.0
& 33.4 / 13.4
& 37.8 / \textbf{45.8}
& \textbf{58.3} / 34.2

& 0.0 
& 60.4 / 11.0
& 44.4 / 9.3
& \textbf{53.9} / 43.5
& 69.4

& 0.0 
& 43.5 / 12.81
& 38.91 / \textbf{56.4}\\

\bottomrule

\end{tabular}}\par
}
\vspace{-0.5cm}
\end{table}

\begin{figure}[t]
    \centering
    \vspace{-10pt}
    \includegraphics[width=0.9\linewidth]{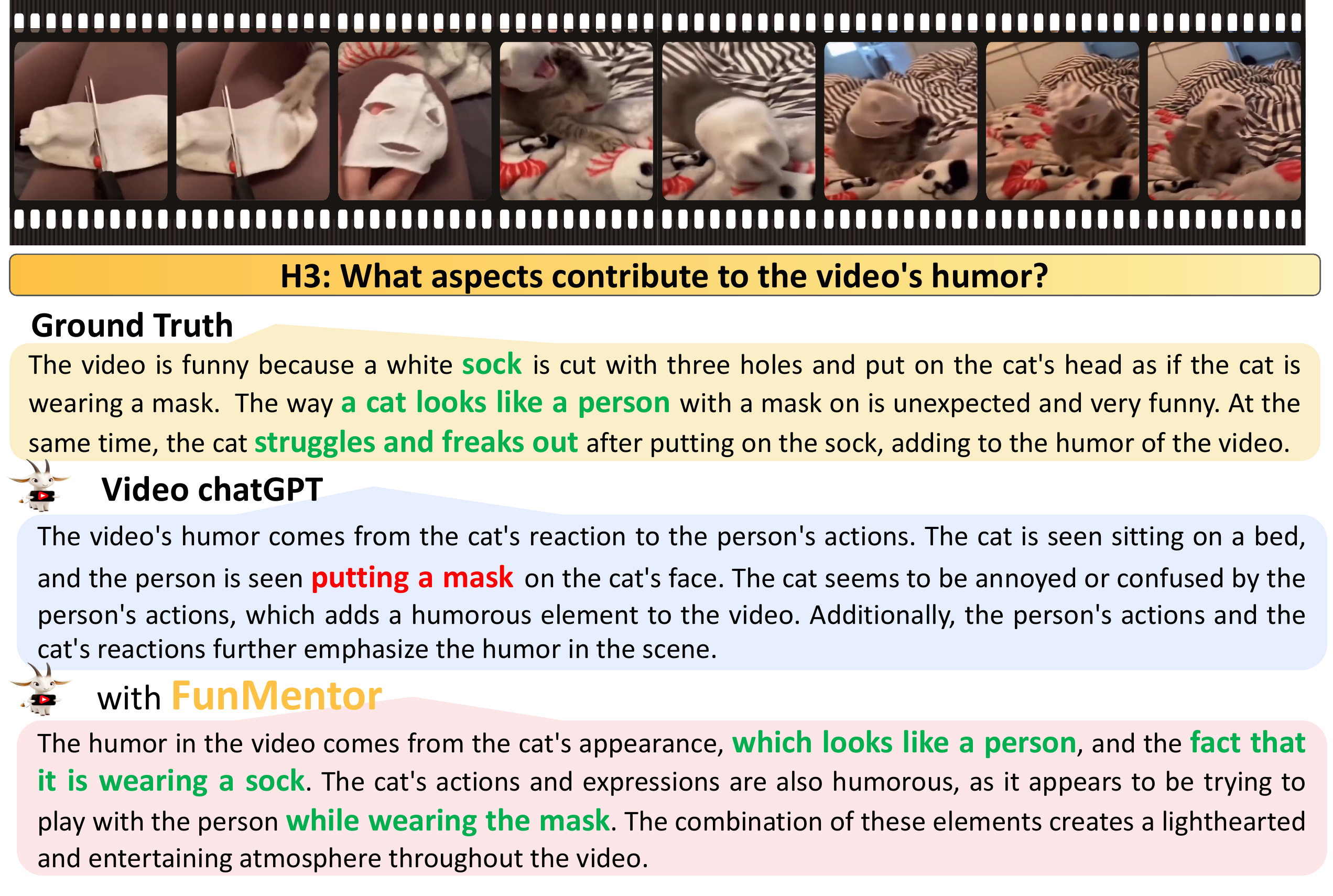}
    \vspace{-10pt}
    \caption{\textbf{VLM responses on before and after FunMentor.} Here shows the answers given by Video-ChatGPT~\cite{maaz2023videochatgpt} on HumorQA video before and after FunMentor. From the GroundTruth, it is evident that \textbf{three} key elements contribute to the humor in the video: \textbf{``sock''}, \textbf{``mask''}, and \textbf{``cat resemble a human''}. Initially, Video-ChatGPT only identified the mask, failing to grasp the full essence of the video's humor. However, after the combination with FunMentor, Video-ChatGPT successfully recognized the \textbf{``looks like a person''} and the \textbf{``fact that it is wearing a sock''}, thus demonstrating a true understanding of what makes the video funny.
    }
    \label{fig:vlm's response}
    \vspace{-25pt}
\end{figure}

In Table \ref{T:experiment}, we show the results of model zoo and
our proposed method. For clarity, we provided a list of dos and don’ts for comparing values in the table:\\ 
\textbf{a. Values from different tasks with the same video type (e.g., H2 and H3) are not comparable.} We observe that the model output in reasoning tasks may contain several words that match the ground truth (GT) while the meaning might be incorrect, resulting in inflated results. Therefore, our comparison is only based on qualitative analysis. \\
\textbf{b. Same metric with different models (see vertically) is comparable.}\\
\textbf{c. Same task for different video types (e.g., H2 and C2) is comparable.}\\
As an example, Fig. \ref{fig:vlm's response} illustrates the responses of VLM before and after applying our method to the HumorQA dataset. Overall, the performance of the models on the FunQA dataset is generally unsatisfactory. However, after fine-tuning VLM with FunMentor, a notable improvement is shown in its ability to comprehend counter-intuitiveness. We have made several key findings:\\
\noindent\textbf{Timestamp localization task is the most challenging.}\quad
Caption-based models focus mainly on captioning and often omit temporal information. 
\begin{wraptable}{r}{5.0cm}
\centering
\vspace{-1cm}
\small
\caption{\textbf{Timestamp Localization Task's baseline.} The result uses IOU metrics.}
\resizebox{\linewidth}{!}{
\begin{tabular}{lp{2cm}<{\centering}p{2cm}<{\centering}p{2cm}<{\centering}}
\toprule
Task         & H1                              & C1                             & M1                             \\ \midrule
Random   &  29.4 & 32.6 &  21.1   \\
Mean  &  55.5  &  82.1 &  46.3 \\
\midrule
TimeChat \cite{Ren2023TimeChat} &  8.4  &  14.6 &  6.0   \\
\bottomrule 
\end{tabular}}
\label{tab:iou}
\caption{\textbf{Human Performance on HumorQA.} The average scores using BLEURT / GPT-4. Full results are in Appendix \ref{appendix_d: Experiments}.}
\resizebox{\linewidth}{!}{
\begin{tabular}{lccc}
\toprule
Task         & H2                              & H3                             & H4                              \\ \midrule
Model SOTA   &   65.2 / 33.2  &   57.5 / 45.8 &  58.3 / 65.1   \\
Human &  70.0 / \textbf{83.6} &  63.1 / \textbf{84.4} &  67.3 / \textbf{79.1}\\
\bottomrule 
\end{tabular}}
\vspace{-1cm}
\label{tab:human}
\end{wraptable}
Due to this, they are not equipped to deal with our task that emphasizes temporal aspects, so they are not scored. Similarly, Instruction-based models, such as Otter, take visual information from specific frames without temporal context. Their outputs are thus confined to individual frames, making them ineffective at addressing temporal localization. In summary, none of the current VLMs could solve H1, C1, or M1 tasks since they do not have a sense of time (refer to Appendix \ref{appendix_d: Failure of t1}). To enhance the timestamp localization task analysis, we add comparative baselines with Random Guess; Mean Time Span derived from the training dataset for an average reference; and TimeChat \cite{Ren2023TimeChat}, a specialized time-sensitive multimodal language model. As shown in Table \ref{tab:iou}, TimeChat’s score is the lowest showing the challenges of the Timestamp Localization Task.\\
\noindent\textbf{No clear winner across all tasks.}\quad
Caption-based models excel in providing detailed descriptions but struggle in tasks that require reasoning, resulting in a notable performance gap between description tasks (e.g., H2) and reasoning tasks (e.g., H3). On the other hand, instruction-based models demonstrate stronger reasoning capabilities than caption-based models but tend to underperform in description tasks. 
One possible explanation is that instruction-based models may generate excessive information in their answers, including a significant amount of incorrect information. We conducted experiments to compare machine versus human performance on FunQA as shown in Table \ref{tab:human}.\\
\noindent\textbf{Performance varies greatly across different video types.}\quad
Generally CreativeQA is the most challenging especially in reasoning. For instance, the GPT-4 scores of Video-ChatGPT and Otter on C3-Reasoning are notably low. One possible reason is that humor and magic videos often depict daily life that models have encountered previously, whereas creative videos that models have never seen before, causing them unable to understand and generate reasonable answers.\\
\noindent\textbf{Insufficient evaluation metrics for free-text tasks.}\quad
Traditional metrics yield near-zero scores on free-text questions (refer to the complete FunQA Benchmark results in the Appendix \ref{appendix_d: Experiments}), as they solely focus on short textual similarity. While BLEURT scores are significantly higher, they still fall short in evaluating more complex similarities. Intuitively, GPT-4 is found to show preliminary capabilities in assessing free-text in deep understanding, which will be detailed in Appendix \ref{appendix_d: Evaluation Metrics}. However, there are still issues of instability, where the same content can receive different scores.\\
\noindent\textbf{Finetuned Otter performs well yet with limitations.}\quad
We finetuned Otter on Dense Caption and FunQA, and Otter (FunQA) shows obvious performance advantages over Otter (D.C.). As shown in Table \ref{T:experiment}, Otter (FunQA) performs better in BLEURT and GPT-4. However, there are still some limitations that Otter (FunQA) falls short of reaching SOTA scores. One possible reason revealed is that the input of Otter is only 128 frames sampled from the video, which is insufficient for comprehensive reasoning.\\
\noindent\textbf{FunMentor demonstrates the powerful potential of agent-based finetuning}\quad Our proposed FunMentor outperforms the previous best method by 10.8 for the GPT-4 score of H3 task, with improvements of 4.8 and 20.9 for Otter (FunQA) and Video-ChatGPT, respectively. Additionally, the results reveal that FunMentor achieves significant performance improvements for Video-ChatGPT, particularly in the H2 and H4 tasks, while the improvement for H3 is relatively modest. This indicates that counter-intuitive reasoning remains a challenging aspect. The substantial performance enhancement by FunMentor highlights the promising prospects of agent-based fine-tuning methods. In the context of VLM requiring extensive training data, research in this direction holds the potential to uncover the vast capabilities of VLM.

\subsection{Comparison with Previous Benchmarks}
\label{sec:compare_nextqa}
We chose NExT-QA as our comparative benchmark, which is designed to emphasize model reasoning abilities. NExT-QA also provides a multi-choice version of the dataset (NExT-OE), similar to FunQA-MC.
\begin{table}[h]
\vspace{-20pt}
\caption{\textbf{Model's Performance in NExT-OE.} This table illustrates that while the previous SOTA model, HGA, excels in the classic WUPS metric, it underperforms with GPT-4. See Appendix~\ref{appendix_d: Evaluation Metrics} for a discussion on potential issues with WUPS.}
\vspace{-10pt}
\label{T:vlm_nextoe}
\centering
\setlength\tabcolsep{5pt}
\resizebox{1.0\linewidth}{!}{
\begin{tabular}{lp{5cm}<{\centering}p{5cm}<{\centering}}
\toprule
Metrics & WUPS & GPT-4  \\
\hline
HGA (SOTA on NExT-OE)& 25.18	& 25.06 \\
Otter (D.C.) & 1.26 & 64.89 \\
Otter (FunQA) & 0.79 & \textbf{73.06} \\
\bottomrule
\end{tabular}
}
\vspace{-20pt}
\end{table}

\noindent\textbf{Traditional metrics are ineffective on VLMs' response.}\quad
Table \ref{T:vlm_nextoe} shows the performance on NExT-OE of Otter and HGA \cite{jiang2020reasoning} using different metrics. We evaluated these models using both traditional evaluation metrics (specifically, WUPS \cite{malinowski2014multi} that is employed by NExT-OE) and our novel evaluation metrics based on GPT-4. Otter exhibits a notably low performance on the WUPS metric that is drastically different than the GPT-4 score, primarily because WUPS is ill-suited for evaluating sentence-based responses and fares poorly when assessing phrases. Specific examples can be found in Appendix \ref{appendix_d: Exp_result}. 

\noindent\textbf{NExT-QA is not a challenge for GPT-4.}\quad
Our investigation into VLMs' performance on NExT-QA and FunQA datasets, shown in Table \ref{T:vlm_next_funqa}, reveals key insights. In the multi-choice format, GPT-4, even without video frames, scores comparably to VLMs on NExT-QA but resembles random guessing on FunQA-MC. Furthermore, while GPT-4V scores 80 and 61 on NExT-QA and FunQA-MC respectively, it drops to 39 on FunQA-MC-R, which focuses on counter-intuitive reasoning. This suggests that NExT-QA's inferential questions are no longer challenging for GPT-4, and the significant score difference in FunQA variants underlines GPT-4V's struggle with complex, non-intuitive questions.

\noindent\textbf{FunQA is challenging for VLMs and GPT-4V.}\quad
As Table~\ref{T:vlm_next_funqa} indicates, GPT-4 scores well on NExT-OE even without video access. The contrast in its performance on NExT-OE versus FunQA highlights FunQA's complexity in video reasoning. A significant performance disparity between GPT-4 and its video-enhanced version, GPT-4V, on FunQA again underscores the value of video content, corroborating findings from Table \ref{T:onlyframes}. Overall, FunQA stands out from prior benchmarks with its focus on reasoning skills and high-quality QA pairs deeply linked to video content, establishing itself as a robust benchmark in the LLM era for assessing VLMs' capabilities in counter-intuitive reasoning.

\begin{table}[t]
\vspace{-0.05mm}
\caption{\textbf{Performance of VLMs in NExT-QA and FunQA.} N.QA, F.MC, and F.MC-R denote NExT-QA, FunQA-MC and FunQA-MC-R, which are multi-choice datasets and we use \textbf{accuracy} as the metric. Regarding N.QE (NExT-OE) and FunQA, which are open-ended answer datasets, we employ metrics based on \textbf{GPT-4}. $^{\ast}$ indicates that this model utilizes a Chat-box mechanism, and we employ GPT-3.5 to automatically convert its output into multiple-choice answers; $^{\dagger}$ signifies the GPT-4 model without any visual information input; $^{\ddagger}$ denotes the GPT-4V(ision) model, where an average sample of four frames of images is used as input. $^{\mathsection}$ denotes an sample version ($10\%$) of all free-text QA pairs.}
\label{T:vlm_next_funqa}
\vspace{-10pt}
\centering
\setlength\tabcolsep{10pt}
\resizebox{1.0\linewidth}{!}{
\begin{tabular}{lcccp{1cm}<{\centering}cc}
\toprule
Metrics & \multicolumn{3}{c}{Acc} & \multicolumn{2}{c}{GPT-4} \\
\cmidrule{2-4}\cmidrule{5-6}
Dataset & N.QA & F.MC & F.MC-R & N.QE & FunQA $^{\mathsection}$ \\
\midrule
Random & 20 & 20 & 20 & - & - \\
Otter (D.C.) $^{\ast}$ & 35 & 27 & 17 & 54 & 22 \\
Otter (FunQA) $^{\ast}$ & 42 & 31 & 26 & 58 & \textbf{28} \\
GPT-4 $^{\dagger}$ & 44 & 34 & 23 & 30 & 2 \\
GPT-4V $^{\ddagger}$ & \textbf{80} & \textbf{61} & \textbf{39} & \textbf{79} & 5\\
\bottomrule
\end{tabular}
}
\vspace{-20pt}
\end{table}




\section{Limitations and Future Work}
This paper has two limitations. \textbf{1)} The current FunQA dataset primarily contains video-level data and annotations. There is potential for enhanced video reasoning through denser annotations, akin to PVSG~\cite{yang2023panoptic}, which might include detailed spatial, temporal, and object-level annotations. \textbf{2)} The initial annotations were made in Chinese and later translated into English. While GPT was used to refine and complete the Chinese text, ensuring comprehensiveness and correctness, differences due to cultural differences between the languages might still persist.\\
Looking ahead, we plan to enrich FunQA with more detailed and varied annotations. We aim to develop new metrics for a more accurate evaluation of models, particularly for open-ended questions. Our goal is to steer models towards deeper video reasoning. However, to ensure fair comparisons and prevent data leakage, it is advised that future research does not utilize the FunQA testing set.
\section{Acknowledgments and Disclosure of Funding}
This study is supported by the Ministry of Education, Singapore, under its MOE AcRF Tier 2 (MOE-T2EP20221- 0012), NTU NAP, and under the RIE2020 Industry Alignment Fund – Industry Collaboration Projects (IAF-ICP) Funding Initiative, as well as cash and in-kind contribution from the industry partner(s).


%
%
\bibliographystyle{splncs04}
\bibliography{main}
\newpage

\appendix
\setcounter{table}{0}
\renewcommand{\thetable}{A\arabic{table}}
\setcounter{figure}{0}
\renewcommand{\thefigure}{A\arabic{figure}}

\section{More Details of FunQA Dataset} \label{appendix_a: FunQA more details}
\subsection{Dataset Construction Pipeline} \label{appendix_a:dataset con}
\noindent\textbf{Video Selection}\quad
In constructing the dataset, we adhered to three principles to address the challenges in video understanding capabilities: our dataset, FunQA, is \textbf{visual centered} and emphasizes \textbf{counter-intuitive reasoning}, \textbf{spatial-temporal reasoning}. Based on these principles, we collect 4365 videos from 3 different art genres and created three subsets: HumorQA, CreativeQA, and MagicQA.\\
\noindent\textbf{HumorQA}\quad
HumorQA composed of 1,769 meticulously curated web videos, serves as a unique source of insight into human humor comprehension. Notably, it contains the shortest average video length of 7s among the three subsets. We believe that the human process of understanding humor is complex and deep, requiring a holistic understanding of the video and adding a degree of common sense to it. Psychological research has demonstrated that humor arises from the incongruity \cite{kant1987critique,sep-humor} between reality and expectations, flourishing with the skillful juxtaposition and transformation of events \cite{latta1999basic,boyd2004laughter,koestler2020act}. This makes humorous videos a valuable asset for the VideoQA dataset, anticipated to enhance a model's proficiency in integrating information and performing deep reasoning.\\
\noindent\textbf{CreativeQA}\quad 
CreativeQA is a collection of 927 videos averaging 48s in length from a TV show called Kasou Taishou \cite{ntv}. This program, showcasing original and novel skits performed by various amateur groups and judged by a panel, boasts a strong creative flair 
 \cite{runco2012standard}. The essence of the show lies in using a mix of people and props to mimic reality, with audiences deriving pleasure from information integration and comparison. We anticipate that the imitation nature of the show will challenge the model's capacity for information extraction, while the longer video length and need for understanding creativity will put to test the model's comprehension of spatial-temporal information.\\
\noindent\textbf{MagicQA}\quad
MagicQA encapsulates 1672 magic performance videos sourced from across the web, spanning various genres like camera magic, close-up magic, and stage magic. The essence of magic revolves around the creation of seemingly impossible illusions \cite{nelms2012magic}, employing diverse effects such as disappearance, creation, and transformation. These illusions are infused with abundant spatial-temporal information. Through this dataset, we aim to empower the model to not only track the ensuing changes in objects but also unravel the underlying mechanics \cite{lamont2005magic} of these transformations.

\begin{table*}[!t]
\centering
\caption{\textbf{Comparison between FunQA and other existing benchmarks (Complete Version).}
Compared to other datasets, FunQA revolves around the captivating realm of interesting and counter-intuitive videos. The tasks within FunQA are specifically designed to challenge the vision capabilities of models, requiring strong skills in producing an in-depth description, interpretation, and spatial-temporal reasoning.
Here we clarify the abbreviation in the table. For annotation type:  \includegraphics[width=0.02\textwidth]{figures/human.png} denotes Manual Annotation and  \includegraphics[width=0.02\textwidth]{figures/robot.png} for Automatic Annotation; \textbf{Avg Len} denotes video average length; \textbf{\# Clips} means number of video clips; \textbf{VC} for visual-centric, \textbf{Des.} for Description, \textbf{Exp.} for Explanation, \textbf{STR} for Spatial-temporal Reasoning, \textbf{MC} means Multiple Choice QA, and \textbf{OE} shows Open Ended QA with \textbf{Average Word Count} per response. 
}
\vspace{-5pt}
\label{T:whole_main}
\setlength\tabcolsep{10pt}
\resizebox{1.0\linewidth}{!}{
\begin{tabular}{llccccccccccl}
\toprule
\multirow{2}{*}{Dataset} & \multirow{2}{*}{Domain} & \multirow{2}{*}{
\includegraphics[width=0.02\textwidth]{figures/human.png} or \includegraphics[width=0.02\textwidth]{figures/robot.png}
} & \multicolumn{2}{c}{Video} & \multicolumn{7}{c}{Question Answer}  \\
\cmidrule(l){4-5} \cmidrule(l){6-12} 
 &  &  & Avg Len & \# Clips                      & \# QA  & VC & Des. & Exp. & STR & MC & OE   \\
 \toprule
TGIF-QA \cite{jang-IJCV-2019} 
& Social Media & \includegraphics[width=0.02\textwidth]{figures/robot.png} & 3s & 72K & 165K & \textcolor{darkgreen}{\ding{51}} & \textcolor{darkgreen}{\ding{51}} & \textcolor{darkred}{\ding{55}} & \textcolor{darkgreen}{\ding{51}} & \textcolor{darkred}{\ding{55}} & 2.1 \\

MSRVTT-QA \cite{xu2017video}
& Social Media & \includegraphics[width=0.02\textwidth]{figures/robot.png} & 15s & 10K & 244K & \textcolor{darkgreen}{\ding{51}} & \textcolor{darkred}{\ding{55}} & \textcolor{darkred}{\ding{55}} & \textcolor{darkgreen}{\ding{51}} & \textcolor{darkred}{\ding{55}} & 1.0 \\

ActivityNet-QA \cite{yu2019activitynetqa}
& Social Media & \includegraphics[width=0.02\textwidth]{figures/human.png} & 180s & 6K & 58K & \textcolor{darkgreen}{\ding{51}} & \textcolor{darkred}{\ding{55}} & \textcolor{darkred}{\ding{55}} & \textcolor{darkgreen}{\ding{51}} & \textcolor{darkred}{\ding{55}} & 1.9 \\

MSVD-QA \cite{xu2017video}
& Social Media & \includegraphics[width=0.02\textwidth]{figures/robot.png} & 10s & 2K & 51K & \textcolor{darkgreen}{\ding{51}} & \textcolor{darkred}{\ding{55}} & \textcolor{darkred}{\ding{55}} & \textcolor{darkred}{\ding{55}} & \textcolor{darkred}{\ding{55}} & 1.0\\

YouTube2Text-QA \cite{Ye_2017}
& Social Media & \includegraphics[width=0.02\textwidth]{figures/robot.png} & 10s & 10K & 123K & \textcolor{darkgreen}{\ding{51}} & \textcolor{darkred}{\ding{55}} & \textcolor{darkred}{\ding{55}} & \textcolor{darkgreen}{\ding{51}} & \textcolor{darkgreen}{\ding{51}} & N/A \\

AGQA \cite{grundemclaughlin2021agqa}                              
& Social Media & \includegraphics[width=0.02\textwidth]{figures/robot.png} & 30K & 10s & 192K & \textcolor{darkgreen}{\ding{51}} & \textcolor{darkred}{\ding{55}} & \textcolor{darkgreen}{\ding{51}} & \textcolor{darkgreen}{\ding{51}} & \textcolor{darkred}{\ding{55}} & TBD \\

AVQA \cite{yang2022avqa}                                          
& Social Media & \includegraphics[width=0.02\textwidth]{figures/human.png} & 60s & 9K & 57K & \textcolor{darkgreen}{\ding{51}} & \textcolor{darkred}{\ding{55}} & \textcolor{darkgreen}{\ding{51}} & \textcolor{darkgreen}{\ding{51}} & \textcolor{darkgreen}{\ding{51}} & N/A \\

\toprule

NExT-QA \cite{xiao2021nextqanext}            
& Daily life & \includegraphics[width=0.02\textwidth]{figures/human.png} & 44s & 5K & 52K & \textcolor{darkgreen}{\ding{51}} & \textcolor{darkgreen}{\ding{51}} & \textcolor{darkgreen}{\ding{51}} & \textcolor{darkgreen}{\ding{51}} & \textcolor{darkgreen}{\ding{51}} & 2.6 \\

Social-IQ \cite{zadeh2019social}                        
& Daily life & \includegraphics[width=0.02\textwidth]{figures/human.png} & 99s & 1K & 8K & \textcolor{darkgreen}{\ding{51}} & \textcolor{darkred}{\ding{55}} & \textcolor{darkgreen}{\ding{51}} & \textcolor{darkred}{\ding{55}} & \textcolor{darkgreen}{\ding{51}} & N/A \\

STAR \cite{wu2021star}                                            
& Daily life & \includegraphics[width=0.02\textwidth]{figures/robot.png} & - & 23K & 60K & \textcolor{darkgreen}{\ding{51}} & \textcolor{darkred}{\ding{55}} & \textcolor{darkred}{\ding{55}} & \textcolor{darkgreen}{\ding{51}} & \textcolor{darkgreen}{\ding{51}} & N/A  \\

FIBER \cite{castro2022fiber}                                      
& Daily life & \includegraphics[width=0.02\textwidth]{figures/human.png} & 10s & 28K & 2K & \textcolor{darkred}{\ding{55}} & \textcolor{darkgreen}{\ding{51}} & \textcolor{darkred}{\ding{55}} & \textcolor{darkgreen}{\ding{51}} & \textcolor{darkred}{\ding{55}} & TBD \\

\toprule

MovieQA \cite{tapaswi2016movieqa}                       & TV shows & \includegraphics[width=0.02\textwidth]{figures/robot.png} & 203s & 7K & 6K & \textcolor{darkred}{\ding{55}} & \textcolor{darkred}{\ding{55}} & \textcolor{darkgreen}{\ding{51}} & \textcolor{darkgreen}{\ding{51}} & \textcolor{darkgreen}{\ding{51}} & N/A \\

TVQA \cite{lei2018tvqa}    
& TV shows & \includegraphics[width=0.02\textwidth]{figures/human.png} & 76s & 22K & 153K & \textcolor{darkred}{\ding{55}} & \textcolor{darkred}{\ding{55}} & \textcolor{darkgreen}{\ding{51}} & \textcolor{darkred}{\ding{55}} & \textcolor{darkgreen}{\ding{51}} & N/A \\

TVQA+ \cite{lei2019tvqa}  
& TV shows & \includegraphics[width=0.02\textwidth]{figures/human.png} & 8s & 4K & 30K & \textcolor{darkred}{\ding{55}} & \textcolor{darkred}{\ding{55}} & \textcolor{darkgreen}{\ding{51}} & \textcolor{darkgreen}{\ding{51}} & \textcolor{darkgreen}{\ding{51}} & N/A \\

KnowIT-VQA \cite{garcia2020knowit}                                 
& TV shows & \includegraphics[width=0.02\textwidth]{figures/human.png} & 60s & 12K & 24K & \textcolor{darkgreen}{\ding{51}} & \textcolor{darkred}{\ding{55}} & \textcolor{darkgreen}{\ding{51}} & \textcolor{darkgreen}{\ding{51}} & \textcolor{darkgreen}{\ding{51}} & N/A \\

\toprule

SUTD-TrafficQA \cite{xu2021sutd}
& Traffic & \includegraphics[width=0.02\textwidth]{figures/human.png} & 5s & 10K & 623K & \textcolor{darkgreen}{\ding{51}} & \textcolor{darkred}{\ding{55}} & \textcolor{darkred}{\ding{55}} & \textcolor{darkgreen}{\ding{51}} & \textcolor{darkgreen}{\ding{51}} & N/A \\
MarioQA \cite{mun2017marioqa} 
& Games  & \includegraphics[width=0.02\textwidth]{figures/human.png} & 5s & 188K & 188K & \textcolor{darkgreen}{\ding{51}} & \textcolor{darkred}{\ding{55}} & \textcolor{darkgreen}{\ding{51}} & \textcolor{darkgreen}{\ding{51}} & \textcolor{darkred}{\ding{55}} & 2.0 \\

CLEVRER \cite{yi2020clevrer}     
& Synthetic Videos & \includegraphics[width=0.02\textwidth]{figures/human.png} & 5s & 20K & 305K & \textcolor{darkgreen}{\ding{51}} & \textcolor{darkred}{\ding{55}} & \textcolor{darkgreen}{\ding{51}} & \textcolor{darkgreen}{\ding{51}} & \textcolor{darkgreen}{\ding{51}} & N/A \\

Env-QA \cite{Gao_2021_ICCV}                                       
& Egocentric & \includegraphics[width=0.02\textwidth]{figures/human.png} & 20s & 23K & 85K & \textcolor{darkgreen}{\ding{51}} & \textcolor{darkred}{\ding{55}} & \textcolor{darkred}{\ding{55}} & \textcolor{darkgreen}{\ding{51}} & \textcolor{darkgreen}{\ding{51}} & N/A \\


\midrule
\textbf{HumorQA (Ours)     }                                     
& Daily life & \includegraphics[width=0.02\textwidth]{figures/human.png} & 7s & 2K & 141K & \textcolor{darkgreen}{\ding{51}} & \textcolor{darkgreen}{\ding{51}} & \textcolor{darkgreen}{\ding{51}} & \textcolor{darkgreen}{\ding{51}} & \textcolor{darkgreen}{\ding{51}} & \textbf{28.2}\\

\textbf{CreativeQA (Ours)   }                               
& Performance & \includegraphics[width=0.02\textwidth]{figures/human.png} & 48s & 1K & 79K & \textcolor{darkgreen}{\ding{51}} & \textcolor{darkgreen}{\ding{51}} & \textcolor{darkgreen}{\ding{51}} & \textcolor{darkgreen}{\ding{51}} & \textcolor{darkgreen}{\ding{51}} & \textbf{59.1}\\

\textbf{MagicQA (Ours)  }                              
& Magic shows & \includegraphics[width=0.02\textwidth]{figures/human.png} & 10s & 2K & 92K & \textcolor{darkgreen}{\ding{51}} & \textcolor{darkgreen}{\ding{51}} & \textcolor{darkgreen}{\ding{51}} & \textcolor{darkgreen}{\ding{51}} & \textcolor{darkgreen}{\ding{51}} & \textbf{27.6}\\

\textbf{FunQA (Ours)} 
& \textbf{Surprising Videos} 
& \includegraphics[width=0.02\textwidth]{figures/human.png} & 19s & 4K & \textbf{312K} & \textcolor{darkgreen}{\ding{51}} & \textcolor{darkgreen}{\ding{51}} & \textcolor{darkgreen}{\ding{51}} & \textcolor{darkgreen}{\ding{51}} & \textcolor{darkgreen}{\ding{51}} & \textbf{34.2} \\ 
\bottomrule 

\end{tabular}}
\end{table*}

\begin{figure}[h]
    \centering
    \includegraphics[width=\linewidth]{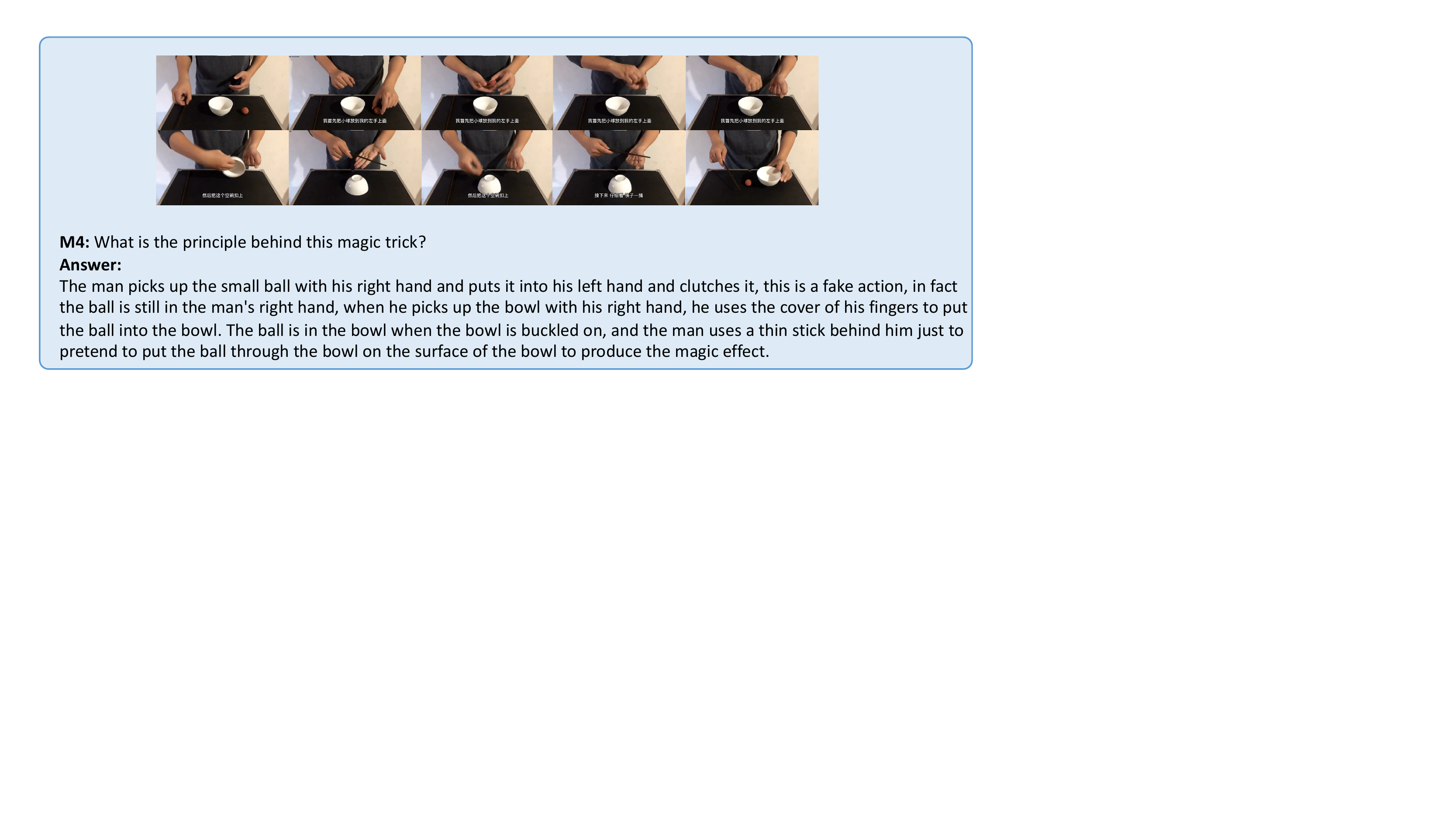}
    \vspace{-0.6cm}
    \caption{\textbf{Example of Magic Method Task.} During the annotation process, we discovered that even as humans, it is difficult to fully understand the complete principles behind the implementation of magic tricks right from the beginning.}
    \vspace{-0.4cm}
    \label{fig:M4}
\end{figure}

\begin{figure}[h]
    \centering
    \includegraphics[width=\linewidth]{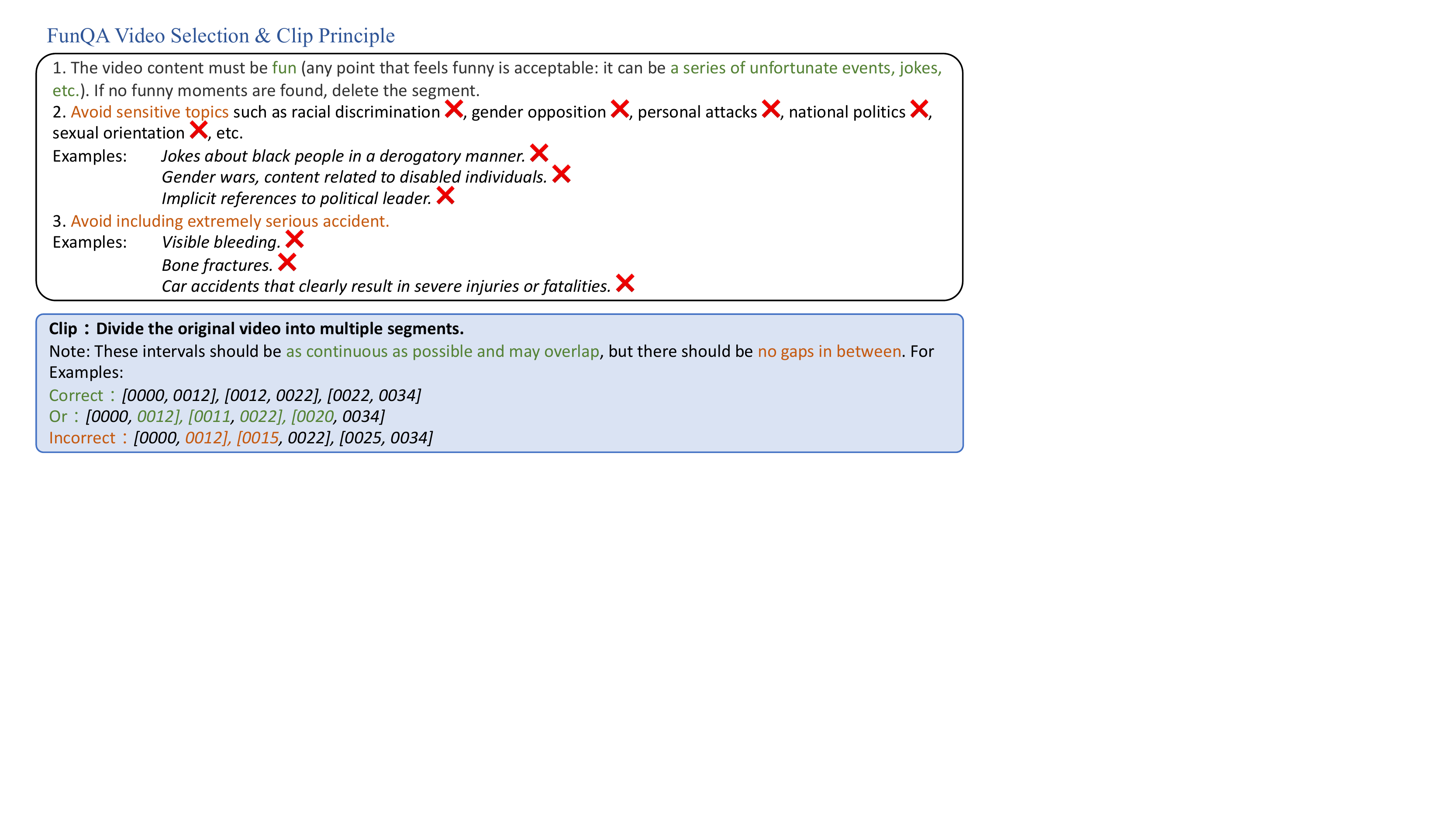}
    \vspace{-0.6cm}
    \caption{\textbf{FunQA Video Selection and Clip Principle.} We have a zero-tolerance policy regarding the inclusion of offensive content in our dataset. During the video sourcing process (video selection and video clipping stages), we ensure that such content is completely eliminated.}
    \vspace{-0.4cm}
    \label{fig:annotation-principle}
\end{figure}

\begin{figure*}[h]
    \centering
    \includegraphics[width=\linewidth]{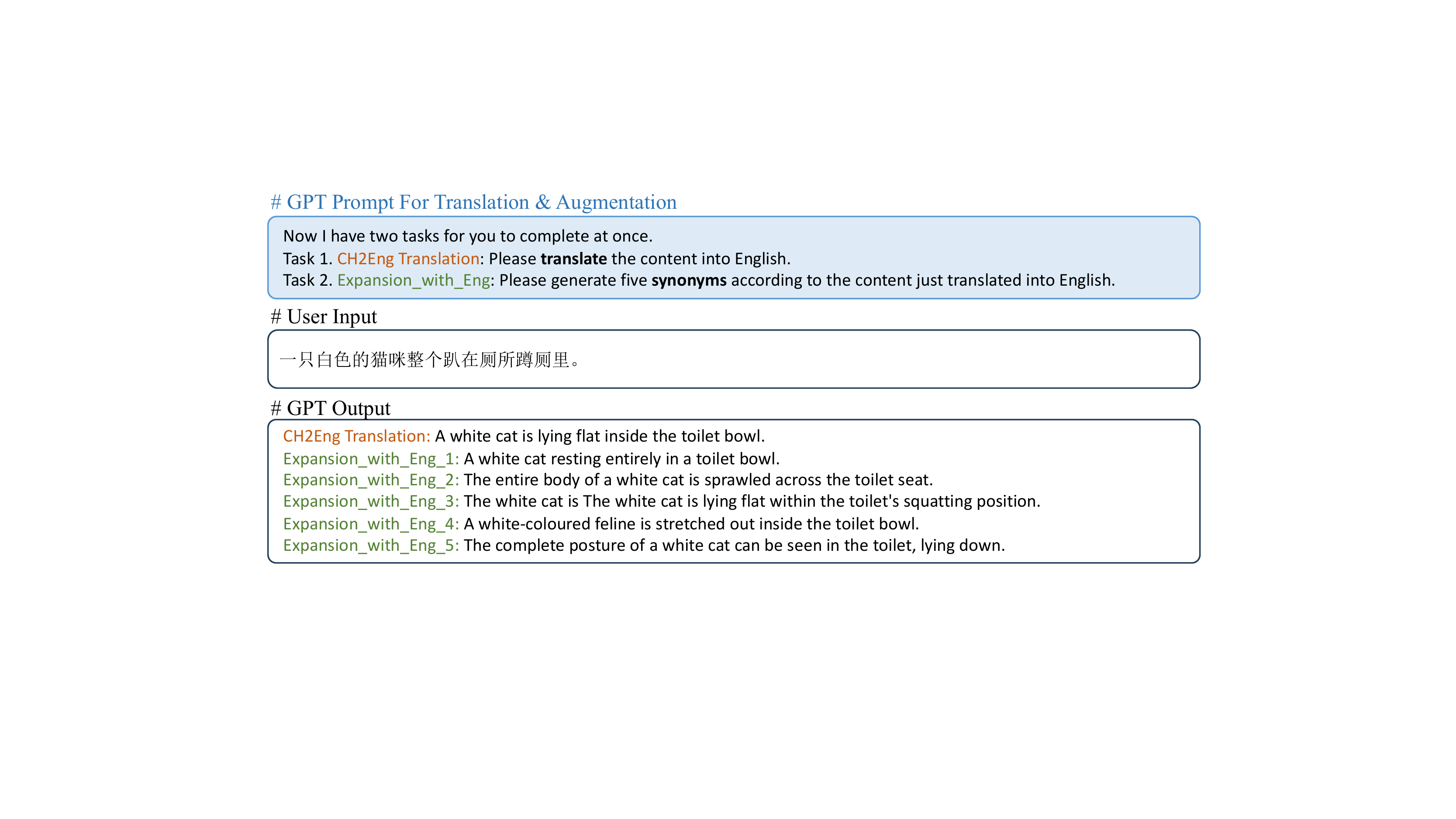}
    \vspace{-0.6cm}
    \caption{\textbf{Examples of incorrect annotations and the suggested modifications.} Here we clarify the meanings of different cell colors. \textbf{Red}: Errors made by the annotator during the annotation process. \textbf{Yellow}: Types of annotation errors. \textbf{Green}: Reasons for the errors and suggested corrections.}
    \vspace{-0.4cm}
    \label{fig:annotation-error}
\end{figure*}

\begin{figure}[ht]
    \centering
    \includegraphics[width=\linewidth]{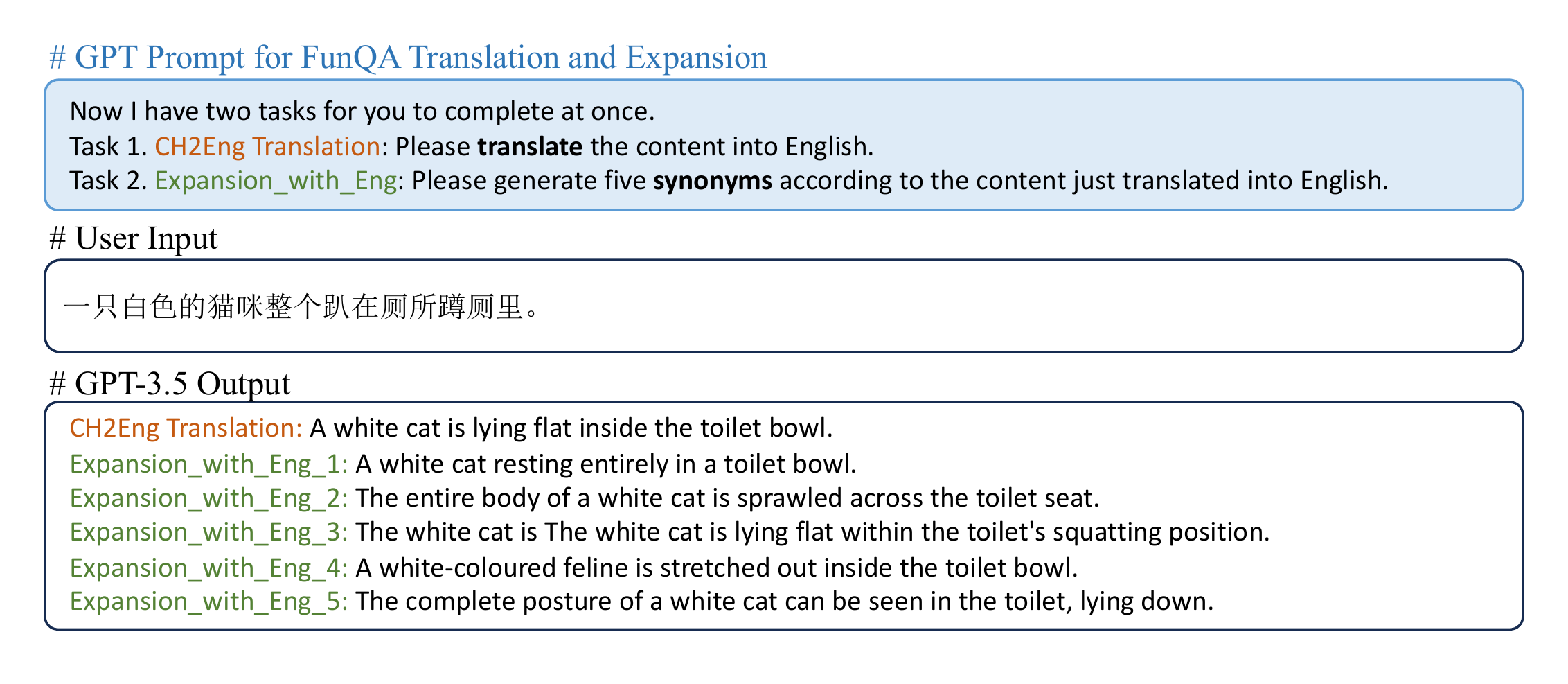}
    \vspace{-0.6cm}
    \caption{\textbf{GPT prompt for Translation and augmentation.} Under our carefully designed prompt, GPT-3.5 can automatically generate English translations and five synonymous sentences from our high-quality Chinese annotations, thereby expanding FunQA dataset.}
    \vspace{-0.4cm}
    \label{fig:prompt-trans}
\end{figure}
\begin{figure}[h]
    \centering
    \includegraphics[width=\linewidth]{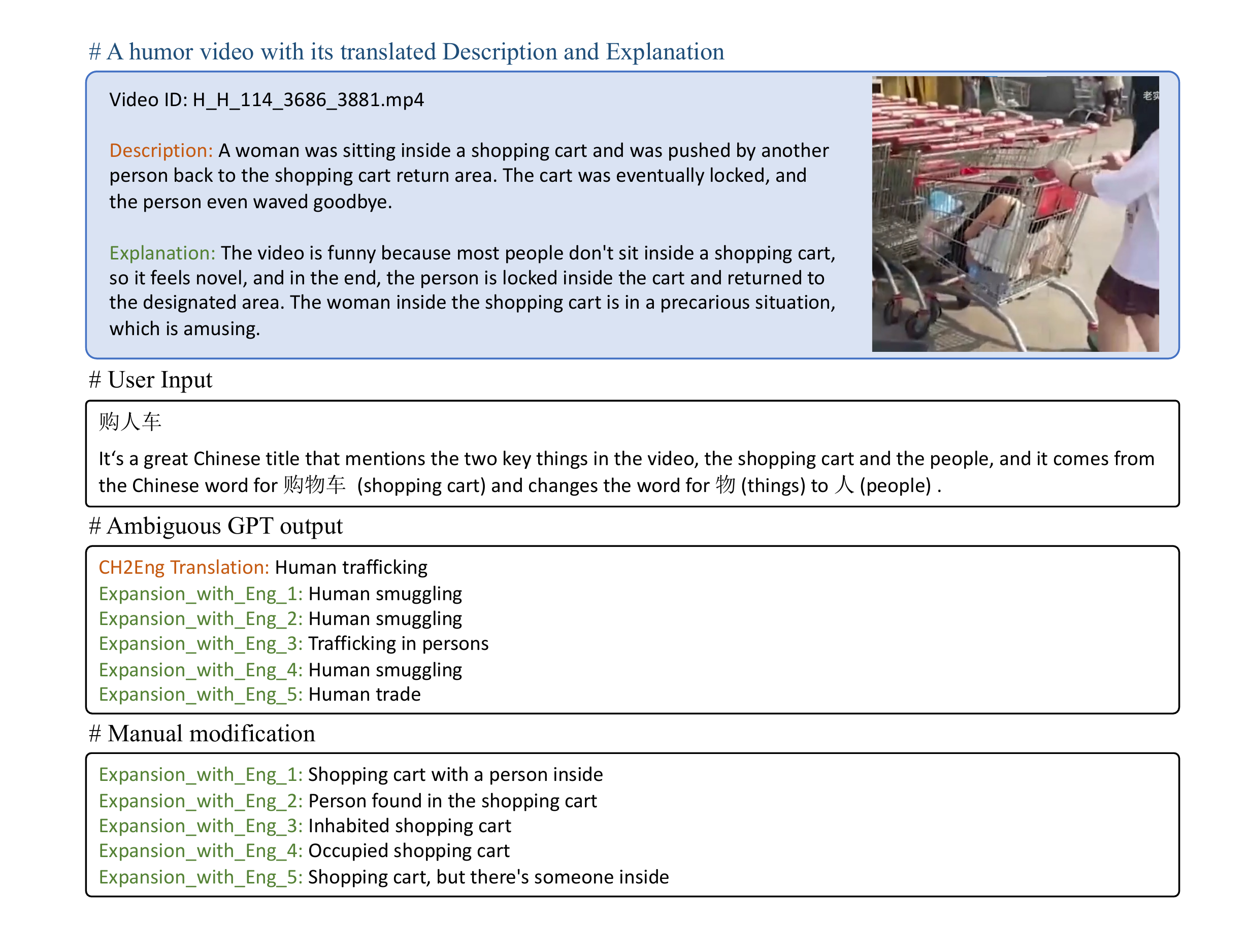}
    \vspace{-0.6cm}
    \caption{\textbf{Issues in translation caused by Chinese-English cultural differences.} In this example of translating and expanding a humorous video title, GPT-3.5 failed to understand the meaning of the original Chinese title, and we filtered out such data and made manual changes.}
    \vspace{-0.4cm}
    \label{fig:trans-err}
\end{figure}

\noindent\textbf{Pre-processing \& Qualification}\quad
For videos related to humor and magic, we downloaded them from different streaming platforms, mostly in the form of compilations. For Creative videos, we downloaded 26 episodes publicly available from Tokyo TV in Japan. We provided rigorous training to the annotators to ensure high-quality video clips in the final compilation. Annotators who successfully completed the Clip task according to the requirements are considered qualified and can proceed to the next stage of annotation.

\noindent\textbf{Training \& Annotation}\quad
We conducted systematic training for all annotators who passed the previous round of annotation, focusing on different tasks: \\
For the \textit{Counter-intuitive Timestamp Localization Task (H1, C1, and M1)}, the annotation format is a pair of numbers enclosed in square brackets, [xxxx, xxxx]. We asked the annotators to record the time intervals in which they felt pleasure (or amusement or shock) while watching the video. \\
For the \textit{Task Detailed Description Task (H2, C2, and M2)}, we requested objective descriptions of what happened at [xxxx, xxxx], emphasizing a "what you see is what you get" approach. It is important to note that the annotations should only cover the selected time intervals and should not include subjective adverbs (such as vividly, vividly, or wildly). When describing characters or objects, be concise and add modifiers if there is ambiguity. An example of a poor label is "a man and a woman," while a relatively better example is "a man wearing a red hat and a woman wearing an apron." 

For the \textit{Counter-intuitiveness Reasoning Task (H3, C3, and M3)}, explain why the video is interesting in the context of the overall content. This part requires interpretive answers based on analysis, reasoning, and prior knowledge to explain why the video is counter-intuitive.

Specifically for the \textit{Magic Method Task (M4)}, we found that most annotators were not professional magicians, and even when watching instructional magic videos, it was challenging for them to provide complete and reasonable annotations for magic methods. Therefore, for M4, we only provided a small number of examples, which do not exist in the FunQA dataset. Fig. \ref{fig:M4} shows an example of M4.

We conducted strict quality reviews in real-time for the annotations, including but not limited to checking if the annotations meet the requirements, if the selected videos contain sensitive content, and if the annotation texts contain sensitive content. The annotation principles (video selection and video clip, etc) can be seen in Fig.~\ref{fig:annotation-principle}. Fig. \ref{fig:annotation-error} also illustrates some errors encountered during the annotation process and the suggested modifications provided. We assured that every annotation included in the final release of FunQA has been subjected to rigorous \textbf{multi-person, multi-round} review processes. Specifically, every piece of released FunQA version annotated content underwent scrutiny by three distinct annotators. After completing all the annotations, we conducted a consensus evaluation study to assess the objectivity of our annotations. We randomly selected 10\% of the videos and asked annotators to provide a consistency score (high consensus, medium consensus, low consensus) between their first and second annotations, considering all the previous annotations shown to them.\\
\noindent\textbf{Copyright \& License}\quad
We respect the copyright of each video. 
Our data sharing approach is informed by the models established by previous studie like Kinetics~\cite{kay2017kinetics}, HD-VILA-100M~\cite{xue2022hdvila}, and others. Instead of providing the original raw data, we only supply the YouTube video IDs necessary for downloading the respective content.

We respect the personal identity information of everyone appearing in our videos and always strive to eliminate offensive content. By conducting strict reviews of the annotators and real-time quality checks during each annotation process, we ensure the absence of offensive content.
\subsection{Language Augmentation} \label{appendix_a: Augmentation}
For each QA pair, we first asked GPT-3.5 to translate the previous Chinese answers into English, and then let GPT-3.5 give five separate answers with the same meaning but different linguistic expressions. Afterward, we filtered out incorrect and incomplete generations. Fig. \ref{fig:prompt-trans} shows the prompt we gave to GPT-3.5.

However, as mentioned in the main text in the limitation, the translation of GPT-3.5 will be ambiguous when the original text uses Chinese harmonics and some special Chinese meanings, and this problem mostly occurs in the Title Task, we screened such problems by hand and modified them, as shown in Fig. \ref{fig:trans-err}.

\subsection{Data Statistics}\label{appendix_a: Dataset statics}
For a complete comparison between FunQA and other existing benchmarks, please see Table \ref{T:whole_main}.
We collected raw videos from multiple video platforms, including short videos, long videos, and video clips, and the detailed data can be seen in Table \ref{T:raw_statics}. The statics of FunQA dataset before GPT-3.5 extension can be seen in Table \ref{T:statics_before_gpt}. The word cloud of the all annotation word is shown in Fig. \ref{fig:wc_funqa} (a-d).

  

\begin{table}[!t]
\caption{\textbf{Statistics of the FunQA raw data.} 
\label{T:raw_statics}}
\centering
\resizebox{1.0\linewidth}{!}{

\begin{tabular}{cccccc}
\hline
FunQA Subset & Type & Source & \# Videos & Avg.len (s) & Total.len (h) \\ 
\hline
\multirow{3}{*}{Humor} & \multirow{2}{*}{Daily Life (Human)} & \multirow{3}{*}{Youtube}                 & 351    & 182  & 15.8   \\
                       &                                     &                    & 1296   & 14      & 5.18    \\
                       & Nature (Animal)                     &                  & 230    & 133  & 8.52    \\ \hline
Creative             & Performance                         & Youtube            & 26     & 6060  & 43.77   \\\hline
\multirow{2}{*}{MagicQA} & Close up Magic                      & \multirow{2}{*}{Youtube} & 765    & 96  & 20.40   \\ 
                       & Camera Magic                        &                          & 334    & 152  & 14.17   \\ \hline
FunQA                    & \textit{-}                          & -                        & 3002   & 129   & 107.87  \\ \hline\\
\end{tabular}   

}
\end{table}
\begin{table}[h]
\caption{\textbf{Statistics of the FunQA before GPT-3.5 extension.}}
\label{T:statics_before_gpt}
\centering
\resizebox{0.8\linewidth}{!}{
\begin{tabular}{ccccc}
\hline
Datasets & Avg length (s) & \# Clips (K) & \# QA pairs (K) & \# QA per clip\\
\hline
HumorQA & 7 & 1.8 & 7.2 & 4.0\\
CreativeQA & 48 & 0.9 & 4.5 & 5.0\\
MagicQA & 10 & 1.6 & 4.8 & 3.0\\
\hline
FunQA & 19 & 4.3 & 16.5 & 3.8\\
\hline
\end{tabular}
}
\label{tab:dataset-info}
\end{table}

\begin{figure*}
    \centering
    \begin{subfigure}[b]{0.2\textwidth}
        \centering
        \includegraphics[width=0.8\textwidth]{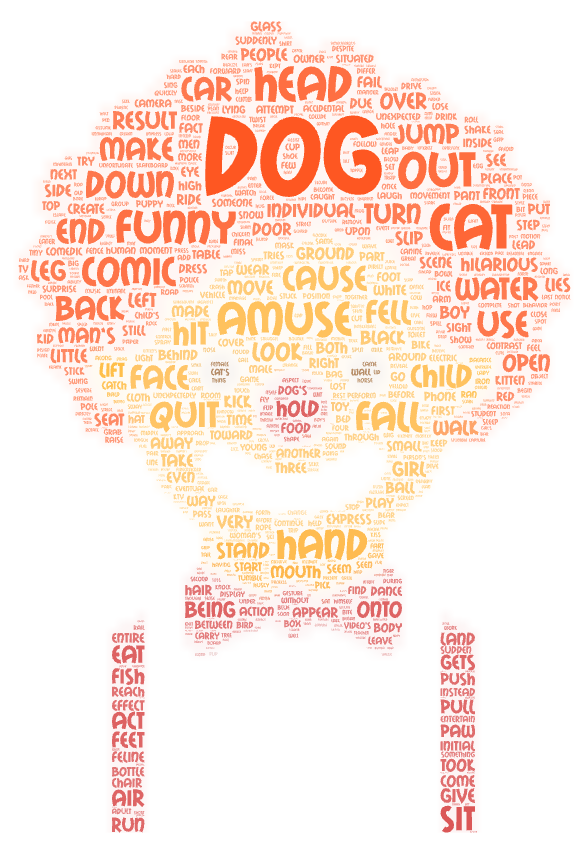}
        \caption{\textbf{HumorQA.}}
        \label{fig:wc_human}
    \end{subfigure}
    \begin{subfigure}[b]{0.2\textwidth}
        \centering
        \includegraphics[width=1\textwidth]{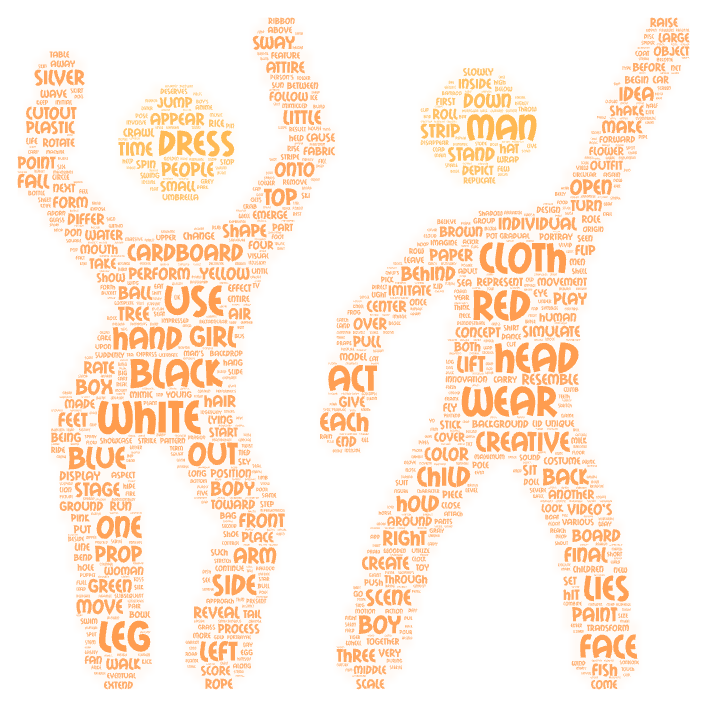}
        \caption{\textbf{CreativeQA.}}
        \label{fig:wc_creat}
    \end{subfigure}
    \begin{subfigure}[b]{0.20\textwidth}
        \centering
        \includegraphics[width=1\textwidth]{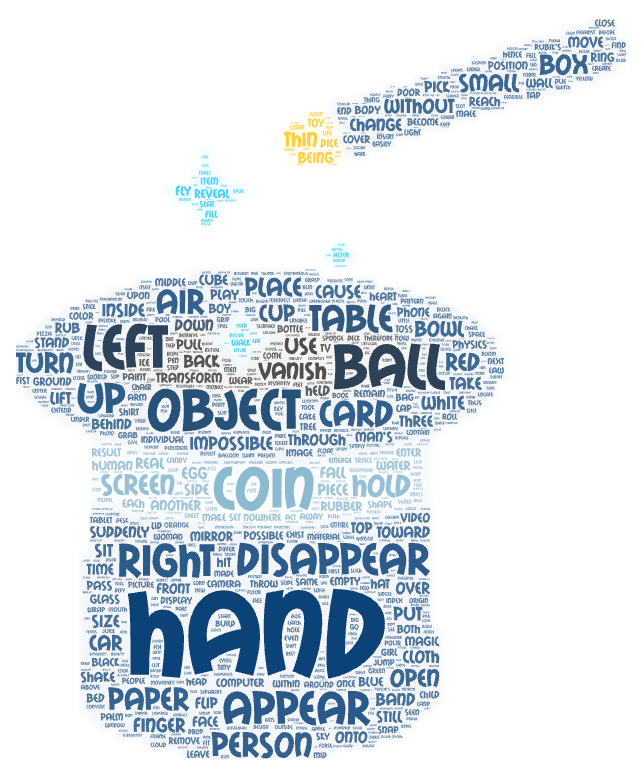}
        \caption{\textbf{MagicQA.}}
        \label{fig:wc_magic}
    \end{subfigure}
    \begin{subfigure}[b]{0.20\textwidth}
        \centering
        \includegraphics[width=1\textwidth]{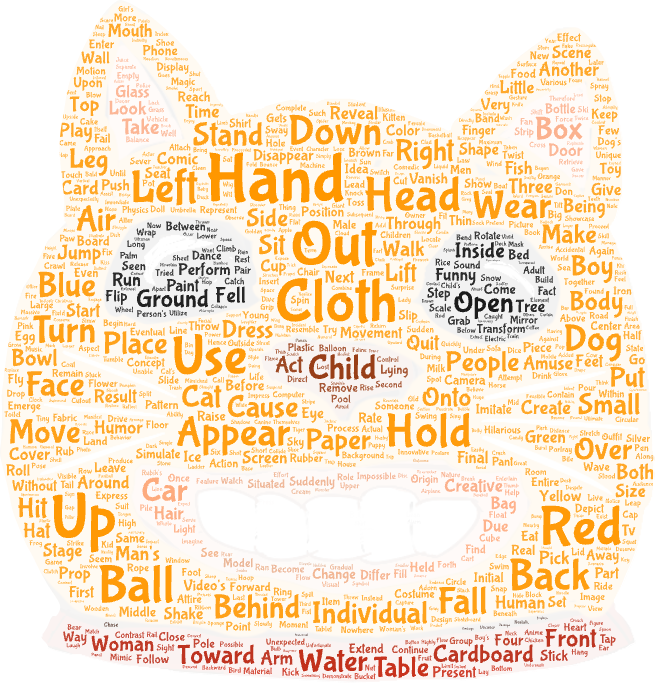}
        \caption{\textbf{FunQA.}}
        \label{fig:wc_funqa}
    \end{subfigure}

    \caption{\textbf{Word cloud of FunQA free-text answer.}}
    \label{fig:wc_funqa}
\end{figure*}
\section{FunQA Extension Datasets}\label{appendix_b: FunQA Extended dataset}
Our main objective in designing these extended datasets is to leverage our high-quality annotated data and provide a rich and suitable data format for models. In addition, we also wanted to test the capability of GPT-3.5 and the quality of the dataset, and it turns out that GPT-3.5 can expand our data tens of times after providing high-quality free-text annotations, and our FunQA dataset is extremely scalable.
\begin{figure*}[htbp]
    \centering
    \includegraphics[width=\linewidth]{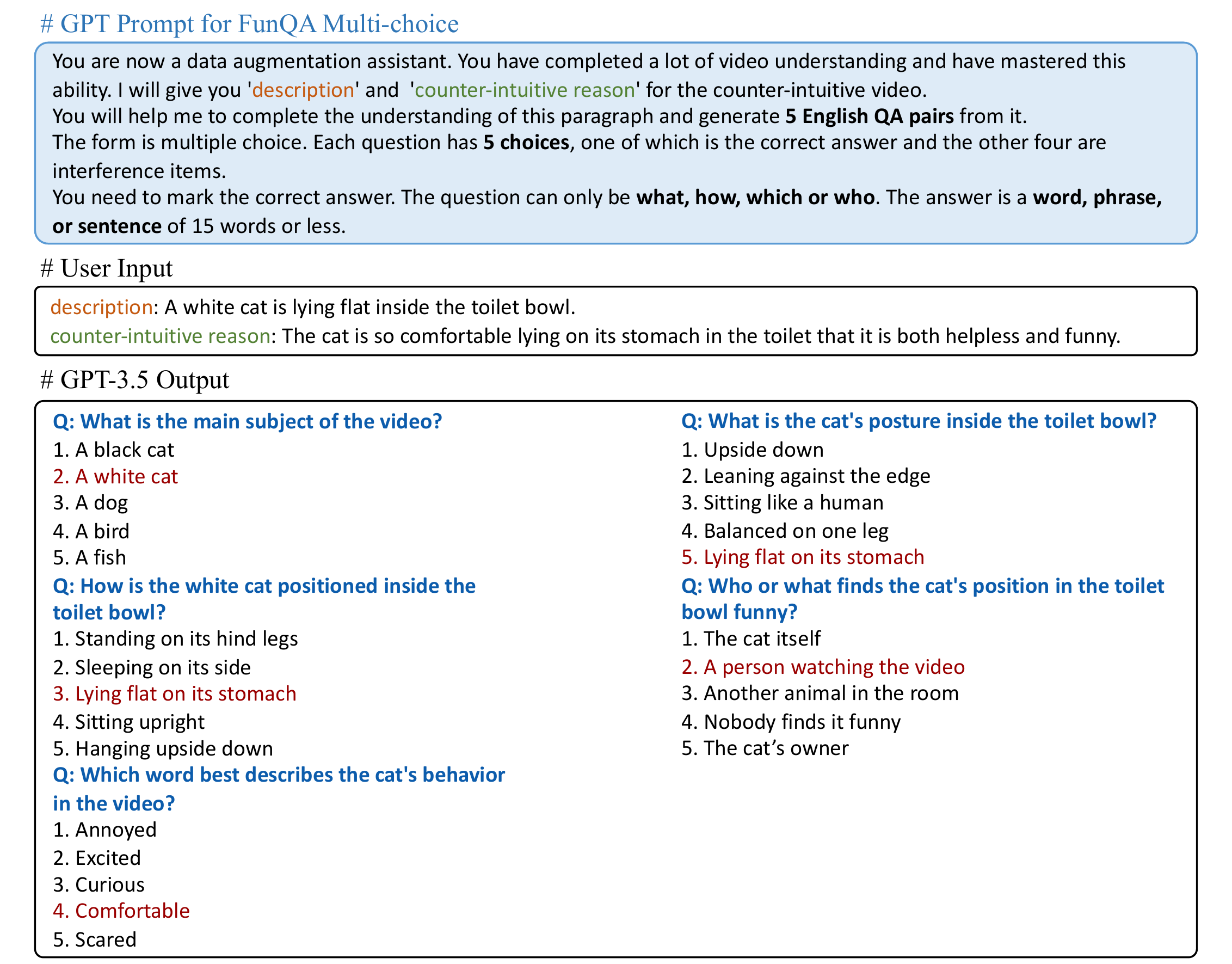}
    \vspace{-0.6cm}
    \caption{\textbf{GPT prompt for FunQA Multi-choice.} We give GPT-3.5 prompt as system, and then enter the description and explanation of the video, and GPT-3.5 will output the data in the form of multiple-choice questions according to the prompt.}
    \vspace{-0.4cm}
    \label{fig:prompt-mcqa}
\end{figure*}

\begin{figure*}[htbp]
    \centering
    \includegraphics[width=\linewidth]{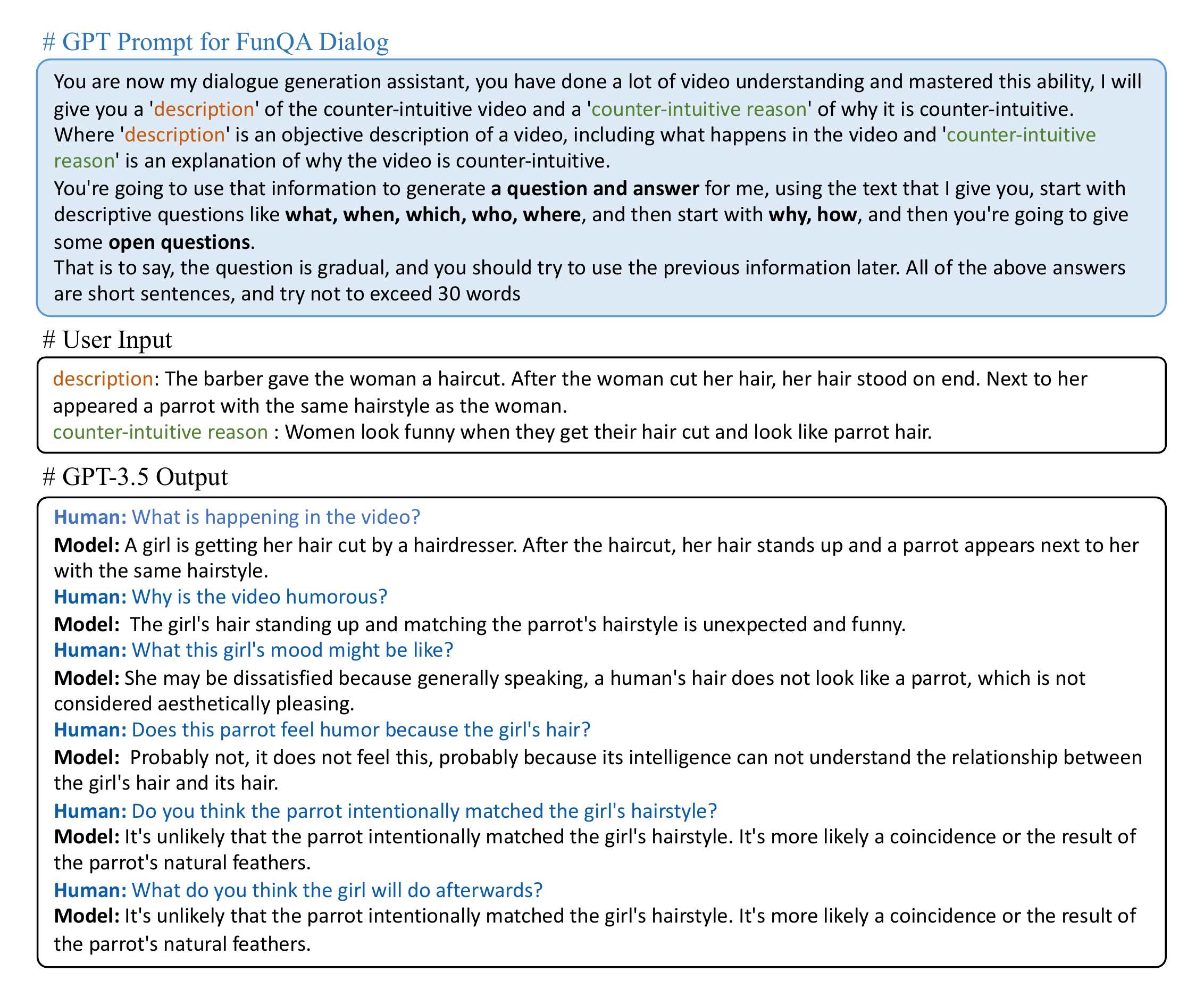}
    \vspace{-0.6cm}
    \caption{\textbf{GPT prompt for FunQA Dialog.} We give GPT prompt as system, and then input the description and explanation of the video, and GPT-3.5 will output the data in the form of dialogue according to the prompt.}
    \vspace{-0.4cm}
    \label{fig:prompt-dia}
\end{figure*}
\subsection{FunQA-MC Dataset} \label{appendix_b: FunQA MC}
FunQA-MC (Multi-choice) Dataset is prepared to provide training and testing for arbitrary models, in this dataset our QA pairs are in the form of multiple choice, the answer is a word, phrase, or short sentence, and the type of questions are all descriptions. An example of the prompt given to GPT-3.5 and the data generated can be seen in Fig. \ref{fig:prompt-mcqa}. For the \textbf{FunQA-MC-R}, which is a multi-choice version specifically containing counter-intuitive reasoning questions, we randomly chose 100 QA pairs from Task 2 ( Counter-intuitiveness Reasoning Task) to create the subset.
\subsection{FunQA-DIA Dataset}\label{appendix_b: FunQA Dia}
Most of the current LLMs are in the form of dialogues. To cater to their data input, we produced the FunQA-DIA (Dialog) dataset, in which we used GPT-3.5 to convert QA pairs into recursive dialogues with added context. In addition to this, we also let GPT-3.5 freely generate future dialogue development, which greatly expands the amount of information. An example of the prompt given to GPT-3.5 and the data generated can be seen in Fig. \ref{fig:prompt-dia}. 

\section{More Details of FunMentor}\label{appendix_c: More Details of FunMentor}
\subsection{Real Fact Question}
\begin{figure}[t]
    \centering
    \includegraphics[width=0.6\linewidth]{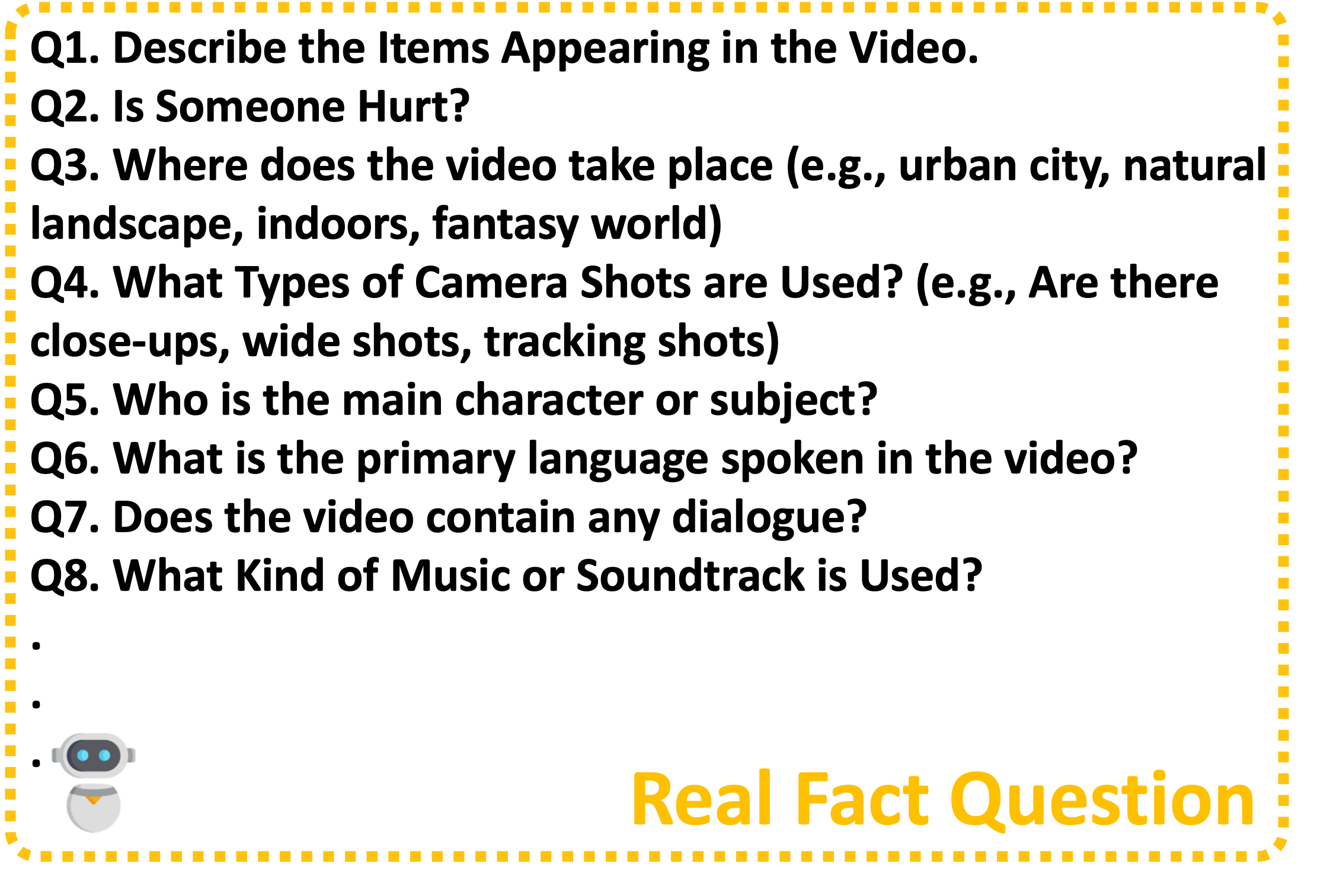}
    \caption{\textbf{Example of Real fact question by FunMentor.} }
    \vspace{-0.4cm}
    \label{fig:real-fact}
\end{figure}

\noindent\textbf{}\quad 
During the \textbf{Real Fact Collection} process, FunMentor relies on designing questions that VLMs can answer accurately and objectively to gain knowledge of the video’s basic information. The specific questions are shown in Fig. \ref{fig:real-fact}.
\begin{figure}[h]
    \centering
    \includegraphics[width=0.8\linewidth]{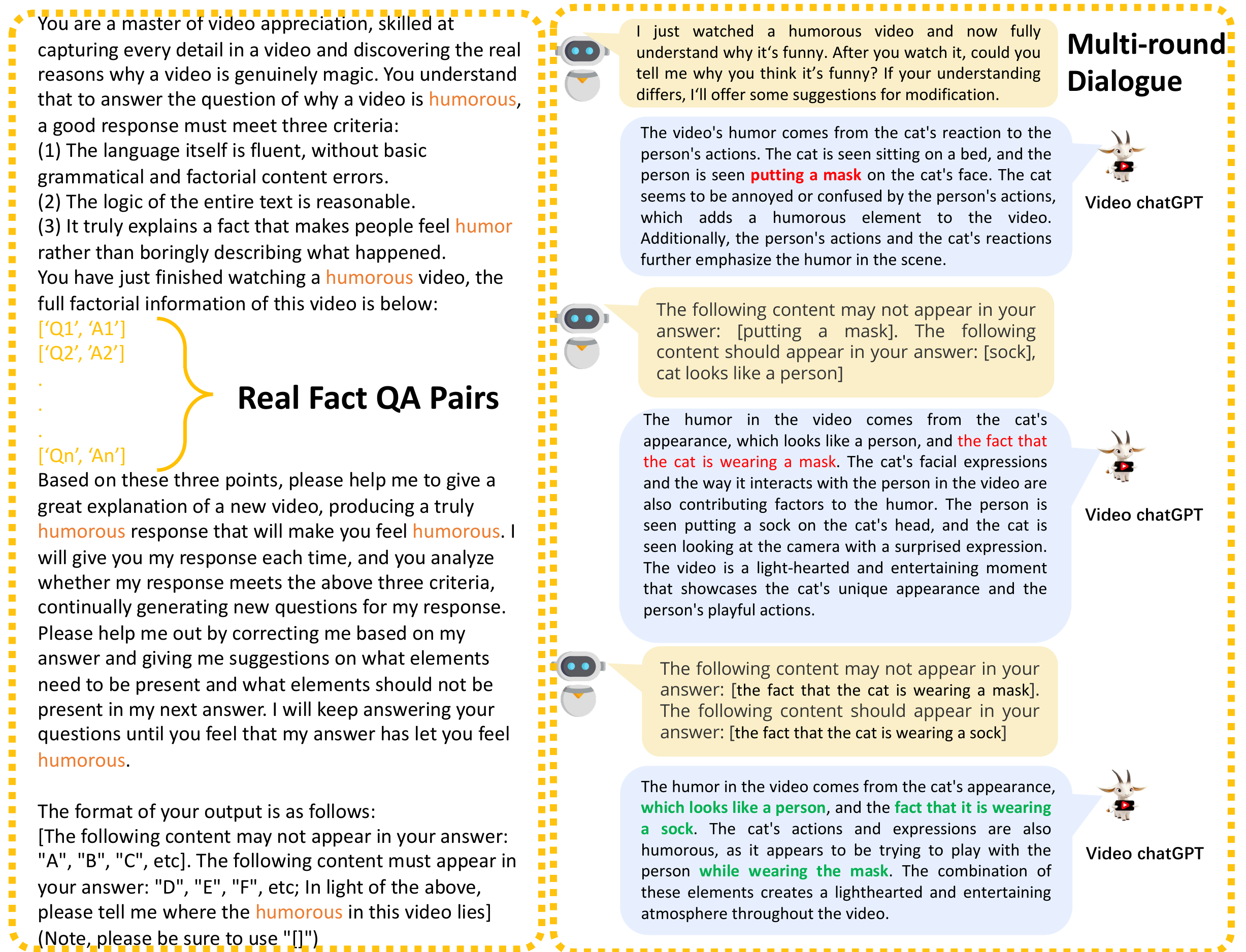}
    \caption{\textbf{Prompt Design of FunMentor in multi-round dialogue.} On the left are the prompts from FunMentor, which incorporate the information from collected \textbf{Real Facts} and conduct \textbf{Answer Judgement} to \textbf{generate suggestions}.}
    \vspace{-0.4cm}
    \label{fig:funmentor_exam}
\end{figure}
\subsection{Prompts Design}
Based on the QA pairs collected as mentioned, FunMentor engages in multi-round dialogues with VLMs through our predefined prompts. This process involves providing increasingly precise instructions, guiding the VLM towards correctly answering the question. Specific examples of this are shown in Fig. \ref{fig:funmentor_exam}.

\begin{table}[h]
\caption{\textbf{Main Results on FunQA Benchmark (Complete Version).} H2, C2, M2 represent the detailed video description task, and H3, C3, M3 represent reasoning around counter-intuitiveness. For the higher-level tasks, H4, C4 involve attributing a fitting and vivid title. The responses for all these tasks in free-text format. We use the following metrics: \textbf{BLEU-4 / ROUGE-L / CIDEr} (shown in the first row) and \textbf{BLEURT / GPT-4} (shown in the second row) for evaluation. C5 represents scoring the video creativity, and the metric is the \textbf{Accuracy} between the predicted score and the official score. We tested the caption-based and instruction-based models. 
Here we clarify the abbreviation in the table.  \textbf{L.M.}: GIT\_LARGE\_MSRVTT; \textbf{L.V.}: GIT\_LARGE\_VATEX; \textbf{D.C.} means finetuned on Dense Caption; \textbf{FunQA} means finetuned on FunQA. }
\label{T:whole_experiment}
\vspace{-10pt}
{\renewcommand\baselinestretch{1.5}\selectfont
\setlength\tabcolsep{10pt} 
\resizebox{1.0\linewidth}{!}{
\begin{tabular}{l|ccc|cccc|cc}
\toprule
\multirow{2}{*}{} & \multicolumn{3}{c}{HumorQA} & \multicolumn{4}{c}{CreativeQA} & \multicolumn{2}{c}{MagicQA} \\

\cmidrule(l){2-4} \cmidrule(l){5-8} \cmidrule(l){9-10}
Task & H2-Des. & H3-Rea. & H4-Title & C2-Des. & C3-Rea. & C4-Title & C5-Score & M2-Des. & M3-Rea. \\
\toprule
\multicolumn{10}{l}{\textbf{- Caption-based Model}} \\

mPLUG \cite{li2022mplug}  
& \makecell[c]{\textbf{1.5} / 16.4 / \textbf{1.0} \\ 19.9 / 3.9 }
& \makecell[c]{1.1 / 12.5 / 0.4 \\ \textbf{25.7} / \textbf{6.0}}
& \makecell[c]{0.6 / 7.5 / 0.1 \\ 22.1 / \textbf{11.2}}

& \makecell[c]{0.4 / 13.4 / 0.0 \\ 14.9 / 3.0}
& \makecell[c]{0.7 / 12.6 / 0.1 \\ \textbf{24.2} / \textbf{6.9}}
& \makecell[c]{0.3 / 3.2 / 0.0 \\ \textbf{20.8} / \textbf{18.8}}
& \makecell[c]{\\ 0.0}

& \makecell[c]{1.2 / 15.8 / 0.5 \\ 19.7 / 4.0}
& \makecell[c]{0.9 / 8.9 / 0.4 \\ \textbf{21.2} / \textbf{8.1}}\\
\toprule
GIT (L.M.) \cite{wang2022git} 
& \makecell[c]{0.5 / 12.8 / 0.2 \\ 22.4 / 3.6}
& \makecell[c]{0.0 / 0.0 / 0.0 \\ 0.0 / 0.0}
& \makecell[c]{\textbf{1.1} / 7.7 / \textbf{0.7} \\ 17.0 / 8.9}

& \makecell[c]{0.0 / 6.40 / 0.0 \\ 14.4 / 3.8}
& \makecell[c]{0.0 / 0.0 / 0.0 \\ 0.0 / 0.0}
& \makecell[c]{0.3 / 1.5 / 0.2 \\ 7.1 / 11.3}
& \makecell[c]{\\ 0.0}

& \makecell[c]{0.2 / 11.2 / 0.1 \\ 19.4 / 8.2}
& \makecell[c]{0.0 / 0.0 / 0.0 \\ 0.0 / 0.0} \\

\toprule
GIT (L.V.) \cite{wang2022git} 
& \makecell[c]{1.2 / 16.9 / 0.6 \\ \textbf{33.3} / \textbf{4.0}}
& \makecell[c]{0.0 / 0.0 / 0.0 \\ 0.0 / 0.0}
& \makecell[c]{1.0 / \textbf{8.8} / \textbf{0.7} \\\textbf{25.9} / 10.0}

& \makecell[c]{0.1 / 8.30 / 0.0 \\ \textbf{20.5} / \textbf{4.2}}
& \makecell[c]{0.0 / 0.0 / 0.0 \\ 0.0 / 0.0}
& \makecell[c]{\textbf{0.5} / 2.8 / \textbf{0.4} \\10.5 / 12.0}
& \makecell[c]{\\ 0.0}

& \makecell[c]{0.6 / 13.7 / 0.1 \\ \textbf{29.8} / \textbf{8.6}}
& \makecell[c]{0.0 / 0.0 / 0.0 \\ 0.0 / 0.0} \\
\midrule
\toprule
\multicolumn{10}{l}{\textbf{- Instruction-based Model}} \\
VideoChat \cite{2023videochat} 
& \makecell[c]{0.5 / 13.7 / 0.0 \\44.0 / 17.9}
& \makecell[c]{0.5 / 13.5 / 0.0 \\45.4 / 31.9}
& \makecell[c]{0.8 / 5.1 / 0.5 \\20.2 / 31.7}

& \makecell[c]{0.3 / 7.50 / 0.0 \\21.7 / 5.9}
& \makecell[c]{0.3 / 7.70 / 0.0 \\22.8 / \textbf{17.7}}
& \makecell[c]{0.2 / 1.2 / 0.2 \\7.3 / 31.1}
& 67.5 

& \makecell[c]{0.6 / 15.5 / 0.0 \\ 47.4 / 8.2}
& \makecell[c]{0.3 / 9.2 / 0.0 \\ 43.1 / 44.6} \\

\toprule
Video-ChatGPT \cite{maaz2023videochatgpt} 
& \makecell[c]{0.5 / 14.0 / 0.1 \\ 39.9 / \textbf{24.3}}
& \makecell[c]{0.7 / 12.4 / 0.1 \\ 40.1 / 24.9}
& \makecell[c]{0.4 / 3.2 / 0.2 \\ 36.5 / 41.2}

& \makecell[c]{1.1 / \textbf{19.8} / 0.2 \\ \textbf{45.8} / 6.6}
& \makecell[c]{0.8 / 17.3 / 0.1 \\ 45.2 / 9.1}
& \makecell[c]{0.2 / 1.9 / 0.2 \\ 30.9 / \textbf{48.8}}
& \textbf{85.4} 

& \makecell[c]{0.7 / 20.8 / 0.0 \\ \textbf{50.8} / \textbf{11.2}}
& \makecell[c]{0.5 / 11.3 / 0.0 \\ 43.3 / 40.4} \\

\toprule
mPLUG-Owl \cite{ye2023mplugowl} 
& \makecell[c]{0.5 / 13.3 / 0.2 \\  44.5 / 10,7}
& \makecell[c]{0.5 / 13.2 / 0.2 \\ \textbf{47.3} / \textbf{35.0}}
& \makecell[c]{0.6 / 5.3 / 0.0 \\ 29.8 / \textbf{48.8}}

& \makecell[c]{0.8 / 16.5 / 0.1 \\ 43.0 / 5.0}
& \makecell[c]{0.7 / 13.3 / 0.0 \\ \textbf{44.7 }/ 10.6}
& \makecell[c]{0.1 / 1.1 / 0.1 \\ 23.9 / 36.3}
& 66.7 

& \makecell[c]{0.4 / 16.7 / 0.0 \\ 46.4 / 8.6}
& \makecell[c]{0.5 / 8.4 / 0.0 \\ 43.9 / 30.9} \\

\toprule
Video-LLaMA \cite{damonlpsg2023videollama} 
& \makecell[c]{0.3 / 12.1 / 0.1 \\ \textbf{48.4} / 7.7}
& \makecell[c]{0.5 / 12.6 / 0.0 \\ 42.9 / 29.0}
& \makecell[c]{0.5 / 0.4 / 0.0 \\ 46.5 / 34.1}

& \makecell[c]{0.7 / 8.80 / 0.1 \\ 45.5 / 7.2}
& \makecell[c]{0.4 / 9.70 / 0.0 \\ 41.1 / 17.2}
& \makecell[c]{0.3 / 0.8 / 0.1 \\ 42.3 / 31.2}
& 64.2 

& \makecell[c]{0.4 / 14.3 / 0.0 \\ 50.1 / 10.2}
& \makecell[c]{0.5 / 7.9 / 0.0 \\ 39.0 / 28.0} \\

\toprule
Otter (D.C.) \cite{li2023otter}  
& \makecell[c]{1.1 / 14.3 / 0.4 \\ 30.2 / 7.7}
& \makecell[c]{1.2 / 14.2 / 0.4 \\ 32.3 / 28.3}
& \makecell[c]{0.5 / 5.4 / 0.1 \\ 21.7 / 20.0}

& \makecell[c]{0.5 / 13.8 / 0.1 \\ 28.7 / 1.7}
& \makecell[c]{1.0 / 16.8 / 0.2 \\32.9 / 7.9}
& \makecell[c]{0.3 / 2.3 / 0.1 \\ 17.7 / 36.3}
& 45.0

& \makecell[c]{1.0 / 15.0 / 0.3 \\ 32.5 / 2.1}
& \makecell[c]{1.1 / 12.8 / 0.2 \\ 27.3 / 36.8} \\
\toprule
Otter (FunQA) \cite{li2023otter} 
& \makecell[c]{\textbf{1.5} / \textbf{18.1} / 0.9 \\ 38.4 / 8.9}
& \makecell[c]{\textbf{1.3} / \textbf{15.4} / \textbf{0.5}\\ 42.6 / 31.7}
& \makecell[c]{0.8 / 5.9 /  0.5 \\ \textbf{47.5} / 32.1}

& \makecell[c]{\textbf{1.5} / 19.6 / \textbf{0.5} \\ 40.0 / \textbf{7.3}}
& \makecell[c]{\textbf{2.2} / \textbf{21.2} / \textbf{0.5} \\ 41.1 / 8.8}
& \makecell[c]{0.3 / \textbf{4.3} / 0.3 \\ \textbf{44.5} / 38.8}
& 69.4

& \makecell[c]{\textbf{2.6} / \textbf{23.8} / \textbf{1.6} \\ 44.7 / 10.3}
& \makecell[c]{\textbf{3.4} / \textbf{20.3} / \textbf{2.6} \\ \textbf{44.5} / \textbf{47.5}} \\
\midrule
\toprule
Video-ChatGPT \cite{maaz2023videochatgpt}  + \textbf{FunMentor (Ours)}
& \makecell[c]{0.6 / 14.2 / 0.6 \\ \textbf{65.2} / \textbf{33.2}}
& \makecell[c]{1.1 / 14.4 / 0.3 \\ \textbf{57.5} / 36.5}
& \makecell[c]{1.0 / 5.7 / 0.6 \\ 50.2 / \textbf{65.1}}

& \makecell[c]{1.4 / 20.2 / 0.3 \\ \textbf{66.3} / \textbf{14.2}}
& \makecell[c]{1.4 / 18.8 / 0.2 \\ \textbf{58.7} / \textbf{23.4}}
& \makecell[c]{0.3 / 2.2 / \textbf{0.3} \\ 45.3 / \textbf{52.2}}
& \textbf{85.4}

& \makecell[c]{1.1 / 23.2 / 0.4 \\ \textbf{55.1} / \textbf{13.3}}
& \makecell[c]{2.1 / 14.4 / 0.9 \\ \textbf{46.3} / 54.8} \\
\toprule
Otter (FunQA) \cite{li2023otter} + \textbf{FunMentor (Ours)}
& \makecell[c]{\textbf{1.9} / \textbf{20.2} / \textbf{1.1} \\ 33.4 / 13.4}
& \makecell[c]{\textbf{1.7} / \textbf{19.3} / \textbf{0.9} \\ 37.8 / \textbf{45.8}}
& \makecell[c]{\textbf{0.9} / \textbf{6.8} / \textbf{0.9} \\ \textbf{58.3} / 34.2}

& \makecell[c]{\textbf{2.4} / \textbf{23.1} / \textbf{1.1} \\ 60.4 / 11.0}
& \makecell[c]{\textbf{3.3} / \textbf{26.5} / \textbf{0.7} \\ 44.4 / 9.3}
& \makecell[c]{\textbf{0.4} / \textbf{4.7} / \textbf{0.3} \\ \textbf{53.9} / 43.5}
& 69.4
 
& \makecell[c]{\textbf{3.1} / \textbf{25.5} / \textbf{2.5} \\ 43.5 / 12.81}
& \makecell[c]{\textbf{5.6} / \textbf{21.2} / \textbf{3.4} \\ 38.91 / \textbf{56.4}}\\

\bottomrule

\end{tabular}}\par
}
\vspace{-0.5cm}
\end{table}
\section{More Details of Experiment} \label{appendix_d: Experiments}
The complete result of FunQA Benchmark (with traditional metircs' scores) are shown in Table \ref{T:whole_experiment}.
\subsection{Significance of New GPT-4 Based Metric} \label{appendix_d: Evaluation Metrics}
\begin{figure*}[htbp]
    \centering
    \includegraphics[width=\linewidth]{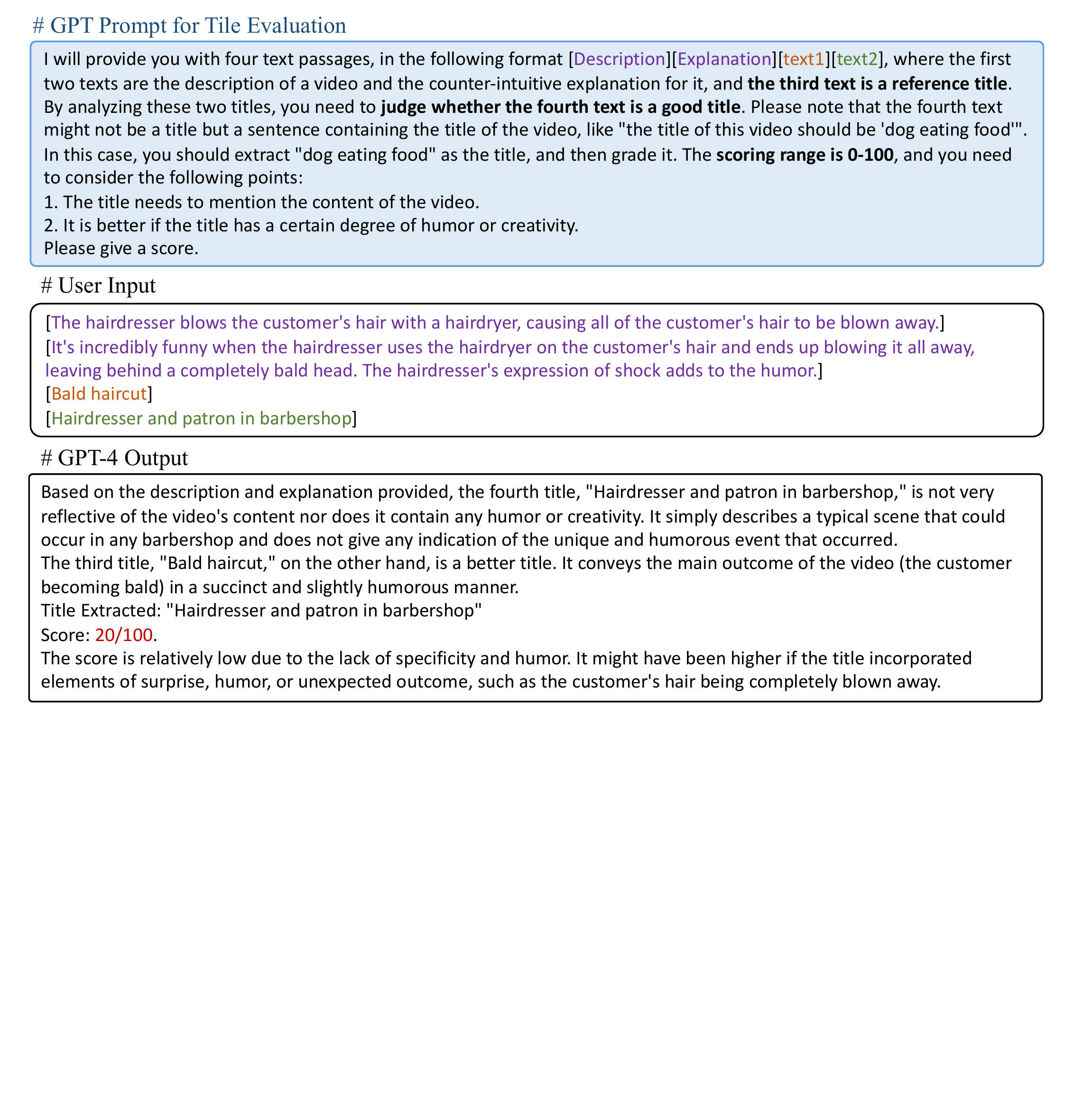}
    \vspace{-0.6cm}
    \caption{\textbf{GPT-4 prompt for Title Evaluation.} We give GPT prompt as system, and then input a description, explanation, and two titles, the first one is our annotation, and the second one is the output of the model, and GPT-4 will evaluate the similarity between these two texts according to the prompt's requirements.}
    \vspace{-0.4cm}
    \label{fig:prompt-eval-title}
\end{figure*}

\begin{figure*}[htbp]
    \centering
    \includegraphics[width=\linewidth]{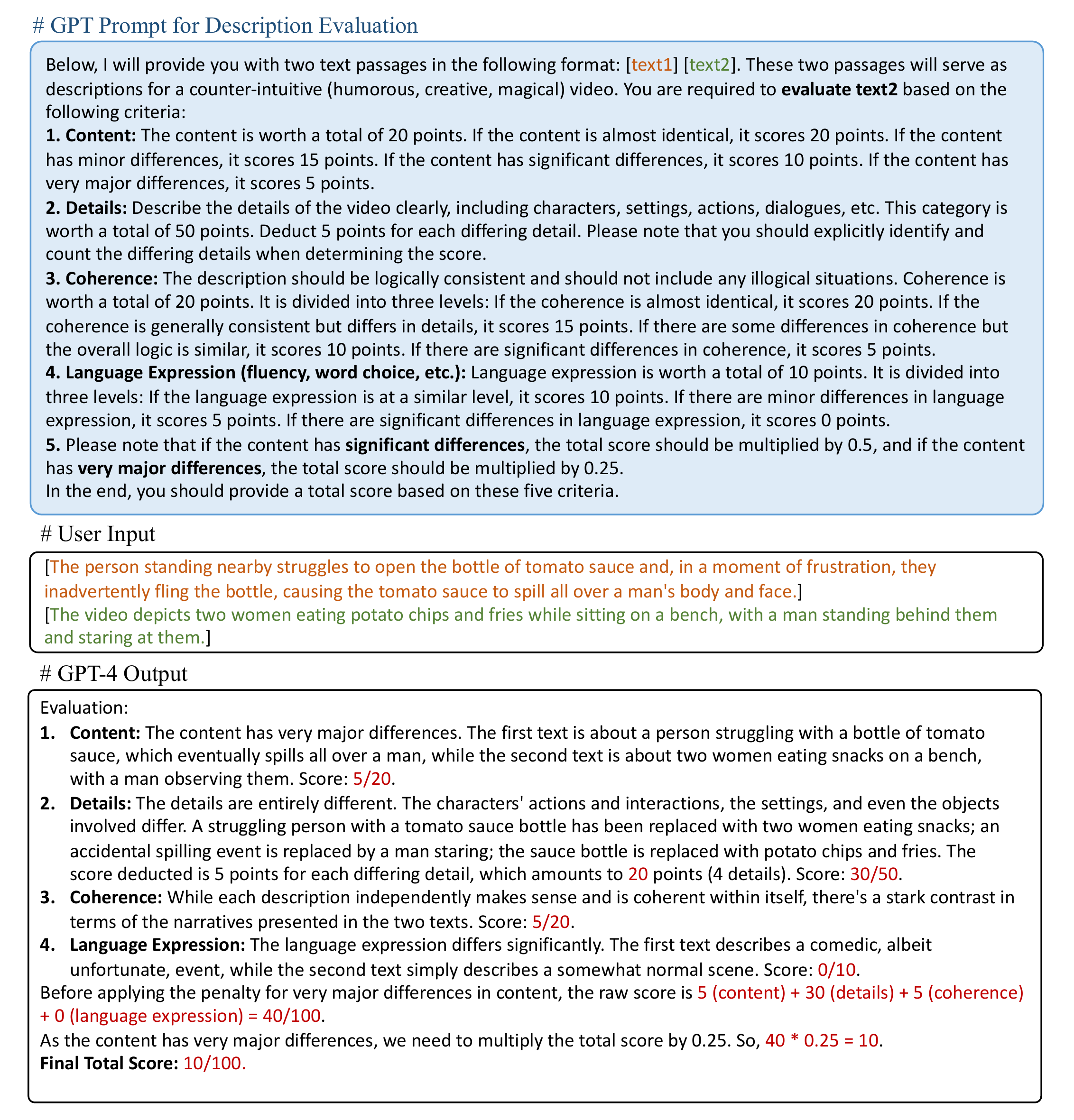}
    \vspace{-0.6cm}
    \caption{\textbf{GPT-4 prompt for Description Evaluation.} We give GPT-4 prompt as system, and then input two descriptions, the first one is our annotation, and the second one is the output of the model, and GPT-4 will evaluate the similarity between these two texts according to the prompt's requirements.}
    \vspace{-0.4cm}
    \label{fig:prompt-eval-des}
\end{figure*}

\begin{figure*}[htbp]
    \centering
    \includegraphics[width=\linewidth]{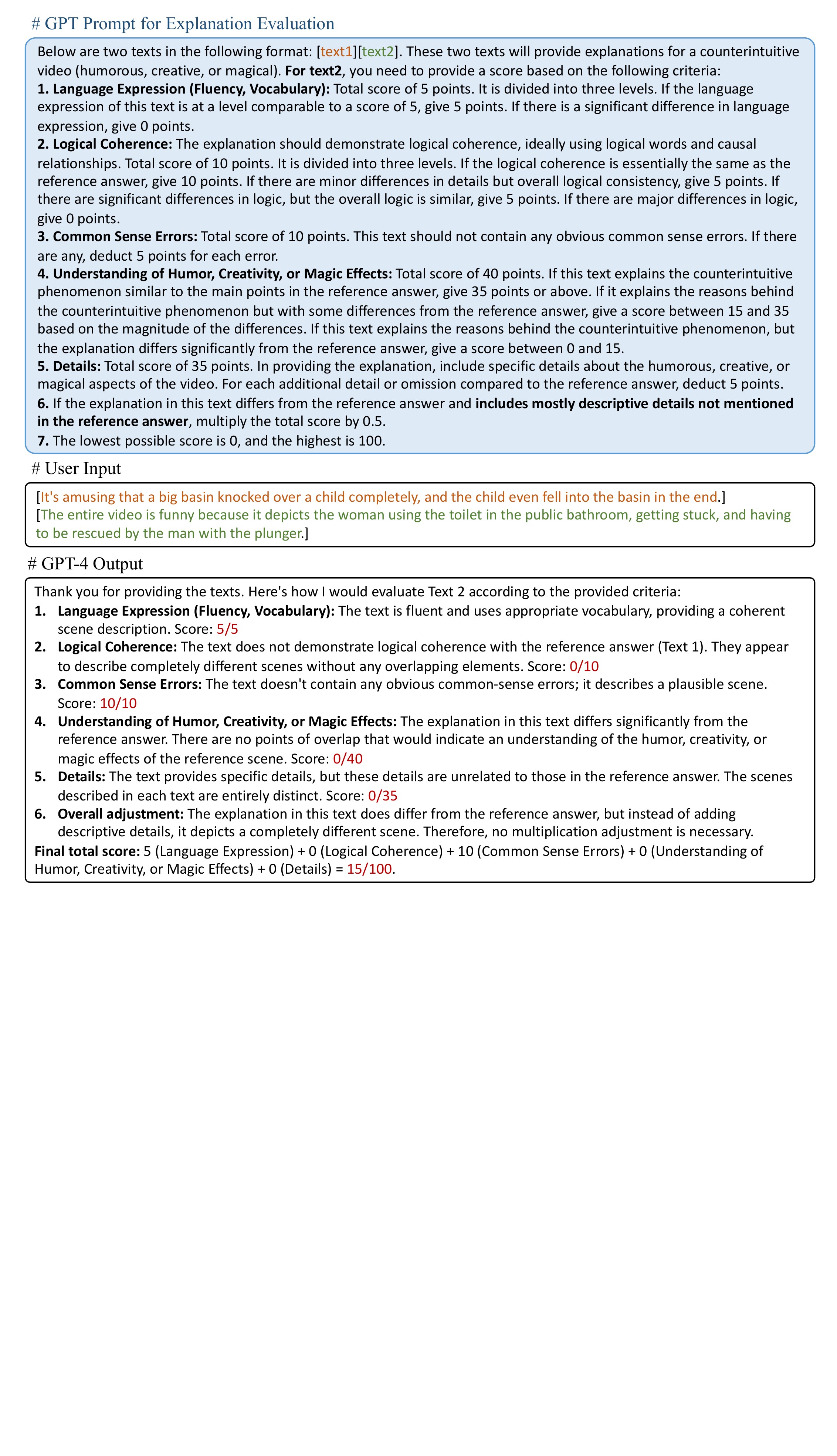}
    \vspace{-0.6cm}
    \caption{\textbf{GPT-4 prompt for Explanation Evaluation.} We give GPT prompt as system, and then input two explanations, the first one is our annotation, and the second one is the output of the model, and GPT-4 will evaluate the similarity between these two texts according to the prompt's requirements.}
    \vspace{-0.4cm}
    \label{fig:prompt-eval-exp}
\end{figure*}

\noindent\textbf{GPT-4 prompt design}\quad 
For each of the three tasks, we designed three prompts for scoring. \textit{For Detailed Description Task (H2, C3, and M2)}, we designed the prompt in five areas: text length, text content variation, text detail variation, logical text description, and linguistic ability.
For \textit{Counter-intuitiveness Reasoning Task (H3, C3, and M3)}, we designed the prompt in six areas: expressiveness of language, the logic of response, the common sense of response, understanding of counter-intuition, differences in text detail, and length of text.
For \textit{Title Task (H4 and C4)}, we used the description, comprehension, and title of the manually annotated video as a reference to score the new title. Each Prompt can be seen in Fig. \ref{fig:prompt-eval-title}, \ref{fig:prompt-eval-des}, and \ref{fig:prompt-eval-exp}.


\begin{figure*}[htbp]
    \centering
    \includegraphics[width=\linewidth]{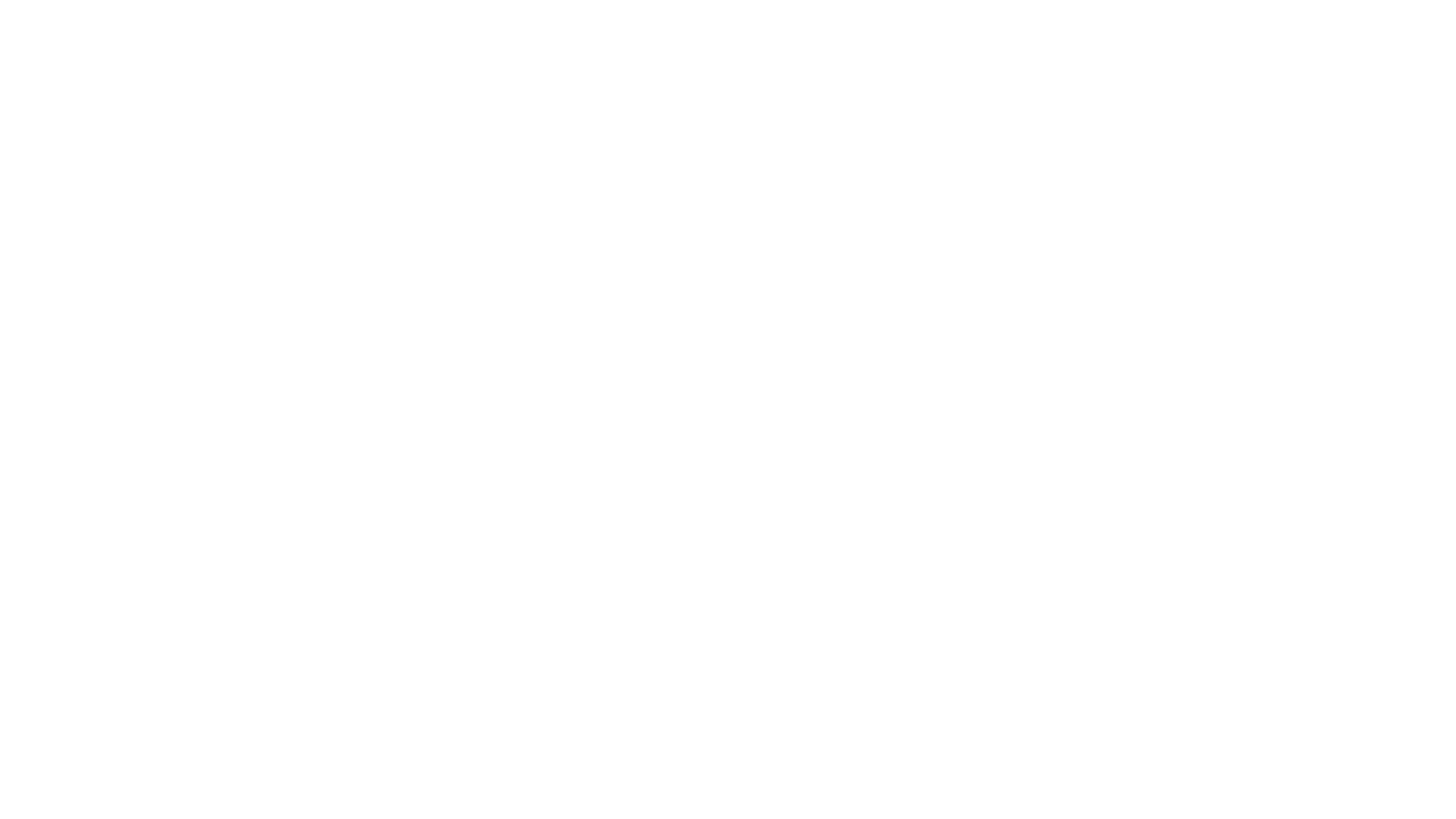}
    \caption{\textbf{The responses of VLMs to NExT-OE are scored using the WUPS and GPT-4 based metric.} It can be observed that Otter's responses, whether correct or close to the answer, receive scores near zero under the WUPS metric. In contrast, HGA's completely incorrect responses (e.g., answering ``cat'' when the correct answer is ``dog'') score highly under WUPS. On the other hand, the scoring by GPT-4 appears to be accurate.}
    \label{fig:wups}
\end{figure*}

\noindent\textbf{Comparison between GPT-4 and Traditional Metrics}\quad
The principle of the traditional metrics is relatively simple. The traditional metric principles are as follows.\\
\noindent\textbf{BLEU-4}\quad BLEU, full name is Bilingual Evaluation Understudy, is a commonly used machine translation evaluation metric. It evaluates how good a machine translation result is by comparing how well it matches the N-gram of one or more human-translated reference results, which is a sequence of N consecutive words. BLEU-4, i.e., evaluates how well a combination of two words (i.e., a binary) matches. BLEU introduces a correction factor, Brevity Penalty (BP), to avoid this problem, which penalizes machine translation if the result is shorter than the reference translation. The BLEU score is the geometric mean of the individual N-gram accuracy multiplied by the shortness penalty. That is, the score of BLEU takes into account the precision and length of the translation result.\\
\noindent\textbf{ROUGE-L}\quad
ROUGE, known as Recall-Oriented Understudy for Gisting Evaluation, is a commonly used evaluation method for tasks such as automatic digesting and machine translation. ROUGE is mainly evaluated by comparing the overlap between the generated abstracts and the reference abstracts. Among them, ROUGE-L is an important variant of ROUGE, where L stands for Longest Common Subsequence (LCS), i.e., the longest common subsequence. Unlike the n-gram, the longest common subsequence does not require consecutive occurrences of items in the sequence.
\\
\noindent\textbf{CIDEr}\quad
CIDEr, known as Consensus-based Image Description Evaluation, is an evaluation metric specifically designed for evaluating image description (Image Captioning) tasks. The main advantage of CIDEr is that it can capture more detailed information because it uses TF-IDF weights to emphasize n-grams that occur frequently in manual annotation but are not common in all image descriptions.\\
\noindent\textbf{BLEURT}\quad
BLEURT, full name Bilingual Evaluation Understudy with Representations from Transformers, is an evaluation method based on the transformer model, specifically for evaluating the output quality of machine translation and natural language generation tasks. Unlike traditional evaluation metrics such as BLEU and ROUGE, BLEURT does not directly compare the n-gram match between generated text and reference text but uses pre-trained language models (e.g., BERT) to understand the semantic information of text. the advantage of BLEURT is that it can capture the deep semantic information of text, and can address some problems that are difficult to be handled by traditional evaluation metrics (e.g. synonym substitution, utterance rearrangement) giving reasonable evaluation.\\
\noindent\textbf{WUPS}\quad
The Wu-Palmer similarity (WPUS) measure calculates how similar two word senses are. It considers the depths of the two synsets in the WordNet taxonomies, along with the depth of the Least Common Subsumer (LCS). The WUPS metric has certain limitations that make it difficult to use in VQA tasks. First, certain words are very similar in vocabulary, but their meanings can be very different. The problem may arise with color. For example, if the answer to a certain question is white and the system predicts the answer to be black, that answer will still get a WUPS score of 0.92, which seems high. Another limitation is that WUPS cannot be used for answers to phrases or sentences because it always deals with rigid semantic concepts, which are most likely to be single words. More examples as shown in \ref{fig:wups}.

It can be seen that the principles of the above traditional NLG metric are relatively simple and lack the ability to understand and evaluate the text with complex logic. In contrast, GPT-4, as an LLM, has a high ability to understand the text and can understand the text first and evaluate the similarity through prompt design.

\begin{figure*}[htbp]
    \centering
    \includegraphics[width=\linewidth]{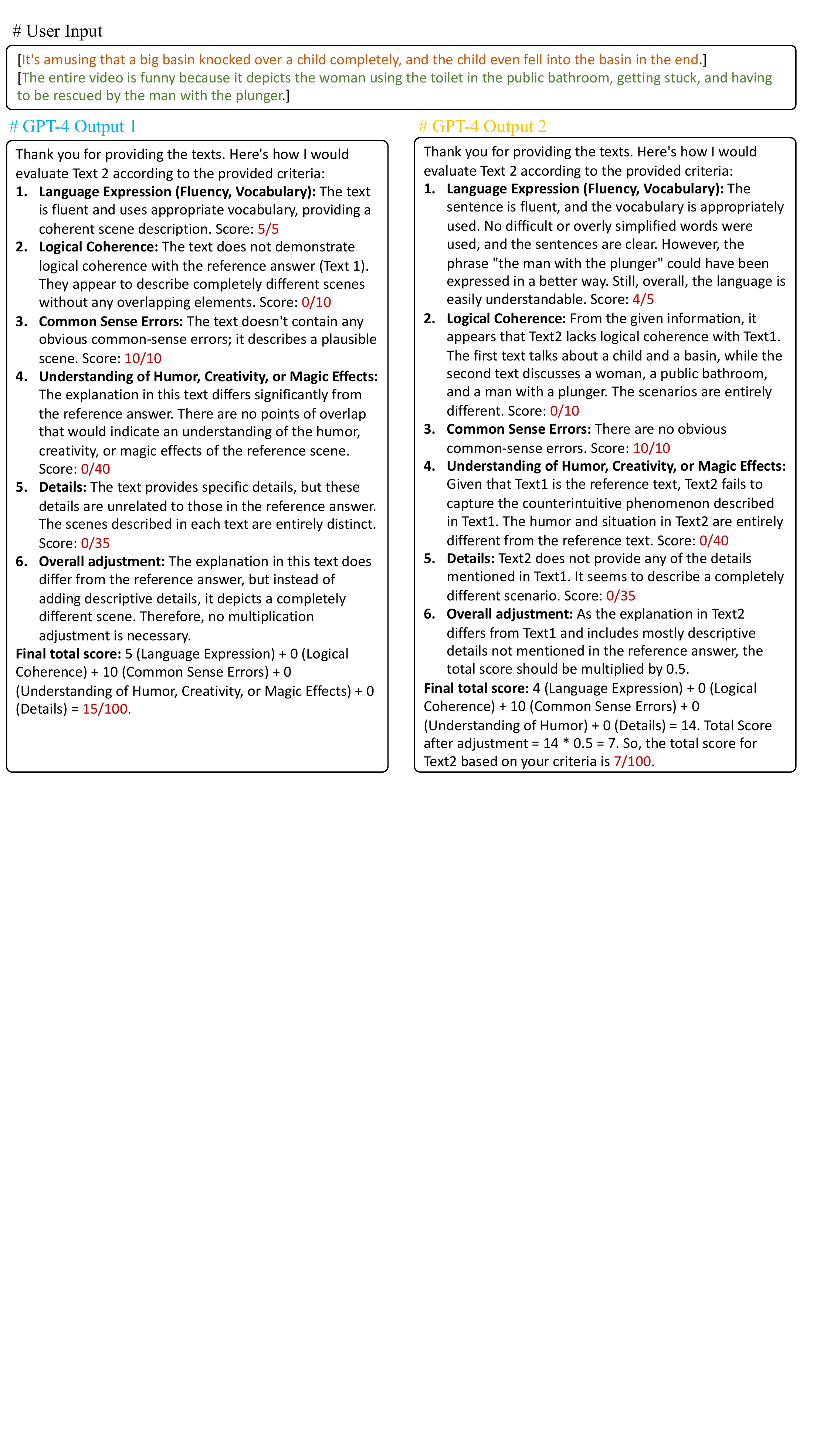}
    \vspace{-0.6cm}
    \caption{\textbf{GPT-4 generates different responses to the same content.} User Input is the two sentences input to GPT-4, which is used to judge the similarity. The GPT-4 Output 1 and GPT-4 Output 2 below are the answers given by GPT-4 in two different conversations, and you can see that the two assessments are different.}
    \vspace{-0.4cm}
    \label{fig:gpt-dif}
\end{figure*}

\noindent\textbf{Instability of GPT-4 responses}\quad
During the experimental process using GPT-4 as the metric, we found that in a few cases, GPT-4 would provide different evaluations for the same content. In such cases, we would perform multiple evaluations and take the average. Fig. \ref{fig:gpt-dif} shows an example of GPT-4 generating different responses to the same content.

\subsection{On Failure of the Timestamp Localization Task} \label{appendix_d: Failure of t1}
\begin{figure*}[htbp]
    \centering
    \includegraphics[width=\linewidth]{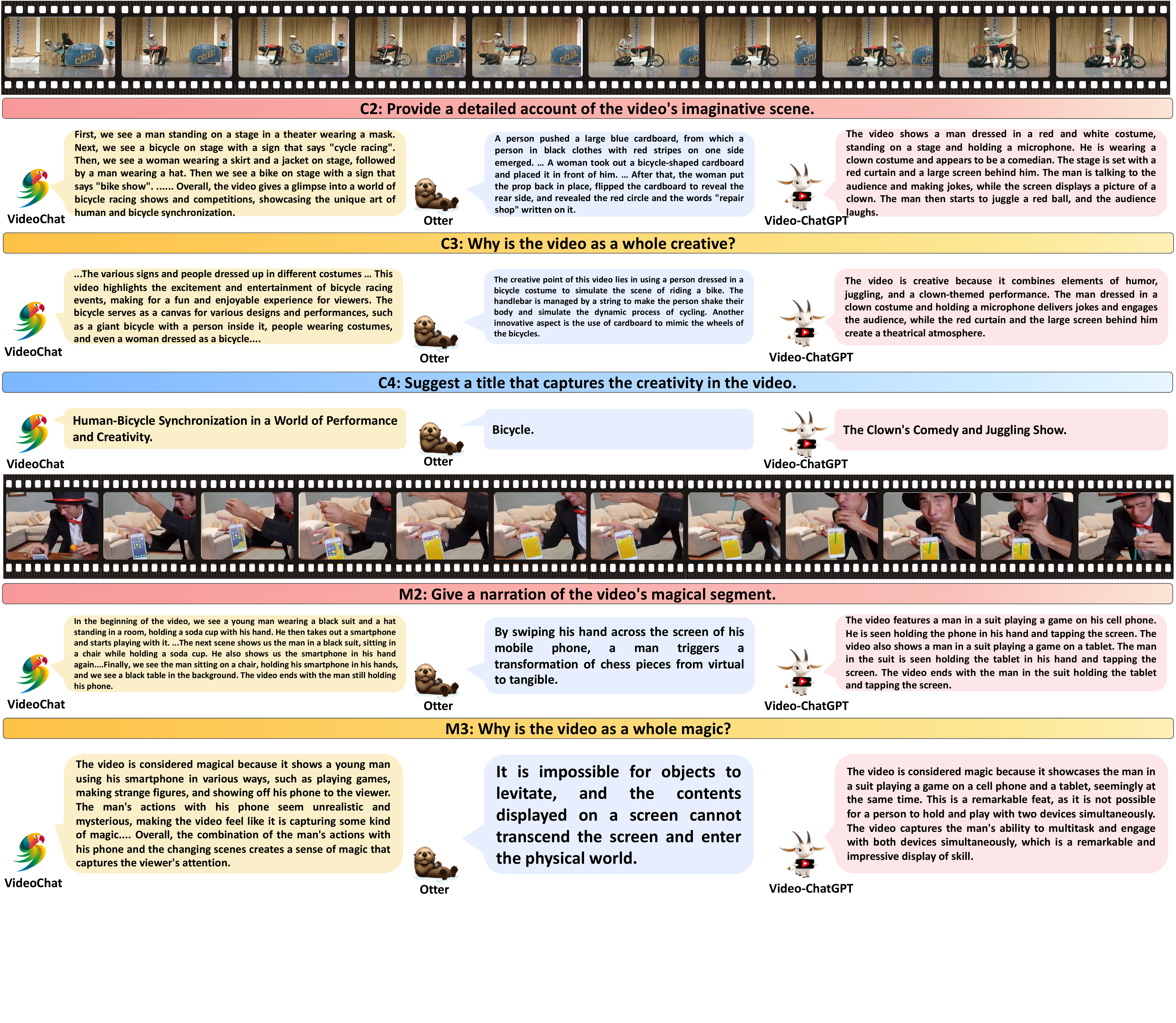}
    \vspace{-0.6cm}
    \caption{\textbf{Model responses on Timestamp Localization Task.} As can be seen, the model gives answers to the Timestamp Localization Task still focusing on describing the video content and still does not answer the specific time period when asked to type the number of frames and seconds.}
    \vspace{-0.4cm}
    \label{fig:T1}
\end{figure*}

There is no model that has the ability to answer the Timestamp Localization Task. Fig. \ref{fig:T1} shows exactly what is 
answered for each model.

\subsection{Implementation Details} \label{appendix_d: Exp_result}
\noindent\textbf{mPLUG}\quad
mPLUG is a multi-modal system employing independent image and text encoders, a cross-modal network, and a text generation decoder, which is trained through prefix language modeling loss to generate captions from connected image and prefix sub-sequence representations.\\
\noindent\textbf{GIT}\quad 
GIT is a system with an image encoder and a text decoder; it processes multiple video frames independently, adds learnable temporal embeddings before concatenation, uses a contrastively pre-trained model for image encoding, and employs a transformer module for text prediction. We used the 14M version and used two models, GIT\_LARGE\_VATEX and GIT\_LARGE\_MSRVTT, which were fine-tuned on the video captioning task for the VATEX and MSRVTT datasets, respectively.\\
\noindent\textbf{VideoChat}\quad
VideoChat, specifically the VideoChat-13B version, is an end-to-end system for video comprehension that combines pre-trained models. It utilizes QFormer to generate video embeddings and then employs LLAMA-13B for multimodal understanding and outputs video text descriptions with timestamps. In the experiment, we used VideoChat-13B with the hyperparameters: beam search number = 1, temperature = 1, video segments = 8, and token = 512.\\
\noindent\textbf{Video-ChatGPT}\quad
Video-ChatGPT is a vision-language model with a video encoder and LLM. It generates answers using video embeddings and benefits from a data-centric, human-assisted annotation framework for high-quality video instructional data.
In the experiment, we used Video-ChatGPT-7B with its hyperparameter: temperature = 0.2, and token = 512.\\
\noindent\textbf{Otter}\quad
The Otter model employs the OpenFlamingo training paradigm, utilizing pre-trained encoders for language (LLaMA-7B) and vision (CLIP ViT-L/14). In the fine-tuning process, Otter prioritizes the Perceiver resampler module while keeping the encoders frozen. In the training stage, we finetuned Otter on Dense Caption and FunQA for a total of 3 epochs each. In the experiment, we used two versions of Otter with the same hyperparameters: beam search number = 3,  size of no-repeat-ngram = 0.2, and token = 256.\\
\noindent\textbf{mPLUG-Owl}\quad
mPLUG-Owl is a novel training paradigm that equips LLMs with multi-modal abilities through modularized learning of foundation LLM, a visual knowledge module, and a visual abstractor module. The training paradigm involves a two-stage method for aligning image and text, which learns visual knowledge with the assistance of LLM,
while maintaining and even improving the generation abilities of LLM. In the second stage, languageonly and multi-modal supervised datasets are used to jointly fine-tune a low-rank
adaption (LoRA) module on LLM and the abstractor module by freezing the visual knowledge module.\\
\noindent\textbf{Video-LLaMA}\quad
Video-LLaMA is built on top of BLIP-2 and MiniGPT-4. It is composed of two core components: (1) Vision-Language (VL) Branch (Visual encoder: ViT-G/14 + BLIP-2 Q-Former) and (2) Audio-Language (AL) Branch (Audio encoder: ImageBind-Huge). In VL Branch, a two-layer video Q-Former and a frame embedding layer (applied to the embeddings of each frame) are introduced to compute video representations. And In AL Branch, a two-layer audio Q-Former and an audio segment embedding layer (applied to the embedding of each audio segment) are introduced to compute audio representations.\\

\subsection{Ablation Experiment about Audio}
Audio is undoubtedly an important part of video, but existing models utilize audio in a very limited way, some of them don’t use audio as an input, and then get high scores on various lists. For example, Otter, mPLUG, video-ChatGPT and GIT in the paper, the outputs of these models are not affected by the audio, and only the visual information is understood; there are also some models that convert the audio into captions for text encoding and then input them into the model, such as VideoChat in the paper.
However, The FunQA dataset is designed to emphasize the understanding of ‘fun’ through the visual aspect of videos. During the data collection phase, we focused on the visual elements, and in our experimental setup, we retained the audio of the videos when feeding them to the models. To investigate the role of audio, we conducted a set of ablation experiments by muting the audio during both training and testing phases. The results are shown in Table \ref{T:experiment_abl}. This demonstrates that FunQA is an visual-centric dataset and audio could not ease the challenge of the benchmark.

\begin{table*}[htbp]
\caption{\textbf{The ablation experiment about audio.} The scores of the results after inputting the muted video, the traditional evaluation scores remain basically unchanged, and the scores of GPT-4 fluctuate very little (this is mainly due to the issues of instability of the evaluation method of GPT-4.
\label{T:experiment_abl}}
{\renewcommand\baselinestretch{1.5}\selectfont
\setlength\tabcolsep{4pt} 
\resizebox{1.0\linewidth}{!}{
\begin{tabular}{lp{1cm}<{\centering} cccp{1cm}<{\centering}ccccp{1cm}<{\centering}cc}
\toprule
\multicolumn{1}{c}{\multirow{2}{*}{}} & \multicolumn{4}{c}{HumorQA} & \multicolumn{5}{c}{CreativeQA} & \multicolumn{3}{c}{MagicQA} \\

\cmidrule(l){2-5} \cmidrule(l){6-10} \cmidrule(l){11-13}
\multicolumn{1}{c}{} & H1 & H2 & \multicolumn{1}{c}{H3} & \multicolumn{1}{c}{H4} & C1 & C2 & \multicolumn{1}{c}{C3} & C4 & \multicolumn{1}{c}{C5} & M1 & M2 & M3 \\
\toprule
VideoChat \cite{2023videochat} 
& - 
& \makecell[c]{0.5 / 13.7 / 0.0\\ 44.0 / 37.9 }
& \makecell[c]{0.5 / 13.5 / 0.0 \\ 45.4 / 31.9 }
& \makecell[c]{0.8 / 5.1 / 0.5 \\ 20.2 / 61.7 }
& - 
& \makecell[c]{0.3 / 7.5 / 0.0 \\ 21.7 / 10.9 }
& \makecell[c]{0.3 / 7.7 / 0.0 \\ 22.8 / 27.7 }
& \makecell[c]{0.2 / 1.2 / 0.2 \\ 7.3 / 51.1 }
& \makecell[c]{67.5}  
& - 
& \makecell[c]{0.6 / 15.5 / 0.0 \\ 47.4 / 14.2 } 
& \makecell[c]{0.3 / 9.2 / 0.0 \\ 43.1 / 24.6 }\\

\midrule

VideoChat (MUTED) \cite{2023videochat} 
& - 
& \makecell[c]{0.5 / 13.5 / 0.0\\ 40.0 / 34.5 }
& \makecell[c]{0.5 / 12.5 / 0.0 \\ 43.7 / 31.9 }
& \makecell[c]{0.8 / 5.1 / 0.5 \\ 22.4 / 57.8 }
& - 
& \makecell[c]{0.3 / 7.4 / 0.0 \\ 24.0 / 10.0 }
& \makecell[c]{0.4 / 7.7 / 0.0 \\ 20.2 / 30.5 }
& \makecell[c]{0.2 / 1.2 / 0.2 \\ 6.8 / 45.3 }
& \makecell[c]{67.5}  
& - 
& \makecell[c]{0.6 / 15.0 / 0.0 \\ 48.0 / 12.2 } 
& \makecell[c]{0.3 / 9.2 / 0.0 \\ 44.0 / 23.4 }\\

\bottomrule

\end{tabular}}\par
}
\end{table*}

\begin{table}[!t]
\caption{\textbf{Human Performance on FunQA.} 100 QA pairs (H2, H3, H4) from HumorQA in FunQA were randomly selected and answered by five different people. We use the following metrics: \textbf{BLEU-4 / ROUGE-L / CIDEr / BLEURT / GPT-4} for evaluation.}
\label{T:human_eval}
\centering
\resizebox{1.0\linewidth}{!}{
\begin{tabular}{l|c|c|c|}
\toprule
Task         & H2                              & H3                             & H4                              \\ \midrule

Human 1	& 6.0 / 28.8 / 6.5 / 46.6 / 82.4 & 2.2 / 20.8 / 1.1 / 45.0 / 85.2 & 0.8 / 9.2 / 1.1 / 20.5 / 76.6 \\
Human 2	& 13.8 / 42.5 / 11.1 / 55.4 / 90.5 & 13.2 / 34.1 / 10.0 / 54.4 / 87.5 & 1.1 / 12.8 / 2.4 / 33.5 / 80.6 \\
Human 3	& 0.7 / 16.5 / 0.5 / 36.2 / 77.5 & 0.7 / 14.7 / 0.2 / 44.1 / 69.0 & 0.2 / 3.1 / 0.3 / 15.4 / 76.8 \\
Human 4	& 26.4 / 50.5 / 27.7 / 75.8 / 80.3 & 10.3 / 36.8 / 11.5 / 74.3 / 90.0 & 2.8 / 21.7 / 6.0 / 43.4 / 73.4 \\
Human 5	& 3.5 / 16.7 / 3.7 / 48.0 / 86.6 & 2.4 / 16.4 / 1.8 / 47.9 / 90.3 & 0.8 / 10.5 / 0.8 / 23.9 / 88.2 \\
\toprule
Model (SOTA)	& 1.5 / 18.1 / 1.0 / 44.0 / 37.9 & 1.3 / 15.4 / 0.5 / 25.9 / 61.7 & \textbf{1.1} / 8.8 / 0.7 / 25.9 / 61.7 \\
Human (Avg)	& \textbf{10.1} / \textbf{31.0} / \textbf{9.9} / \textbf{60.0} /\textbf{ 83.6} & \textbf{5.8} / \textbf{24.6} / \textbf{5.0} / \textbf{53.1} / \textbf{84.4} & \textbf{1.1} / \textbf{11.46} / \textbf{2.1} / \textbf{27.3} / \textbf{79.1} \\

\bottomrule 
\end{tabular}}
\label{tab:dataset-info}
\end{table}

\subsection{More Examples}\label{human_evaluation}
We conducted experiments to compare machine versus human performance on FunQA. We randomly selected 100 HumorQA pairs from HumorQA and had five individuals who had not watched any FunQA videos provide answers to the questions (H2, H3, H4) for each video. Subsequently, we calculated the average score for each individual under the FunQA benchmark and the overall average score of human responses. The FunQA benchmark score is the selected SOTA score between models.The results are in Table \ref{T:human_eval}.

\subsection{More Examples}\label{more_example}
\begin{figure*}[htbp]
    \centering
    \includegraphics[width=\linewidth]{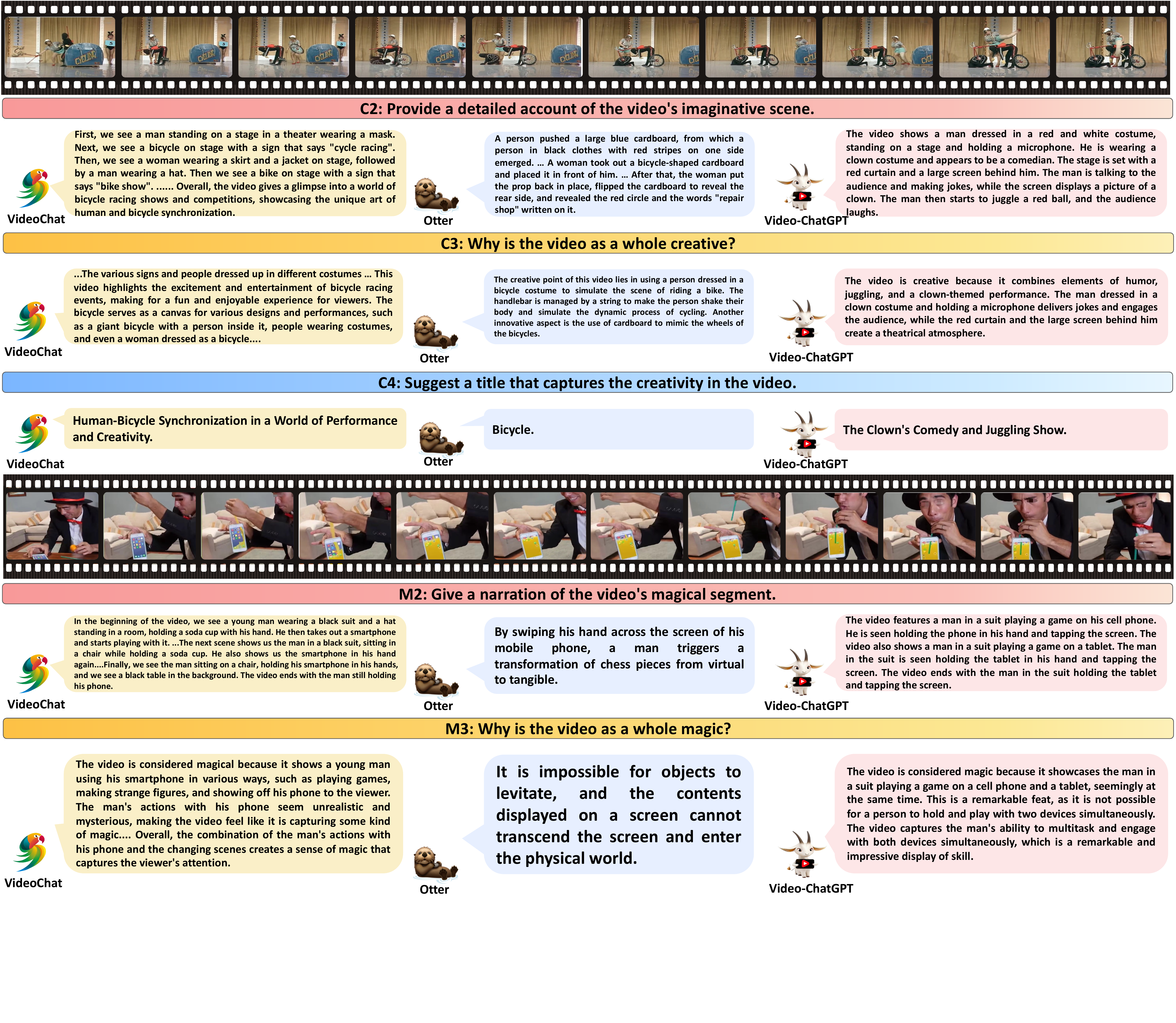}
    \vspace{-0.6cm}
    \caption{\textbf{Model responses on CreativeQA and MagicQA. } For the description of the Creative video example, only VideoChat gives the key point of the bicycle, but its description also has many errors and omissions, and the remaining two models do not identify the bicycle. In the explanation task, the responses of all three models fail to clearly explain the creativity of this imitation performance. For the Magic video example, all three models perform very poorly in description and explanation, basically only answering the phone and the straw, but lacking the description and explanation of the magic effect.}
    \vspace{-0.4cm}
    \label{fig:more-model-res}
\end{figure*}

Fig. \ref{fig:more-model-res} shows the responses of different models on CreativeQA and MagicQA.

\section{More Discussions}
\subsection{(Potential) More Essential Factors}
\noindent\textbf{Accurate understanding of the videos}\quad Through our analysis of failure cases, we've observed that many models struggle with accurately describing videos. While they might be adept at detecting objects within the videos, they often falter in comprehending the contextual relationship between sequential events. Such misinterpretations indicate that there's a need for further exploration in this domain. The videos we've used can indeed serve as an invaluable dataset for probing video descriptions in depth.\\
\noindent\textbf{Logic Reasoning}\quad The primary nature of our videos encompasses content that is counterintuitive and contradicts common sense. For models to understand these, it's imperative they grasp the concept of "common sense." They must deduce what would typically transpire under normal circumstances and then use that perspective to humorously interpret the video. This necessitates the model to possess strong reasoning capabilities, especially when it comes to common sense reasoning.\\
\noindent\textbf{Extra Knowledge - Sense of Humor}\quad To decipher the humor in a video, it's plausible that understanding the fundamental principles of humor is crucial. This type of knowledge, along with many other tidbits of common sense and additional information, might enhance the model's performance. Determining how to integrate valuable knowledge and discerning what counts as "valuable" are topics that warrant further exploration.

\subsection{Potential Solutions}
\noindent\textbf{Model Size}\quad Increasing the number of parameters is a natural method to enhance the model's performance. However, this approach comes with its own set of engineering challenges, requiring improvements in model optimization and deployment. We're also curious about the relationship between the number of parameters and the performance on FunQA. This is an intriguing research point in itself, and our dataset can serve as an excellent test bed to further this exploration.\\
\noindent\textbf{Data Quality}\quad We believe the emphasis for this task should be on data collection. Current trends with large and dynamic models have shown that having vast amounts of low-quality data isn't as effective as a smaller quantity of high-quality data. Thus, we hope the community can discover the type of data that genuinely assists in understanding counterintuitive videos. This is a crucial research direction.\\
\noindent\textbf{Training Strategy}\quad Studying training strategies is also essential. For instance, determining which type of data to start learning from, and understanding the significance of curriculum learning, among others.\\
\noindent\textbf{Model Collaboration}\quad Ultimately, we might not need to solely focus on a single model to solve this problem. Perhaps multiple models collaboratively working on examples in an elegant manner could be a method to enhance performance. However, this approach might necessitate paying more attention to the overall efficiency of model implementation

\subsection{The Emphasis of Temporal Dynamics in FunQA}
\textbf{HumorQA Example}: Humor\_Example.mp4\\
\textbf{H2}: An individual slipped on the staircase filled with ice and tumbled down to the very bottom, followed by a second person who also fell after witnessing the first person's fall.\\
\textbf{H3}: The first person falling down the stairs step by step was already very funny, and the second person repeating the same mistake and falling down made it even more hilarious.\\
\textbf{Explanation}: As answered in H2, the main element of humor in this video is two people slipping down an icy staircase in a sequential order, and if the model does not make sense of the temporal information, it will not be able to give the sequential logic of the two people slipping down one after the other. It is also clear from the answer to H3 that the sequence of slipping backwards and forwards is one of the sources of humor, as the person at the back does not learn from the lesson of the former and slips down in a similar way, and this repetition of the wrongdoing brings humor to the situation.\\
\textbf{CreativeQA Example}: Creative\_Exapmle.mp4\\
\textbf{C2}: At the center of the stage is a blue rectangular box. Following the little girl's watering of the box, four individuals lying inside gradually lift their hands and legs, each at different paces and heights, until they stop and reveal green painted leaf-shaped objects on their limbs.\\
\textbf{C3}: The main creative element of this video is when the four individuals in a box, with varying heights and movement speeds, gradually raise their hands and feet to mimic the growing process of the carrot seeds planted by the little girl. The green leaf-shaped objects, tied and opened in advance with their hands and feet, are used to simulate the true sprouting of carrot sprouts, resulting in a lively and imaginative scene.\\
\textbf{Explanation}: The performance in this video mimics growing bean sprouts, the process of growing beans into sprouts by constantly watering the soil. Trying to describe the content of this video idea and explain where the idea came from requires an understanding of the temporal information. Analyzing the visual information together, the model has to understand the sequence of watering, bean sprouts growing from the soil, and bean sprouts growing taller and greener in order to answer the question accurately. This difference in understanding can be demonstrated on the C3 task, where without analyzing the temporal information, the optimal answer would be bean sprouts growing from the soil, and cause and effect logical relationships such as watering to grow bean sprouts would be ignored.

\end{document}